%% file: main.tex
\documentclass[10pt,twocolumn,letterpaper]{article}

\usepackage{cvpr}              


\input{math_commands.tex}

\definecolor{cvprblue}{rgb}{0.21,0.49,0.74}
\usepackage[pagebackref,breaklinks,colorlinks,allcolors=cvprblue]{hyperref}

\def\paperID{993} 
\def\confName{CVPR}
\def\confYear{2025}

\usepackage{hyperref}
\usepackage{url}
\usepackage{xcolor}
\usepackage{graphicx}
\usepackage{multirow}
\usepackage{wrapfig}
\usepackage{booktabs}
\usepackage{makecell}
\usepackage[export]{adjustbox}

\title{FlexiDiT: Your Diffusion Transformer Can Easily Generate High-Quality Samples with Less Compute}

\author{$\text{Sotiris Anagnostidis}^{\dagger, \S, *}$ \and $\text{Gregor Bachmann}^{\dagger, \S}$ \and $\text{Yeongmin Kim}^{\dagger, \ddagger}$ \and $\text{Jonas Kohler}^{\dagger}$ \and $\text{Markos Georgopoulos}^{\dagger}$ \and $\text{Artsiom Sanakoyeu}^{\dagger}$ \and $\text{Yuming Du}^{\dagger}$ \and $\text{Albert Pumarola}^{\dagger}$ \and $\text{Ali Thabet}^{\dagger}$ \and $\text{Edgar Schönfeld}^{\dagger}$}

\usepackage[rawfloats=true]{floatrow}
\usepackage{indentfirst}

\makeatletter
\renewcommand{\paragraph}{%
  \@startsection{paragraph}{4}%
  {\z@}{1.0ex \@plus 1ex \@minus .2ex}{-1em}%
  {\normalfont\normalsize\bfseries}%
}
\makeatother

\newcommand\blfootnote[1]{%
  \begingroup
  \renewcommand\thefootnote{}\footnote{#1}%
  \addtocounter{footnote}{-1}%
  \endgroup
}

\begin{document}

\twocolumn[{
\maketitle
    \begin{center}
        \vspace{-9mm}
        \includegraphics[width=1.\linewidth,center]{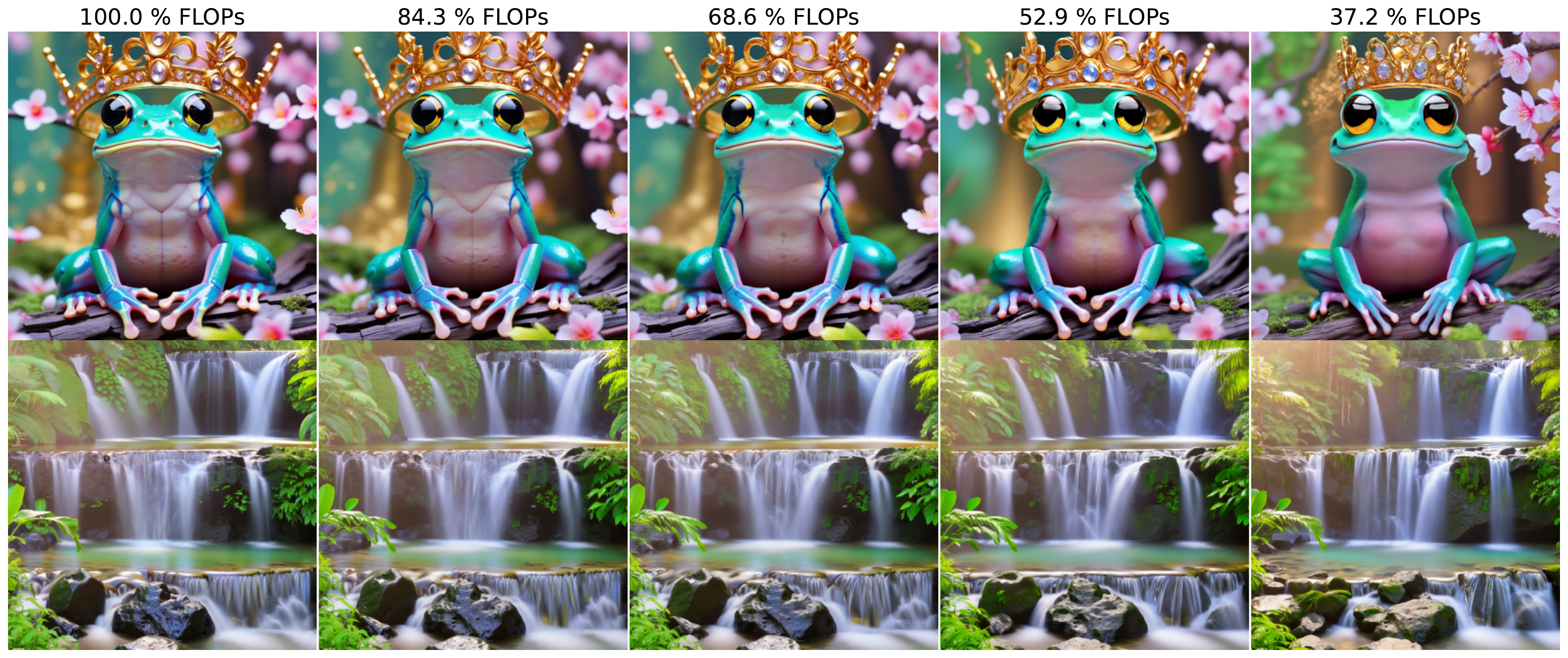}
        \captionof{figure}{We \emph{flexify} DiTs and adjust the compute per diffusion step, generating high-quality samples with significantly less compute.}
        \label{fig:emu-examples-0}
    \end{center}
}]


\newcommand{\pixart}{\textit{T2I Transf.}}
\newcommand{\emu}{\textit{Emu}}
\newcommand{\moviegen}{\textit{Video DiT}}
\newcommand{\flexidit}{\textit{FlexiDiT}}

\definecolor{pcolor}{HTML}{18255f}
\newcommand{\pcolorcommand}[2]{\newcommand{#1}{{\color{pcolor} #2}}}
\pcolorcommand{\p}{p}
\pcolorcommand{\q}{q}
\pcolorcommand{\condition}{c}
\pcolorcommand{\qzero}{q_0}
\pcolorcommand{\pt}{p_t}
\pcolorcommand{\pzero}{p_0}
\pcolorcommand{\pdata}{p_\text{data}}
\pcolorcommand{\pnoise}{p_\text{noise}}
\pcolorcommand{\ptheta}{p_{\theta}}
\pcolorcommand{\thetat}{{\theta}_t}
\pcolorcommand{\pthetat}{p_{\theta_t}}
\pcolorcommand{\etheta}{\epsilon_{\theta}}
\pcolorcommand{\mutheta}{\mu_{\theta}}
\pcolorcommand{\Sigmatheta}{\Sigma_{\theta}}
\pcolorcommand{\x}{\mathbf{x}}

\definecolor{datacolor}{HTML}{bd3817}
\newcommand{\datacolorcommand}[2]{\newcommand{#1}{{\color{datacolor} #2}}}
\datacolorcommand{\wdim}{w}
\datacolorcommand{\hdim}{h}
\datacolorcommand{\fdim}{f}
\datacolorcommand{\guidancescale}{s_{\text{cfg}}}
\datacolorcommand{\guidancescaleone}{s_{\text{cfg1}}}
\datacolorcommand{\guidancescaletwo}{s_{\text{cfg2}}}
\datacolorcommand{\patchsize}{p}
\datacolorcommand{\patchsizef}{p_\text{f}}
\datacolorcommand{\patchsizeh}{p_\text{h}}
\datacolorcommand{\patchsizew}{p_\text{w}}
\datacolorcommand{\patchsizeexp[1]}{p^{#1}}
\datacolorcommand{\patchsizecondition}{p_{\text{cond}}}
\datacolorcommand{\patchsizeuncondition}{p_{\text{uncond}}}
\datacolorcommand{\denoisestep}{s}
\datacolorcommand{\denoisestepexp[1]}{s^{#1}}
\datacolorcommand{\patchsizecurrent}{p_{\text{current}}}
\datacolorcommand{\patchsizeunderscore[1]}{p_{\text{#1}}}

\definecolor{modelcolor}{HTML}{2F3A6F}
\newcommand{\modelcolorcommand}[2]{\newcommand{#1}{{\color{modelcolor} #2}}}
\modelcolorcommand{\embedding}{M_{\text{embed}}}
\modelcolorcommand{\flexembedding}{M^{\text{flex}}_{\text{embed}}}
\modelcolorcommand{\weightembedding}{w_{\text{embed}}}
\modelcolorcommand{\flexweightembedding}{w^\text{flex}_{\text{embed}}}
\modelcolorcommand{\biasembedding}{b_{\text{embed}}}
\modelcolorcommand{\flexbiasembedding}{b^\text{flex}_{\text{embed}}}

\modelcolorcommand{\deembedding}{M_{\text{de-embed}}}
\modelcolorcommand{\flexdeembedding}{M^\text{flex}_{\text{de-embed}}}
\modelcolorcommand{\weightdeembedding}{w_{\text{de-embed}}}
\modelcolorcommand{\flexweightdeembedding}{w^\text{flex}_{\text{de-embed}}}
\modelcolorcommand{\biasdeembedding}{b_{\text{de-embed}}}
\modelcolorcommand{\flexbiasdeembedding}{b^\text{flex}_{\text{de-embed}}}

\modelcolorcommand{\project}{Q}
\modelcolorcommand{\projectembedding}{Q_{\text{embed}}}
\modelcolorcommand{\projectdeembedding}{Q_{\text{de-embed}}}

\newcommand{\ImageNet}{\textit{ImageNet}}

\definecolor{constantcolor}{HTML}{bd3817}
\newcommand{\constantcolorcommand}[2]{\newcommand{#1}{{\color{constantcolor} #2}}}
\constantcolorcommand{\depth}{L}
\constantcolorcommand{\tokens}{N}
\constantcolorcommand{\hiddensize}{d}
\constantcolorcommand{\hiddensizeinput}{d_\text{input}}
\constantcolorcommand{\hiddensizeoutput}{d_\text{output}}
\constantcolorcommand{\hiddensizelora}{d_\text{lora}}
\constantcolorcommand{\diffusionsteps}{T}
\constantcolorcommand{\diffusionstepsweak}{T_\text{weak}}
\constantcolorcommand{\diffusionstepspowerful}{T_\text{powerful}}
\constantcolorcommand{\inchannels}{c_\text{in}}
\constantcolorcommand{\outchannels}{c_\text{out}}

\blfootnote{\thanks{${}^*$Work done during an internship at Meta GenAI. ${}^{\dagger}$Meta GenAI. ${}^{\S}$ETH Zürich. $\ddagger$KAIST. Correspondence to: \texttt{sanagnos@ethz.ch}}}

\input{sec/0_abstract}    
\input{sec/1_intro}
\input{sec/2_background}
\input{sec/3_flexible_dit}
\input{sec/4_1_class_conditioned}
\input{sec/4_2_text_conditioned}
\input{sec/4_3_video_generation}
\input{sec/4_4_latency}
\input{sec/5_coclusion}

{
    \small
    \bibliographystyle{ieeenat_fullname}
    \bibliography{main}
}

\newpage
\onecolumn
\appendix

\input{sec/appendix}

\end{document}

%% file: math_commands.tex
\usepackage{amsmath,amsfonts,bm}

\def\eqref#1{equation~\ref{#1}}

\def\1{\bm{1}}

\DeclareMathAlphabet{\mathsfit}{\encodingdefault}{\sfdefault}{m}{sl}
\SetMathAlphabet{\mathsfit}{bold}{\encodingdefault}{\sfdefault}{bx}{n}

\newcommand{\R}{\mathbb{R}}



%% file: sec/0_abstract.tex
\begin{abstract}
Despite their remarkable performance, modern Diffusion Transformers (DiTs) are hindered by substantial resource requirements during inference, stemming from the fixed and large amount of compute needed for each denoising step. In this work, we revisit the conventional static paradigm that allocates a fixed compute budget per denoising iteration and propose a dynamic strategy instead. Our simple and sample-efficient framework enables pre-trained DiT models to be converted into \emph{flexible} ones --- dubbed~\flexidit --- allowing them to process inputs at varying compute budgets. We demonstrate how a single \emph{flexible} model can generate images without any drop in quality, while reducing the required FLOPs by more than $40$\% compared to their static counterparts, for both class-conditioned and text-conditioned image generation. Our method is general and agnostic to input and conditioning modalities. We show how our approach can be readily extended for video generation, where~\flexidit~models generate samples with up to $75$\% less compute without compromising performance.
\end{abstract}

%% file: sec/1_intro.tex
\section{Introduction}
\label{sec:intro}

\begin{figure*}[!ht]
    \centering
    \begin{minipage}{0.66\textwidth}
        {{\includegraphics[width=1\textwidth]{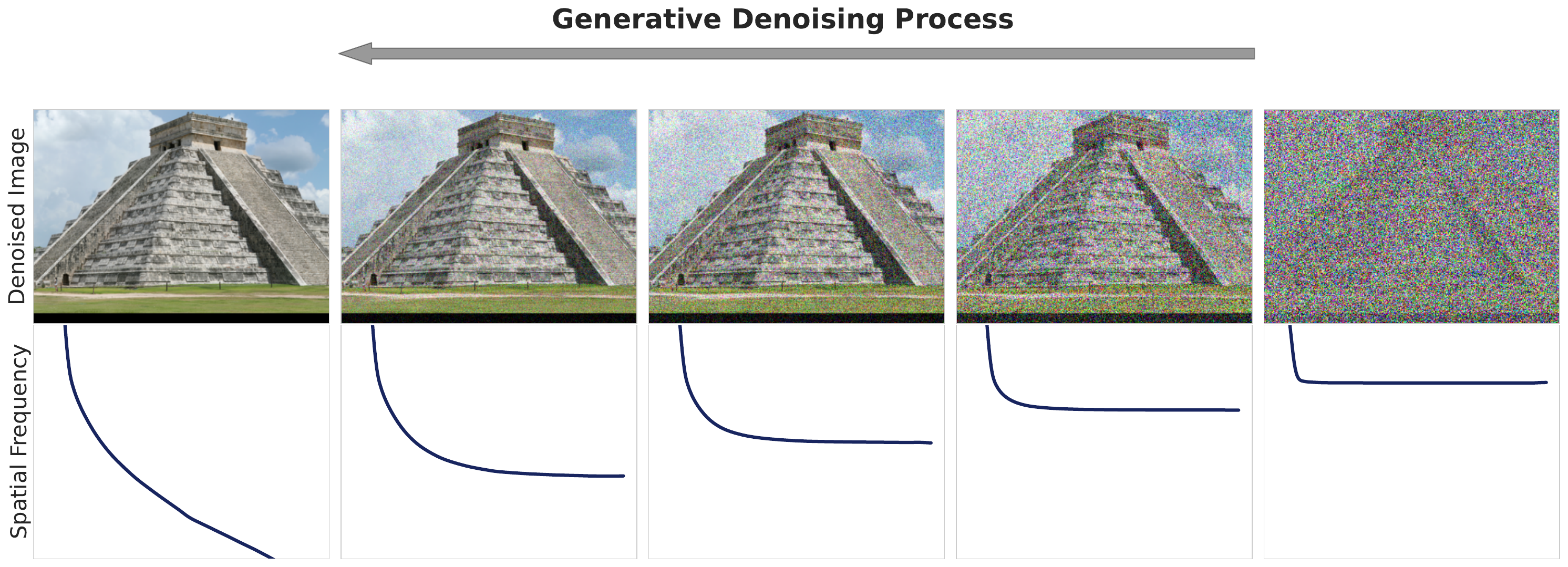} }}%
    \end{minipage}
    \begin{minipage}{0.32\textwidth}
        \vspace{4mm}
        {{\includegraphics[width=1\textwidth]{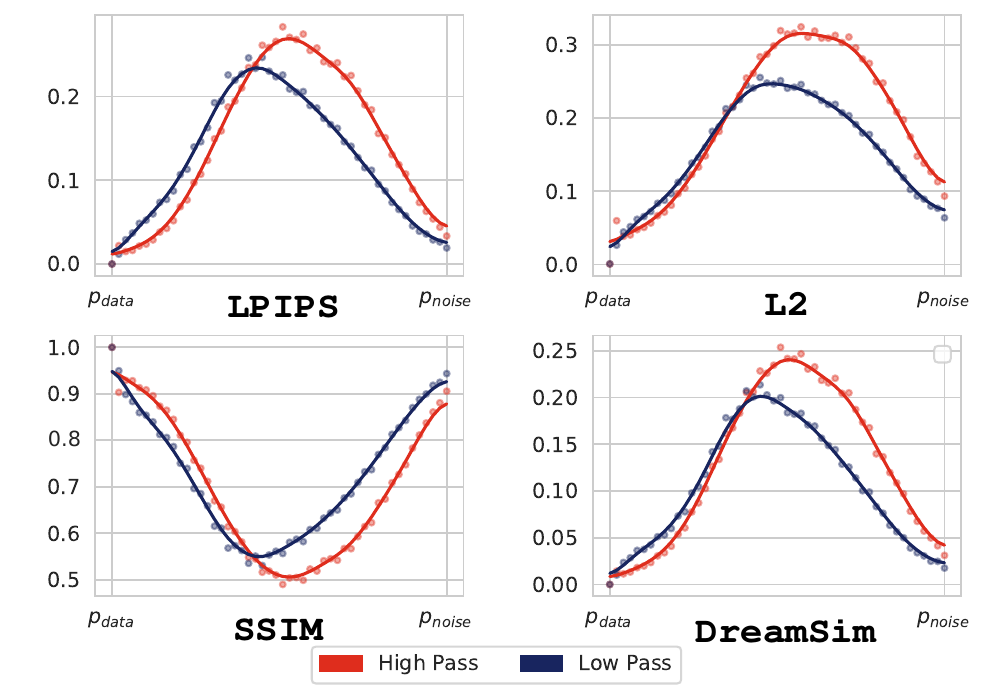} }}%
    \end{minipage}
    \caption{Diffusion can be viewed as spectral autoregression~\citep{dieleman2024spectral}. \textbf{Left:} Diffusion and its effect on the spatial frequency of images. \textbf{Right:} To investigate the role of different frequency components in image generation, we apply a low or high pass filter to a single diffusion step update (while keeping all other updates unchanged). With all other sources of randomness fixed, we compare the generated samples with and without filtering using LPIPS~\citep{zhang2018unreasonable}, $L_2$ distance of the pixels, SSIM~\citep{wang2004image} and DreamSim~\citep{fu2023dreamsim}. Notably, the influence of low and high pass filters varies depending on whether they are applied early or late in the denoising process.}%
    \label{fig:motivation}
    \vspace{1mm}
\end{figure*}

Diffusion models~\citep{sohl2015deep} have recently become the core building block of major improvements in image generation~\citep{dhariwal2021diffusion, videoworldsimulators2024, sd, flux, yang2024cogvideox}. These models gradually denoise a random sample $\x_t$, drawn typically from  $\pnoise = \mathcal{N}(\mathbf{0}, \mathbf{I})$, by iteratively calling a neural network trained to reverse a pre-defined noise corruption process $\ptheta (\x_{t - 1} | \x_{t})$. This process enables the generation of samples from the desired distribution $\pdata$.
The remarkable performance of these models is closely tied to the amount of computational resources invested in them, as evidenced by established scaling laws \citep{kaplan2020scaling}. The Transformer architecture~\citep{vaswani2017attention} has proven to be highly scalable across various modalities, leading to its adoption in diffusion models, in the form of the recent Diffusion Transformer~\citep{peebles2023scalablediffusionmodelstransformers}. In DiTs, the denoising process $\ptheta (\x_{t - 1} | \x_{t})$ is parametrized using Transformer blocks instead of traditional convolutional layers.
As popularized in Vision Transformers~\citep{dosovitskiy2020image}, (latent) images of dimension $\hdim \times \wdim$ are divided into patches of size $\patchsize \times \patchsize$, which serve as input tokens that are transformed via a series of attention and feed-forward layers. The use of DiTs offers two key advantages: (i) a unified architecture and input processing framework that facilitates multimodal applications and generalization to other domains, such as audio and video; and (ii) exceptional scalability due to their signal propagation characteristics and efficient training on modern hardware.

The computational complexity of a DiT with hidden dimension $\hiddensize$ and depth $\depth$ is $\mathcal{O}(\depth \tokens \hiddensize^2 \ + \depth \tokens^2 \hiddensize)$, where $\tokens$ corresponds to the total number of tokens\footnote{For high-resolution image generation --- even when diffusion is performed in a latent space --- the second term $\mathcal{O}(\depth \tokens^2 \hiddensize)$ that corresponds to the attention operations, can quickly become the main bottleneck.}. Given $\diffusionsteps$ steps of the diffusion process, function calls to the same monolithic DiT model are repeated for all $\diffusionsteps$ steps, and the total amount of compute is thus \emph{uniformly} allocated. However, image generation via diffusion exhibits distinct, non-uniform, and temporally --- with respect to the noise process --- varying characteristics. Consistent with intuition, prior work has observed that high-level image features tend to emerge at early generative diffusion steps (i.e., large $t$'s). In contrast, later steps refine and progressively generate high-frequency and detailed visual features~\cite{castillo2023adaptive}. We illustrate further differences during the denoising generation in Fig.~\ref{fig:motivation}. Concisely, \textit{different denoising steps have profoundly different influence on high and low-level features of the resulting images}.

In theory, different denoising methods and models can be employed for each step $t$, i.e. one can use separate parameters $\thetat$ to learn $\pthetat (\x_{t - 1} | \x_{t})$. While this notion has been previously explored in the literature, prior works propose computationally intensive solutions, such as training separate expert denoisers for different $t$'s~\citep{balaji2022ediff} or using model cascades~\citep{stablecascades}. These approaches face two significant limitations: (i) they require multiple models to be managed during inference, increasing memory demands. This can quickly become a major issue, especially as current trends favor the use of larger and larger models. Moreover, (ii) using separate models restricts the opportunity for knowledge sharing across steps. Although one can treat denoising steps independently, the denoising process itself retains certain shared characteristics across steps. This inherent smoothness implies that separate models would need to learn these shared properties individually. In this work, we address these challenges and instead propose to perform different denoising steps with varying levels of compute using a single model. 

\vspace{4mm}

In summary, our contributions are the following:
\begin{itemize}
    \item We present a simple framework that \emph{flexifies} DiT models, allowing them to convert samples into different sequences of tokens by adjusting the patch size. Processing samples as different sequences enables us to control the compute budget being used for each denoising step.
    \item By leveraging specific image characteristics at different denoising steps, we demonstrate that strategically allocating less compute to certain steps can yield significant computational savings (over 40\%) \emph{without compromising} the quality of generated samples for both class-conditioned and text-conditioned image generation. We also show how denoising based on a larger patch size can serve as a more effective guidance signal.
    \item  We illustrate the versatility of our framework by extending it to other modalities. For video generation, we achieve substantial computational savings (up to $75$\%) \emph{with no considerable} drop in performance or conceptual differences in the generated samples.
\end{itemize}

%% file: sec/2_background.tex
\section{Background}
\label{sec:background}

\begin{wrapfigure}{r}{0.18\textwidth}
  \begin{center}
    \vspace{-8mm}
    \hspace{-8mm}
    \includegraphics[width=0.22\textwidth]{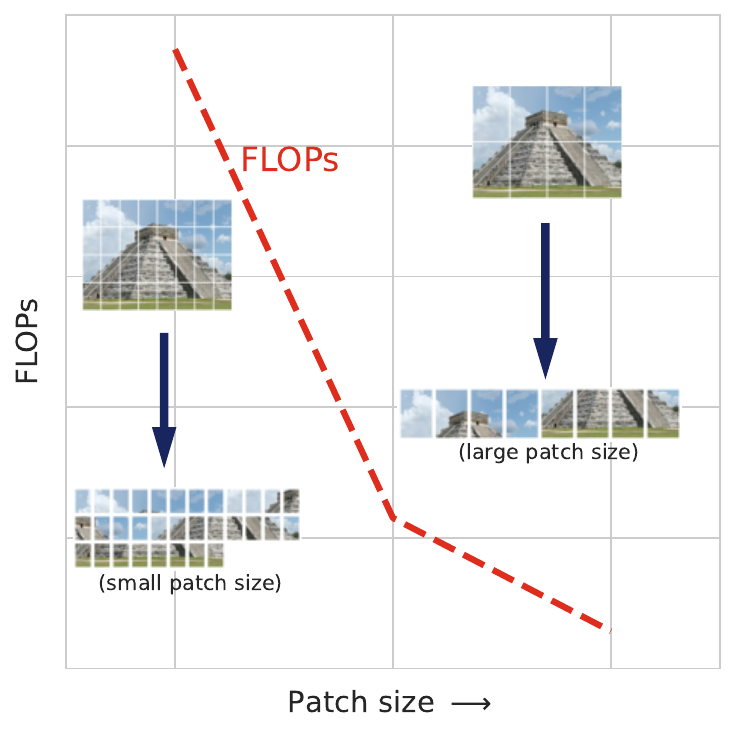}
    \caption{Tokenizing images into patches.}
    \vspace{-6mm}
  \end{center}
\end{wrapfigure}
For simplicity, we limit the discussion here to images, and detail later changes due to different modalities.
DiTs use a Transformer encoder to process image patches as tokens. Hereafter, we refer as \emph{tokenization} to the process of converting a (latent) image into a series of tokens and as \emph{de-tokenization} to the opposite process, of transforming a series of tokens back into an image. Given an image of size $\hdim \times \wdim$ and a chosen patch size\footnote{In reality, different patch sizes can be considered along the height and the width dimensions. We will however refrain from doing that.} $\patchsize$, an input image is cropped into non-overlapping patches of dimensions $\R^{\patchsize \times \patchsize \times \inchannels}$, where $\inchannels$ denotes the number of channels of the input image. The total number of tokens is then equal to $\tokens = (\hdim / \patchsize) \times (\wdim / \patchsize)$. This (flattened) sequence of patches is then projected using a linear layer $\embedding$ with weights $\R^{\patchsize * \patchsize * \inchannels \times \hiddensize}$ and potential biases $\R^\hiddensize$. This tokenization process is equivalent to performing a $2$D convolution, where the kernel size and stride are both equal to $\patchsize \times \patchsize$. The embedded $\tokens$ tokens are then processed using $\depth$ Transformer encoder layers. The output tokens of dimension $\R^{\tokens \times \hiddensize}$ are projected back to the image space with a linear de-embedding layer $\deembedding$ with weights $\R^{\hiddensize \times \outchannels * \patchsize * \patchsize}$ and potential biases $\R^{\outchannels * \patchsize * \patchsize}$. Here $\outchannels$ denotes the number of output channels, typically $\outchannels = 2 \inchannels$ if the prediction includes the variance, else $\outchannels = \inchannels$. DiTs adhere to the scaling properties of Transformers, as demonstrated in various other applications~\citep{kaplan2020scaling,hoffmann2022training,henighan2020scaling,zhai2022scaling, bachmann2024scaling}. 

DiTs are an increasingly popular alternative to convolutional networks for denoising corrupted images during generation, i.e. modeling $\ptheta (\x_{t - 1} | \x_{t})$. Diffusion defines two Markov chains, the forward and the backward process. During the forward process, Gaussian noise\footnote{Although we focus on Gaussian noise here, other corruptions apart from Gaussian noise have also been analysed~\citep{bansal2024cold, nachmani2021denoising, daras2022soft}.} is added to samples of the real distribution $\x_0 \sim \q(\x_0)$~\citep{ho2020denoising}:

\begin{multline}
    \label{eq:noise_images}
    \q (\x_{1:\diffusionsteps} | \x_0) = \prod_{t=1}^{\diffusionsteps} \q(\x_{t} | \x_{t - 1}), \quad \text{where} \\ \quad \q(\x_{t} | \x_{t - 1}) = \mathcal{N}(\x_{t - 1}; \sqrt{1 - \beta_t}\x_{t - 1}, \beta_t \mathbf{I}).
\end{multline}
In the backward process, samples are drawn from a Gaussian noise distribution $\p(\x_\diffusionsteps) = \pnoise = \mathcal{N}(\mathbf{0}, \mathbf{I})$ and then gradually denoised, using Tweedie’s Formula ~\citep{efron2011tweedie}:

\begin{multline} 
\label{eq:denoise_images}
\ptheta(x_{\diffusionsteps:0}) = \p(\x_\diffusionsteps) \prod_{t=\diffusionsteps}^1 \ptheta(\x_{t - 1} | \x_{t}), \quad \text{where} \\ \quad \ptheta(\x_{t - 1} | \x_{t}) = \mathcal{N}(\x_{t}; \mutheta(\x_t, t), \Sigmatheta(\x_t, t)).
\end{multline}
Typically, a \emph{single} model is used to model the prediction $\ptheta(\x_{t - 1} | \x_{t})$ for every $t$.

%% file: sec/3_flexible_dit.tex
\section{Flexible Diffusion Transformers}
\label{sec:class_conditioned}

\begin{figure*}[!ht]
    \centering
    \hfill
    \includegraphics[width=0.61\textwidth]{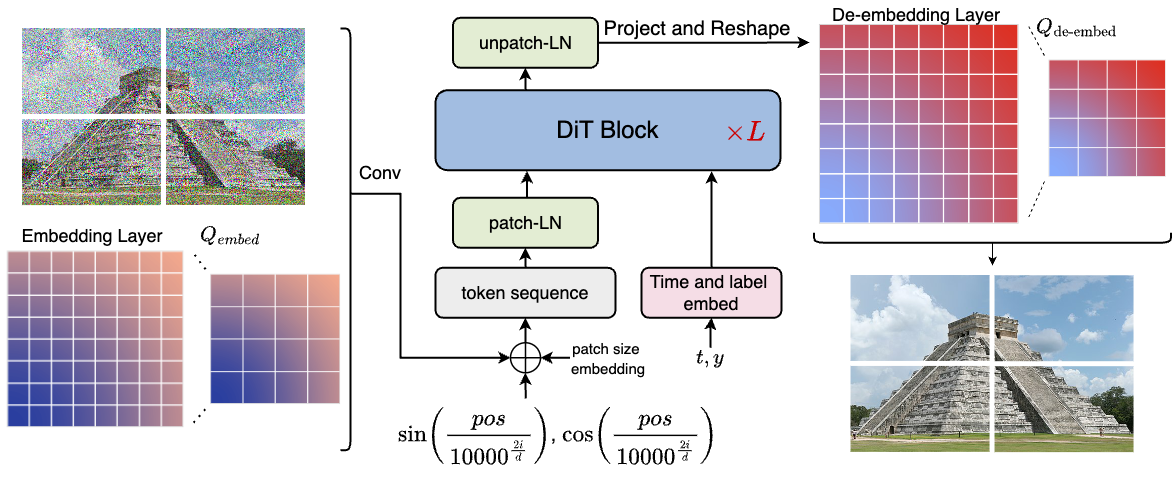}
    \hfill
    \includegraphics[width=0.35\textwidth]{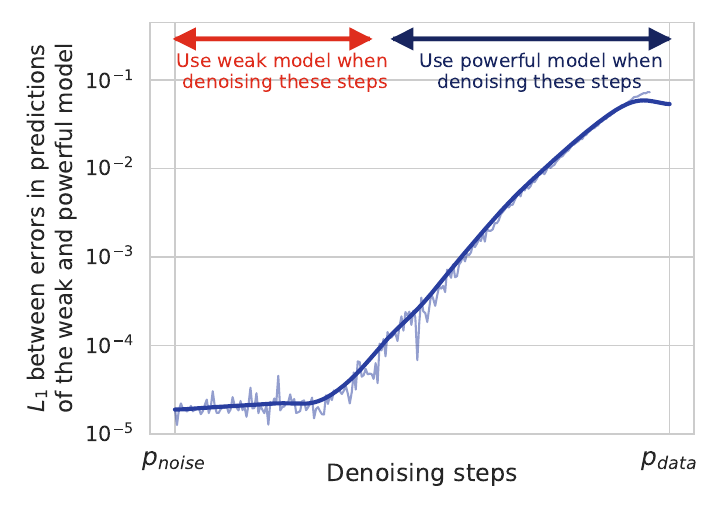}
    \hfill
    \caption{\textbf{Left:} We flexify DiTs by allowing them to process images with more patch sizes, by changing the lightweight embedding and de-embedding layers. We showcase this for a class-conditioned~\ImageNet~model. \textbf{Right:} We plot the difference in predictions between a weak and a powerful model. For the first denoising steps, differences are small, and thus using the weak model there allows accelerated generation without performance degradation.}
    \label{fig:methodology}
\end{figure*}

We \emph{flexify} DiTs, via small architectural changes, that allow them to process images as different length sequences, by adjusting the patch size $\patchsize$ used in tokenization. 
Flexible tokenization of images has been utilized before for single-step inference applications~\citep{beyer2023flexivit} and to accelerate training~\citep{anagnostidisnavigating}. We instead propose to use different patch sizes at different steps in the denoising process of the same image. This is based on the following intuition:
\begin{center}
    \textit{Early steps focus on low-frequency details, which can be performed with bigger patch sizes at the same quality.}
\end{center}
Changing the patch size $\patchsize$ directly affects the total number of tokens (recall that $\tokens = (\hdim / \patchsize) \times (\wdim / \patchsize)$) and thus the overall compute required for a function evaluation. Hence, by leveraging bigger patch sizes only for early steps, we can reduce the overall generation time while maintaining both low and high-frequency details in the produced images.

We obtain a flexible model --- coined {\flexidit} --- by modifying and fine-tuning a pre-trained DiT, which enables it to process and understand images with new patch sizes. In our experiments we focus on efficiency, so newly added patch sizes are always larger than the one of the pre-trained model, leading to fewer tokens and thus a smaller computational footprint. The single~\flexidit~model can be \emph{instantiated} in different modes depending on the selected patch size. We will refer to instances of the model that use the original and smaller patch size $\patchsizeunderscore[powerful]$ as \emph{powerful}, compared to using a \emph{weak} model with a larger patch size $\patchsizeunderscore[weak]$. To simplify the discussion, we largely ignore how additional conditioning may be applied for now, and instead refer to specific implementation details in the experiments section.

DiTs --- as any Transformer --- can process sequences of any arbitrary length $\tokens$. Fundamentally, we just need to modify (i) how tokenization and de-tokenization are performed to ensure that the input and output representation space stay unchanged and (ii) ensure that the model can interpret these different length sequences. In the following, we discuss and outline two different ways that {\flexidit}s can be derived, based on whether the forward pass for the pre-trained model (we refer to this as the \emph{target} model) is preserved exactly or not. In both cases, we emphasize that any additional fine-tuning and new parameters are minimal. In all our experiments, training is stable, and no special tricks are necessary, while the total compute for fine-tuning is less than $5$\% of the original pre-training compute.

\subsection{Shared Parameters for all Sequences}
\label{sec:single_forward_pass}

When the original training data is available, it becomes possible to fine-tune the pre-trained model while preserving its performance, along with its existing abilities, biases, and potential safety features. We demonstrate that this approach enables processing images with varying patch sizes by introducing only minimal additional trainable parameters, as shown in Fig.~\ref{fig:methodology} (left). Below, we detail the specific components of the architecture that require adaptation.

\paragraph{Tokenization:}~We introduce a new embedding layer $\flexembedding$, for some underlying patch size $\patchsizeexp[\prime]$ with new weights $\flexweightembedding \in \R^{\patchsizeexp[\prime] * \patchsizeexp[\prime] * \inchannels \times \hiddensize}$ and biases $\flexbiasembedding \in \R^\hiddensize$. Then, when instantiating a~\flexidit~with a patch size $\patchsizecurrent$, we project these weights with a fixed projection matrix $\projectembedding \in \R^{\patchsizecurrent * \patchsizecurrent \times \patchsizeexp[\prime] * \patchsizeexp[\prime]}$ to the desired shape, and perform the $2$D convolution with kernel size and stride equal to $\patchsizecurrent \times \patchsizecurrent$. We use as $\projectembedding$ the pseudo-inverse of the bi-linear interpolation projection, as this leads to better norm preservation of the output~\citep{beyer2023flexivit}. We initialize the weights based on the pre-trained parameters and the pre-trained patch size as $\flexweightembedding = \projectembedding^\dagger \weightembedding$\footnote{Here $\dagger$ denotes the pseudo-inverse. All projection matrices $\project$ multiply each channel separately.} and $\flexbiasembedding = \biasembedding$, preserving the functional form of the model for the pre-trained patch size. When adding positional encodings, we identify for each patch its pixel coordinates in the original image~\citep{xie2023difffit, chen2023pixart}. We additionally introduce a patch size embedding for each of the patch sizes used by the model, which we add to every token in the sequence, and patch size dependent layer-normalization layers. These layers help with signal propagation~\citep{he2015delving, xiong2020layer, noci2022signal} by preserving the norms of the activations and let the model recover its expressivity~\citep{wimbauer2024cache, ioffe2015batch}.

\paragraph{De-tokenization:}~We similarly adapt the de-embedding layer $\deembedding$. For an underlying patch size $\patchsizeexp[\prime]$, we define a new layer $\flexdeembedding$ with weights $\flexweightdeembedding \in \R^{\hiddensize \times \outchannels * \patchsizeexp[\prime] * \patchsizeexp[\prime]}$ and biases $\flexbiasdeembedding \in \R^{\outchannels * \patchsizeexp[\prime] * \patchsizeexp[\prime]}$. Depending on the current patch size $\patchsizecurrent$ in the neural network evaluation, we project with a fixed projection matrix $\projectdeembedding \in \R^{\patchsizeexp[\prime] * \patchsizeexp[\prime] \times \patchsizecurrent * \patchsizecurrent}$ to $\flexweightdeembedding \projectdeembedding$ and $\flexbiasdeembedding \projectdeembedding$. Again, we use as $\projectdeembedding$ the pseudo-inverse of the bi-linear interpolation, now with flipped dimensions. We initialize the new parameters as $\flexweightdeembedding = \weightdeembedding \projectdeembedding^\dagger$ and $\flexbiasdeembedding = \biasdeembedding \projectdeembedding^\dagger$. 
In total, less than $0.005$ \% of auxiliary parameters are introduced to attain a~\flexidit~for the models tested in this case. 
\begin{figure*}[!t]
    \centering
    \includegraphics[width=\textwidth]{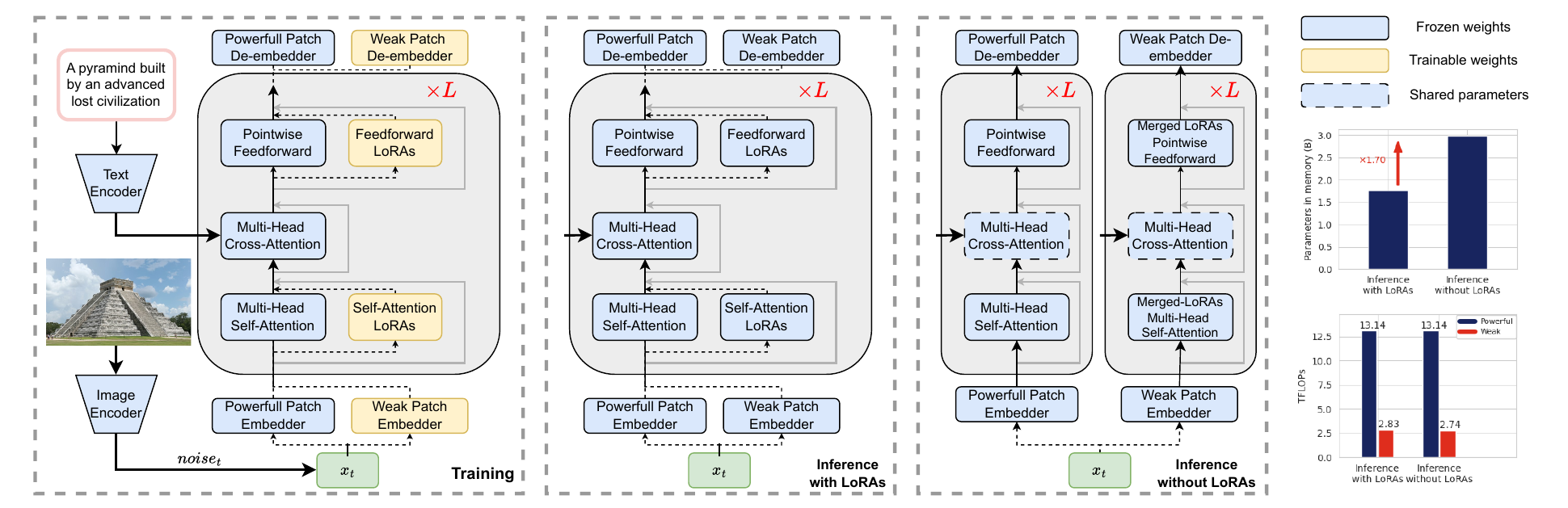}
    \caption{We preserve the functional form of the target model for the pre-trained patch size and add new trainable parameters (LoRAs) for each additional patch size we want to fine-tune the model to operate with. We showcase this for a text-to-image/video model that uses cross-attention for text conditioning. We find that freezing cross-attention layers without any additional LoRAs works the best. During inference, we can either keep the LoRAs unmerged (\textit{Inference with LoRAs}) leading to a slight FLOPs increase that depends on the LoRAs' dimensions, or create different copies of the model for each patch size, by merging the LoRAs (\textit{Inference without LoRAs}). The latter leads to additional memory requirements. FLOPs and parameter numbers on the right correspond to our flexible T2I~\emu~model.}%
    \label{fig:t2i_methodology}
\end{figure*}

\subsection{Different LoRAs for each Sequence}
\label{sec:different_forward_pass}

In practice, however, pre-training often requires extensive computational resources and may span multiple stages using various datasets, which may not always be accessible, even for models with open weights. In such cases, fine-tuning can have unintended effects, potentially diminishing model capabilities~\citep{ibrahim2024simplescalablestrategiescontinually} or compromising safety guarantees~\citep{qi2023finetuningalignedlanguagemodels}. For such situations, where it is essential to preserve the original forward pass of a pre-trained model, we demonstrate a method that enables fine-tuning across different patch sizes with minimal additional compute and a small supplementary dataset. Fig.~\ref{fig:t2i_methodology} illustrates our approach.

We perform similar changes to the model, but instead of training the DiT block parameters, we freeze the original weights and add trainable LoRAs~\citep{hu2021loralowrankadaptationlarge}. These LoRAs are specialized for each new patch size and activated only when our~\flexidit~is instantiated with that. There are no LoRAs for the patch size of the pre-trained model. For the tokenization and de-tokenization layers, we simply add new layers $\embedding, \deembedding$ for each new patch size. As before, we use a patch size embedding, but only for the new patch sizes. Note that functional preservation is imposed for the pre-trained patch size, i.e. inference using only the powerful model generates exactly the same samples. We fine-tune by using the predictions of the original model (powerful model) as labels to distill knowledge~\citep{hinton2015distilling} for the new patch sizes, i.e. we train to minimize:$$\mathbb{E}_{t, \x_t} \| \etheta(\x_{t - 1} | \x_{t}; \patchsizeunderscore[powerful]) - \etheta(\x_{t - 1} | \x_{t}; \patchsizeunderscore[weak]) \|_2.$$We find that this leads to improved performance, faster convergence~\citep{cho2019efficacy} and better alignment between predictions of different patch sizes, which can be important given potential discrepancies in the data used during pre-training and our fine-tuning. Here we use $\etheta$ to denote the model's prediction parametrizing the distribution $\ptheta$. Note that $\etheta(\x_{t - 1} | \x_{t}; \patchsizeunderscore[powerful])$ has no trainable parameters. 

During inference, we have two options: (i) keep the new LoRA parameters unmerged, which incurs a minor additional computational cost, or (ii) merge these weights into a copy of the original model parameters, with some additional memory cost. If we keep the LoRA parameters of dimension $\hiddensizelora$ unmerged, the computational complexity of the corresponding linear layer with input and output $(\hiddensizeinput, \hiddensizeoutput)$ will change from $\tokens \hiddensizeinput \hiddensizeoutput$ to $\tokens (\hiddensizeinput \hiddensizeoutput +  \hiddensizeinput \hiddensizelora + \hiddensizelora \hiddensizeoutput)$. If we merge LoRAs, there is no computational overhead, but the new merged parameters need to be kept in memory. Depending on the model requirements and the available resources, one can choose between the two options. We note that compute overhead by keeping LoRAs unmerged is minimal, see also Fig.~\ref{fig:t2i_methodology} (right). For all models tested, additional parameters in this case are less than 5\% of the original model parameters. Further details and ablations can be found in App.~\ref{app:implementation-details}. 

\subsection{Inference Scheduler}

At every denoising step, we need to decide how to instantiate our~\flexidit, i.e. which patch size to use. Following our intuition, we find that for early steps of the denoising process, weak and powerful models produce similar predictions, and thus using the weak model preserves quality while reducing computational complexity, as also seen in Fig.~\ref{fig:methodology} (right). We therefore propose an inference scheduler that, starting from random noise $\pnoise$, first denoises using a weak model for the first $\diffusionstepsweak$ steps and switches to the powerful model for the last $\diffusionstepspowerful = \diffusionsteps - \diffusionstepsweak$ steps. By adjusting the steps performed by the weak model $\diffusionstepsweak$, we can adjust the amount of compute that we are saving. In practice, unless otherwise mentioned, we fine-tune models to process images with one additional patch size. We choose this patch size, corresponding to the weak model, as $2 \times$ larger than the one corresponding to the powerful model. Then, the sequence length corresponding to tokenizing with the new patch size is $4 \times$ smaller, and thus compute required for the powerful model is $> 4 \times$ compared to the weak model. 




\begin{figure*}[!h]
    \centering
    \floatsetup{heightadjust=all, valign=c}
    \begin{floatrow}
        {{\includegraphics[width=0.33\textwidth]{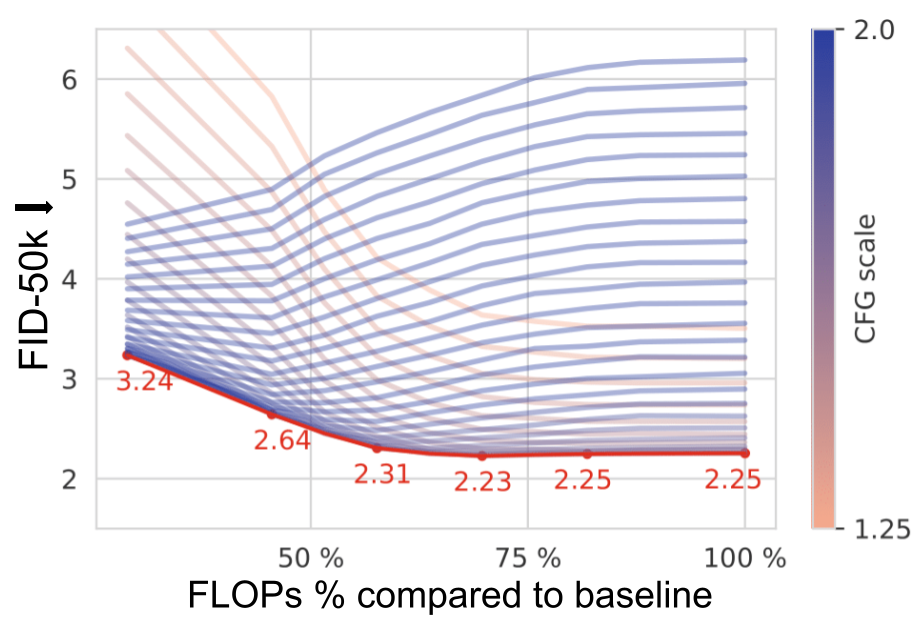} }}%
        \hfill
        {{\includegraphics[width=0.33\textwidth]{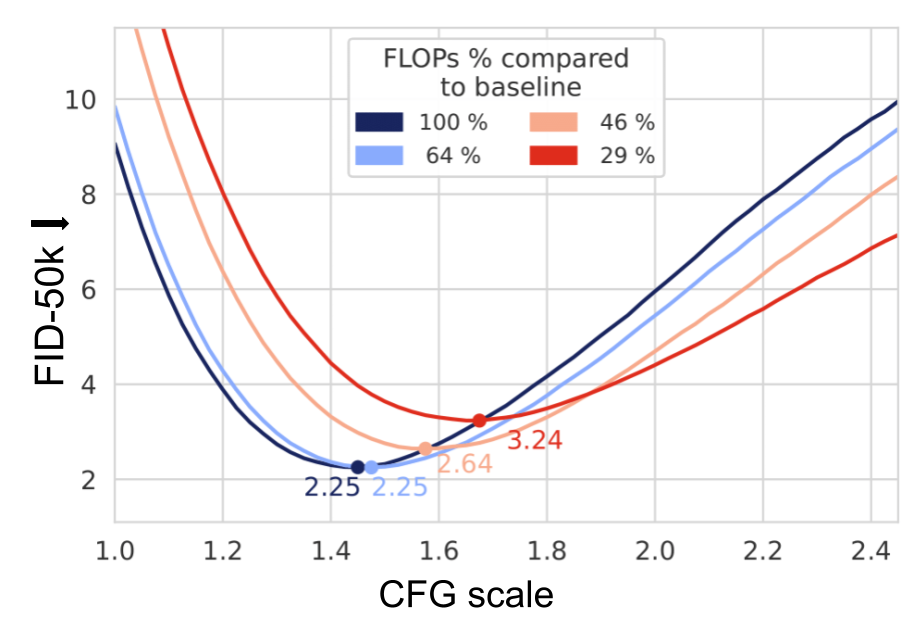} }}%
        \hfill
        {{\includegraphics[width=0.32\textwidth]{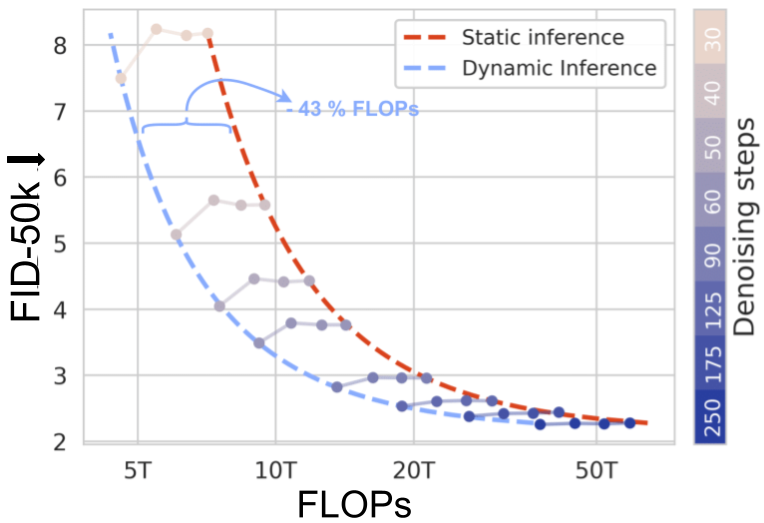} }}%
    \end{floatrow}
    \caption{\textbf{Left:} As the weak model is used more extensively during generation, compute benefits increase, but at the cost of some performance degradation. \textbf{Middle:} The optimal CFG scale varies depending on the extent to which the weak model is used. Each line corresponds to an inference scheduler that applies the weak model for a different proportion of denoising steps. \textbf{Right:} Benefits from our inference scheduler are orthogonal to performing a smaller overall number of diffusion steps. We plot FID for different overall number of steps $\diffusionsteps$ and different number of weak steps $\diffusionstepsweak$, using in every case the DDPM scheduler.}%
    \label{fig:xl-experiments}
\end{figure*}

\subsection{Generation Guidance}

For conditional generation, classifier-free guidance (CFG) is typically employed~\citep{ho2022classifier, esser2024scaling} to enhance sample quality. This entails performing two neural function evaluations (NFEs) (or one NFE with twice the batch size) to compute predictions with and without the conditioning $\condition$, i.e. $\etheta(\x_{t - 1} | \x_{t}, \condition)$ and $\etheta(\x_{t - 1} | \x_{t}, \emptyset)$. Sampling can then take place as $\etheta(\x_{t - 1} | \x_{t}, \emptyset) + \guidancescale (\etheta(\x_{t - 1} | \x_{t}, \condition) - \etheta(\x_{t - 1} | \x_{t}, \emptyset))$, where $\guidancescale$ is the guidance scale. Recent work~\citep{karras2024guiding} has shown that using a smaller or less well-trained version of the model rather than an unconditional model can lead to better guidance signal~\citep{ahn2024selfrectifyingdiffusionsamplingperturbedattention, sadat2024trainingproblemrethinkingclassifierfree}. We adapt these findings in our setting, leading to better generation quality \emph{without the need to train or deploy additional models}. For each denoising step, given a patch size used for the conditional $\patchsizecondition$ and a patch size used for guidance $\patchsizeuncondition$, we compute:
\begin{align*}
    \begin{cases}
      \etheta(\x_{t - 1} | \x_{t}, \emptyset; \patchsizeuncondition) + \guidancescale (\etheta(\x_{t - 1} | \x_{t},\condition; \patchsizecondition) - \\ \etheta(\x_{t - 1} | \x_{t}, \emptyset; \patchsizeuncondition)), \quad \text{if}\ \patchsizecondition = \patchsizeuncondition \\
      \etheta(\x_{t - 1} | \x_{t}, \condition; \patchsizeuncondition) + \guidancescale (\etheta(\x_{t - 1} | \x_{t}, \condition; \patchsizecondition) - \\ \etheta(\x_{t - 1} | \x_{t}, \condition; \patchsizeuncondition)), \quad \text{if}\ \patchsizecondition < \patchsizeuncondition
    \end{cases}
\end{align*}
In this setup, we use the powerful model for the conditional prediction and leverage the weak model's output as guidance. Unlike traditional approaches, our method applies guidance based on the \emph{conditional} prediction from the weak model. Our guidance scheme requires performing inference using both the weak (for the unconditional) and the powerful (for the conditional) model, for some denoising steps. We show in Fig.~\ref{fig:packing} (appendix) how this can be efficiently implemented, making use of packing~\citep{dehghani2024patch}. Depending on the guidance signal used, optimum generation quality can vary with respect to the guidance scale $\guidancescale$. This will become more apparent in the following experiments.

%% file: sec/4_1_class_conditioned.tex
\section{Experiments}

For clarity, we present efficiency gains with respect to FLOPs and point to Section~\ref{sec:latency} for a detailed analysis of the relationship between FLOPs and latency.

\subsection{Class-Conditioned Image Generation}
\label{sec:class_conditioned_experiments}

We fine-tune models (\textit{DiT-XL/2}) on~\ImageNet, using the same setup as in~\citep{peebles2023scalablediffusionmodelstransformers}. Since training data is publicly available, we fine-tune pre-trained models using the same parameters for all sequences, without the use of LoRAs, as described in Sec.~\ref{sec:single_forward_pass}. During training, we randomly noise images according to~\cref{eq:noise_images}, and learn to denoise using one of the available patch sizes. We primarily report FID and point to App.~\ref{sec:additional-experiments} for more experiments and different metrics. In practice, we use a pre-trained model with a patch size of $2$ (\emph{powerful}), that we fine-tune to also process images with a patch size of $4$ (\emph{weak}). Since we are fine-tuning the powerful model, we can also ``teach'' it how to correct specific mistakes made by the weak model, accumulated in the backward process during the first $\diffusionstepsweak$ steps. We provide more details in App.~\ref{sec:exposure_bias} on how to reduce this exposure bias~\citep{ning2023elucidating, li2023alleviating}. Unless otherwise mentioned, reported metrics are computed by generating images at resolution $256 \times 256$, using $250$ steps of the DDPM scheduler~\citep{ho2020denoising, peebles2023scalablediffusionmodelstransformers}.

\paragraph{Compute gains.}~We generate images with our~\flexidit~and the proposed inference scheduler,  varying the amount of compute by adjusting the number of initial denoising steps $\diffusionstepsweak$ performed with the weak model. For each level of compute, we report the FID of the generated images in Fig.~\ref{fig:xl-experiments} (left and middle). In general, performing a few steps with the weak model (60-100\% of baseline compute) leads to \emph{no drop} in performance. Saving even more compute is possible, albeit at the cost of a minor drop in the quality of the generated images. When using only the powerful mode of our~\flexidit, we get the same performance --- $\textit{FID-}50\textit{k} = 2.25$ --- as the pre-trained \textit{DiT-XL/2} model ---  $\textit{FID-}50\textit{k} = 2.27$. Thus, \textit{fine-tuning for more patch sizes does not reduce the capacity of the model with respect to the pre-trained one}. We also show that other inference schedulers, such as starting with the powerful model and switching to the weak model, lead to worse results (appendix Fig.~\ref{fig:opposite-scheduler}), validating our intuition.
\begin{figure*}[!h]
    \begin{minipage}{\textwidth}
    \begin{minipage}[b]{0.29\textwidth}
        \centering
        \includegraphics[width=1\textwidth]{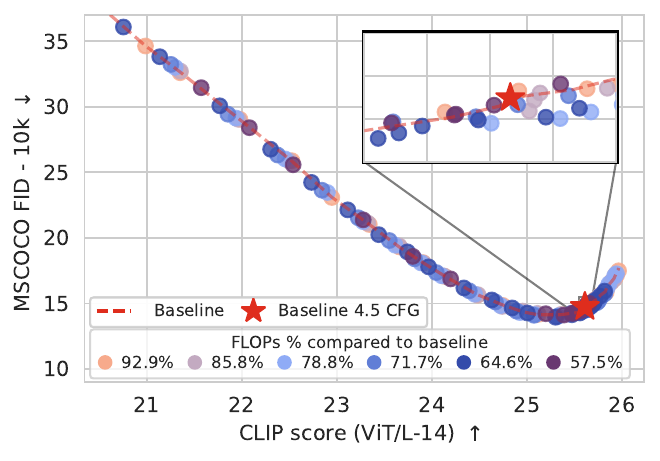}
    \end{minipage}
    \hfill
    \begin{minipage}[b]{0.38\textwidth}
        {\scriptsize
        \begin{tabular}{cccc}
        \toprule  
        Model (compute \%) & \textit{FID} $\downarrow$ & \textit{CLIP} $\uparrow$ & \textit{VQAScore} $\uparrow$ \\
        \midrule
        \pixart~($100$ \%) & $14.75$ & $25.60$ & $63.29$ \\  
        \pixart~( $86$ \%) & $14.72$ & $25.62$  & $63.30$ \\  
        \pixart~( $72$ \%) & $14.77$ & \underline{$25.63$} & \underline{$63.40$} \\  
        \pixart~( $58$ \%) & \underline{$14.71$} & $25.58$ & $63.26$ \\  
        \midrule
        \emu~($100$ \%) & \underline{$26.00$} & $26.05$ & $70.17$ \\  
        \emu~( $84$ \%) & $25.96$ & $26.06$ & $70.19$ \\  
        \emu~( $69$ \%) & \underline{$26.00$} & \underline{$26.08$} & \underline{$70.37$} \\  
        \emu~( $53$ \%) & $26.10$ & $26.07$ & $70.09$\\  
        \bottomrule
        \end{tabular}
        \par\vspace{8pt}
        }
    \end{minipage}
    \hfill
    \begin{minipage}[b]{0.32\textwidth}
        {\scriptsize
        \begin{tabular}{|cc|ccc|}
        \toprule
        \multicolumn{2}{|c|}{Steps $_\textit{(Powerful/Weak)}$} & \multicolumn{3}{c|}{Votes (in \%)} \\
        Ours & Baseline  & Win & Tie & Lose \\
        \midrule
        50 $_{(40, 10)}$ & 42 $_{(42, 0)}$ & $33.5$ & $42.5$ & $24.0$ \\
        50 $_{(30, 20)}$ & 34 $_{(34, 0)}$ & $35.5$ & $41.5$ & $23.0$ \\
        50 $_{(20, 30)}$ & 26 $_{(26, 0)}$ & $35.5$ & $44.0$ & $20.5$ \\
        50 $_{(10, 40)}$ & 18 $_{(18, 0)}$ & $43.0$ & $41.5$ & $15.5$ \\
        \midrule
        \multicolumn{5}{|p{5.00cm}|}{\tiny{We asked $8$ annotators to express preferences between pairs of images for a given prompt. We generate images for $200$ prompts and collect $3$ different votes per each pair of images, collectively $2400$ votes.}} \\
        \bottomrule
        \end{tabular}
        \par\vspace{7pt}
        }
    \end{minipage}
    \end{minipage}
    \vspace{-3mm}
    \caption{\textbf{Left:} We plot FID vs CLIP score for images generated with different CFG scales using the~\pixart~model (we refer to the appendix for results regarding our~\emu~model). The red line represents images generated with varying CFG scales using only the (powerful) target model. By employing our dynamic scheduler, we can match image quality in terms of both FID and text alignment while significantly reducing compute requirements. \textbf{Middle:} Our flexible models can match the baseline (for a fixed pre-defined guidance scale $\guidancescale$) across benchmarks, with significantly less compute. \textbf{Right:} Human study results show votes indicating a win, tie, or loss for our method compared to a baseline, which corresponds to running the pre-trained model (only the powerful model) for fewer steps. Comparisons are between~\emu~inference modes that require approximately equal FLOPs and time.}%
    \label{fig:t2i-experiments}
\end{figure*}

\paragraph{Relation between $\diffusionsteps$ and $\diffusionstepsweak$.}~Naturally, the question arises whether doing fewer overall diffusion steps $\diffusionsteps$ leads to the same efficiency improvements compared to performing more steps with the weak model. After applying the DDPM scheduler for a different number of overall diffusion steps $\diffusionsteps$, we plot in Fig.~\ref{fig:xl-experiments} (right) the efficiency gains across them. Results indicate that gains from performing weak steps are orthogonal to performing fewer overall diffusion steps. In other words, \textit{more steps are required to achieve better image fidelity, but performing some of the initial steps with our weak model is sufficient to achieve the targeted fidelity}. In App.~\ref{sec:intuition}, we provide further insights on how the predictions of the weak model closely align with the ones of the powerful model. This similarity supports parameter sharing, which not only reduces resource requirements during inference but also enables faster convergence during fine-tuning.


%% file: sec/4_2_text_conditioned.tex
\vspace{2mm}
\subsection{Text-conditioned Image Generation}
\label{sec:text-condition}

The universality of Transformers and the holistic view of different input sources as sequences of tokens implies that the generalization of the architecture to more modalities and a variety of different inputs is straightforward. This is also the case for DiTs, which have been already extended to accommodate text conditioning~\citep{chen2023pixart, betker2023improving}, video generation~\citep{ma2024latte} and speech synthesis~\citep{liu2024autoregressive}. We showcase how our framework can be applied out of the box to state-of-the-art text-to-image (T2I) model architectures. T2I DiTs have the same architecture as class-conditioned DiTs, with the exception that conditioning is imposed via cross-attention.

We fine-tune T2I models, by introducing new parameters in the form of LoRAs. Specifically, we use a DiT following PIXART~\citep{chen2023pixart} --- we refer to this as~\pixart~--- generating $256 \times 256$ images and a 1.7B DiT based on EMU~\citep{dai2023emu} --- we refer to this as~\emu~--- generating $1024 \times 1024$ images. Implementation details are provided in App.~\ref{sec:details_text_to_image}, in short, we fine-tune both low and high-resolution target models with a pre-trained patch size of $2$, to also support a patch size of $4$. For~\pixart, we follow the inference scheduler protocol in~\citep{chen2023pixart}, and use the DDPM solver for $100$ steps. We point to the App.~\ref{sec:additional-experiments} for results with different solvers and number of steps (namely $20$ steps of DPM, $25$ steps of SA-solver, and a varying number of steps of the DDPM solver) showing that performance benefits are orthogonal to these choices. As before, we vary the percentage of denoising steps with the weak model for either the conditional or the guidance predictions.

\begin{figure*}[!h]
    \begin{minipage}{\textwidth}
        \begin{minipage}[b]{0.65\textwidth}
        \centering
        \includegraphics[width=1\linewidth]{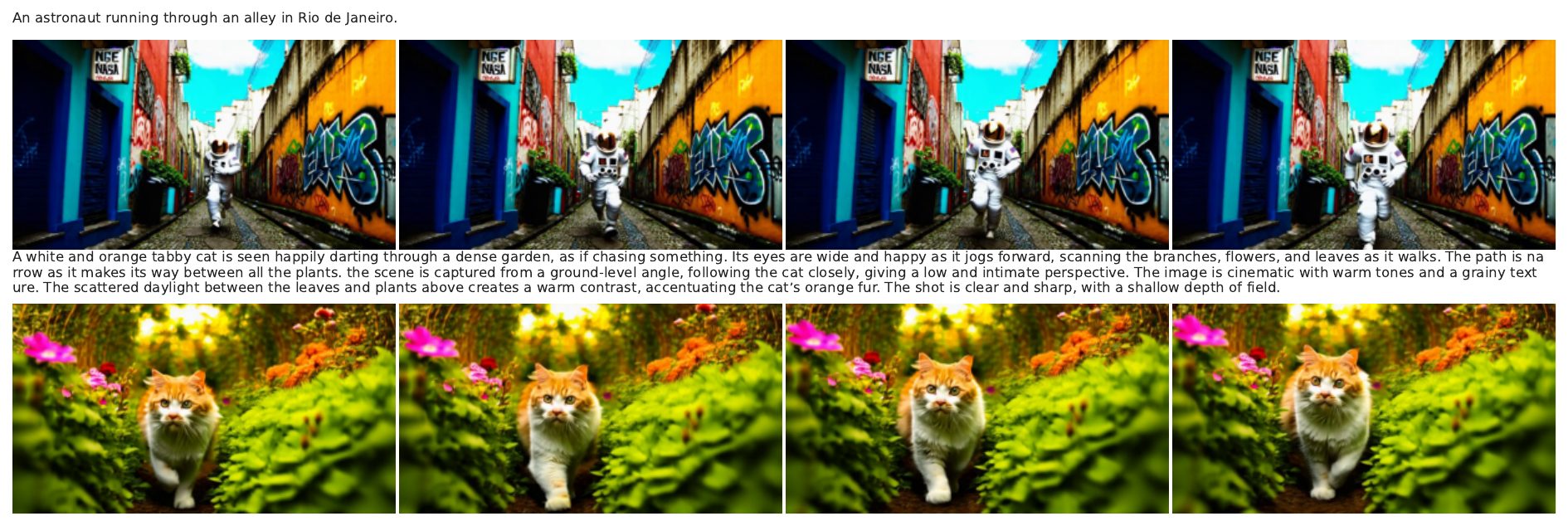}
        \end{minipage}
        \begin{minipage}[b]{0.33\textwidth}
        \includegraphics[width=1\linewidth]{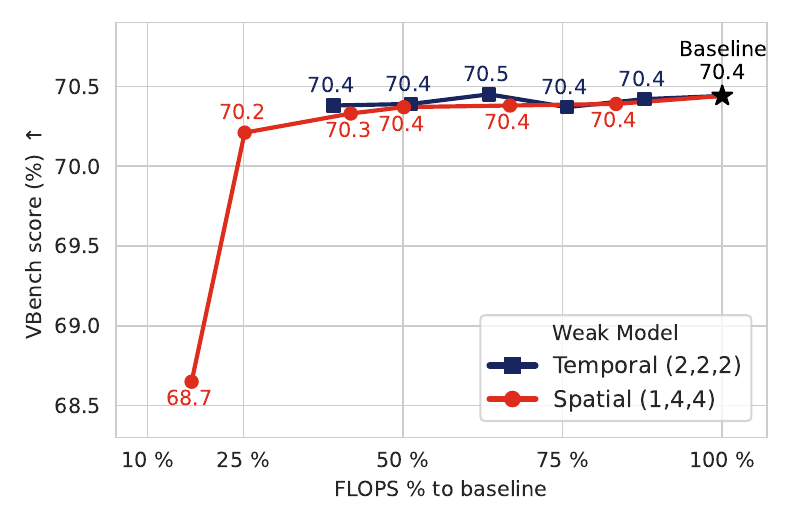}
        \end{minipage}
    \end{minipage}
    \caption{\textbf{Left:} Samples from our flexible~\moviegen~model using $25.2$\% of compute compared to the pre-trained baseline. \textbf{Right:} We perform a varying number of steps of the denoising process with a weak model, using either our spatial or our temporal weak model. Both weak models lead to significant compute savings with \emph{little to no} degradation in performance.}
    \label{fig:moviegen-experiments}
\end{figure*}

\paragraph{CLIP-vs-FID.}~We report FID and CLIP score alignment for captions from the \textit{MS COCO} dataset with varying CFG scale values in Fig.~\ref{fig:t2i-experiments} (left). As before, we notice that lower compute versions of our model require a higher CFG scale to generate images that match similar FID vs CLIP score values of the baseline (full compute) model. We show that for a fixed CFG scale of the baseline model --- we chose $\guidancescale = 4.5$ for the~\pixart~model as proposed in~\cite{chen2023pixart} and $\guidancescale = 6.0$ for our~\emu~model --- we can match the attained performance for less than 60\% of the original compute. Detailed scores are given in Fig.~\ref{fig:t2i-experiments} (middle). 
\paragraph{Additional benchmarks.}~We perform additional evaluations on the final generated images. We follow~\citep{lin2024evaluating} and present text alignment for visual question-answering (VQA). We generate images based on the captions of \textit{DrawBench}, \textit{TIFA160}, \textit{Pick-a-Pic}, \textit{Winoground} and report average VQA scores (we point to App.~\ref{sec:additional-experiments} for detailed results). We compare the baseline model (same CFG scales as before), against dynamic inference with our flexible models. Consistent with our previous findings, results indicate that performing a few denoising steps with our weak model generally generates high-quality images. We can again save up to $40$\% of compute without a performance drop. For high-resolution images generated from our~\emu~model, we also embark on a human study, as shown in Fig.~\ref{fig:t2i-experiments} (right). There we validate that human annotators prefer images from our dynamic scheduler instead of using similar compute, but this time uniformly allocated.




%% file: sec/4_3_video_generation.tex
\subsection{Text-Conditioned Video Generation}
\label{sec:video_generation}
Finally, we also explore other modalities, such as video generation. Text-to-video (T2V) generation typically follows the same setup as T2I~\citep{polyak2024movie, ma2024latte}. Given a video of dimensions $\R^{\fdim \times \hdim \times \wdim \times \inchannels}$, where $\fdim$ is the temporal component corresponding to the number of frames, and a chosen patch size $(\patchsizef, \patchsizeh, \patchsizew)$ for each of the dimensions, the input (latent) video is cropped into non-overlapping spatial-temporal patches of dimensions $\R^{\patchsizef \times \patchsizeh \times \patchsizew \times \inchannels}$. Patches are then embedded using now a $3$D convolutional layer into tokens, subsequently transformed by the DiT blocks where every token attends to every other token in the sequence, and finally projected back to patches with a linear layer. 

We perform the same changes as for our T2I models in Sec.~\ref{sec:different_forward_pass}, adding LoRAs for new patch sizes. We initialize tokenization and de-tokenization layers as before, and when the temporal patch size $\patchsizef$ is increased, we duplicate weights along the temporal dimension. We fine-tune a 4.9B video model as the one in~\citep{polyak2024movie} --- referred to as~\moviegen~--- and train with distillation as before. We generate videos of resolution $(\fdim \times \hdim \times \wdim) = (256, 384, 704)$ using $250$ steps of the DDPM scheduler as in~\citep{polyak2024movie}. Diffusion is performed in a latent space that down-samples each dimension $\fdim, \hdim, \wdim$ by $8$. More dimensions of the latent space now offer more possibilities in terms of determining the characteristics of our weak model. For our experiments, we fine-tune a pre-trained model with patch size $(\patchsizef, \patchsizeh, \patchsizew) = (1, 2, 2)$ to also support a patch size $(2, 2, 2)$, we denote this weak model as `temporal', and $(1, 4, 4)$, denoted as `spatial'. We then use either one of these modes as a weak model during inference. Both modes could also be used iteratively for the generation of a single sample, which we leave for future work.

By adjusting the number of steps performed with our weak model, we can again control the overall compute required per generated sample. We use \textit{VBench}~\citep{huang2024vbench} to evaluate the quality of the generated videos and report results in Fig.~\ref{fig:moviegen-experiments}. In this case, we can save up to $75$\% of compute without a significant drop in performance. Recent training-free methods~\citep{liu2024faster, zhao2024real, kahatapitiya2024adaptive}, while appealing due to their lack of training requirements, achieve significantly lower compute savings before performance begins to degrade noticeably. This demonstrates that \textit{allowing the model to learn optimized compute allocation is more effective than relying on predefined rules for inference efficiency}. Additional comparisons to previous work are provided in the appendix.




%% file: sec/4_4_latency.tex
\subsection{FLOPs vs Latency}
\label{sec:latency}

\begin{figure}
    \centering
    \includegraphics[width=\linewidth]{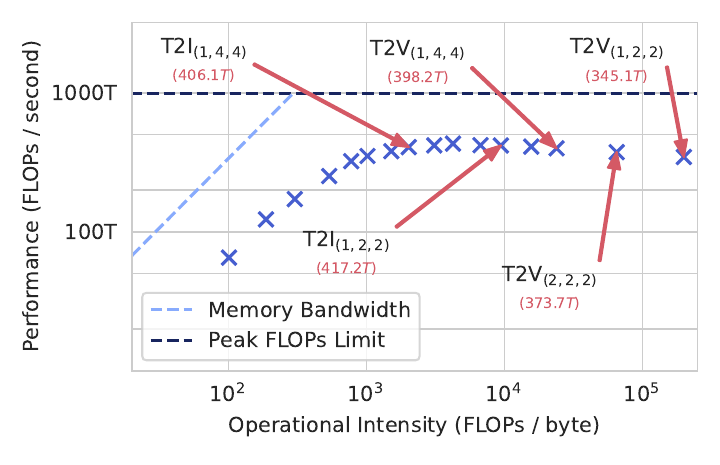}
    \caption{GPU utilization for one denoising step, when propagating sequences with different overall number of tokens, corresponding to different patch sizes $(\patchsizef, \patchsizeh, \patchsizew)$. Our T2I model has no temporal dimension, but we overload notation and set $\patchsizef = 1$. For simplicity, we use a DiT of similar configuration (width and depth) for both the T2I and T2V reported in this plot numbers, but results generalize across model shapes.}
    \label{fig:latency}
\end{figure}

High-resolution image/video generation is predominantly compute-bound. To verify, we propagate sequences of different lengths through a fixed size DiT and measure performance --- FLOPs and latency --- on a \textit{NVIDIA H100 SXM5} with a batch size of $2$, simulating inference with CFG\footnote{In all cases, we compile using \textit{torch.compile} with \textit{fullgraph=True} and \textit{mode = 'reduce-overhead'}.}. In Fig.~\ref{fig:latency} we show that the weak~\emu~and~\moviegen~models are also compute-bound, for the setup (generated image/video resolution) that we presented in the paper. Indeed, for T2V, FLOPs utilization is higher for our weak models, due to inefficiencies of the self-attention operation for large sequence lengths when using the powerful model. This effect can be expected to be even more predominant for multi-GPU inference. Consequently, \emph{latency benefits are even higher than FLOPs benefits} presented so far.

%% file: sec/5_coclusion.tex
\section{Conclusion}

We have demonstrated how regular DiTs can be converted into \emph{flexible} ones, that can process samples with different patch sizes after minimal fine-tuning. Adjusting the compute for some denoising steps in the diffusion process readily allows accelerating inference without compromising. Notably, the efficiency benefits of our approach are independent of the chosen solver or the number of denoising steps. We have displayed how our approach is generic and can be straightforwardly applied to class-conditioned image generation, low and high-resolution text-conditioned image generation, and text-conditioned video generation. Looking ahead, we anticipate further applications of our flexible DiT framework across various modalities, such as audio and $3$D modeling. As computational resources become increasingly in demand, developing efficient and adaptable models like ours will be crucial for enabling generative capabilities that are of high-quality, but also more scalable.

%% file: sec/appendix.tex
\section{Detailed Related Work}

Efficiency in Vision Transformers~\citep{dosovitskiy2020image} falls primarily into two broad categories: reducing the computation per token or the number of tokens overall. 

\begin{figure}[!h]
    \centering
    \includegraphics[width=0.37\linewidth]{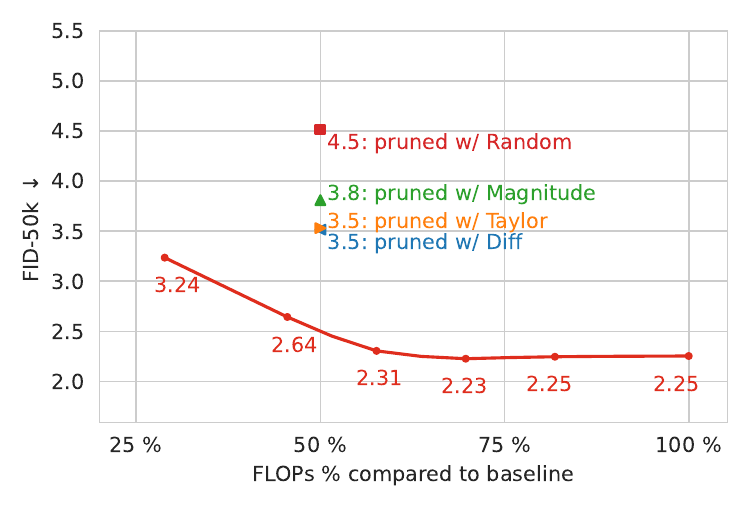}
    \caption{We compare our dynamic scheduler with more baselines.}
    \label{fig:baseline}
\end{figure}

\paragraph{Reducing the amount of computation per token.}~Reducing the amount of computation per token can be achieved by reducing the model size when training via distillation~\citep{beyer2022knowledge} or pruning the network after training~\citep{zhu2021vision, imfeld2023transformer}. These methods again though typically work with a static inference protocol. As in~\citep{zhao2024dynamic}, we compare in Fig.~\ref{fig:baseline} our class-conditioned~\ImageNet~\flexidit~model, against popular based pruning techniques, based on Diff pruning~\citep{fang2023structuralpruningdiffusionmodels}, Taylor, magnitude or random pruning~\citep{imfeldtransformer}.
As one can see, our dynamic scheduler outperforms these baselines. \textit{We note that our method can also be applied in conjunction with pruning techniques to achieve even higher efficiency gains, offering potential orthogonal benefits.} In concurrent work,~\citep{zhao2024dynamic} propose to adjust, in a dynamic way, the model during inference in different steps. In our work, we do not train separate models for each target FLOPs ratio like they do, meaning that we can train a single model and decide how many FLOPs we want to invest during inference. This makes our approach more versatile. Additionally, our training is very stable, and no specific training tricks are required to converge successfully. That is how we were able to extend experiments to high-resolution image and video generation, achieving significantly better speed-ups than the ones they reported. We also outperform their results when the compute budget is very small. Nonetheless, this work could inspire adaptive per-sample schedulers, that could open new future directions.

Given the large computational requirements of the attention operation, many methods nowadays focus on that to reduce the overhead imposed. These methods commonly use some form of hierarchical attention~\citep{liu2021swin, hatamizadeh2023fastervit}, skip (usually the first) attention layers altogether~\citep{xiao2021early}, or reduce the number of the attended keys~\citep{chen2024pixart, yuan2024ditfastattn, fan2021multiscale}, by commonly aggregating keys in a spatial neighborhood or applying some form of windowed attention.

\paragraph{Reducing number of tokens.}~Our method primarily falls within the second category of reducing the overall number of tokens. Previous work here, typically relied on filtering~\citep{rao2021dynamicvit, liu2022adaptive, wu2023ppt}, merging~\citep{bolya2022token, lu2023content, huang2023vision} or dropping~\citep{anagnostidis2023dynamic}. Although merging works well for applications that eventually lead to some pooling operations (like classification tasks or for the task of creating an image-specific embedding), it works significantly less well for applications that require dense (token-level) predictions, where some un-merging operation has to be defined. In other concurrent work,~\citep{wang2024qihoo} reduces the number of representative tokens to calculate the attention over. Our approach resembles most~\citep{beyer2023flexivit}, where vision Transformers are trained to handle inputs with varying patch resolution. By applying less compute for some steps, we can reduce computational complexity significantly, without a drop in performance. 

\paragraph{Image generation.}~In the context of image generation, diffusion has been largely established as the method for attaining state-of-the-art results. There have been previous works that try to take advantage of potential correlations between successive denoising step predictions~\citep{shi2024resmaster}, by either caching intermediate results in the activations~\citep{ma2024deepcache, wimbauer2024cache}, or in the attention~\citep{zhao2024real}. Caching has the advantage of a training-free method. Nonetheless, potential benefits are lower. Similar to our work, ~\citep{balaji2022ediff} use different experts for different denoising steps. Instead of using different experts that require separate training and separate deployment, we show how a single model can be easily formed into a flexible one that can be instantiated in different modes, with each of its modes corresponding essentially to a different expert. Similar in-spirit approaches have been proposed that rely on the smaller compute requirements for lower resolution image generation~\citep{kim2024pagoda, zhang2023i2vgen}. ~\citep{jing2022subspace} also adapt the computation per step, by projecting into smaller subspaces. We instead, keep the dimension of the latent space and the characteristics of it the same across diffusion steps. Orthogonal gains to our approach are also possible through methods such as guidance distillation~\citep{meng2023distillation, kohler2024imagine} and consistency models~\citep{song2023consistency}. Our approach is also largely agnostic to the diffusion process and can be applied out of the box for flow matching methods~\citep{lipman2022flow}. We point the interested reader to~\citep{ma2024efficient} for a survey for further efficiency in diffusion models. Finally, compared to other established techniques~\citep{hatamizadeh2025diffit, liu2024alleviating} we do not fundamentally change the architecture, which allows us to apply our framework effortlessly for numerous pre-trained models across different modalities.

\paragraph{Video generation.}~Our approach can be easily extended for video generation, and in principle for the generation of any modality where some inductive bias (spatial, temporal, etc) is employed in the diffusion (latent) space. In video generation, typically, latent video tokens are processed in parallel~\citep{liu2024sora, blattmann2023stable, polyak2024movie}. Training-free methods~\citep{liu2024faster, zhao2024real, kahatapitiya2024adaptive} have been proposed in this case to accelerate video generation. Benefits with training-free methods are nonetheless minimal before performance degradation kicks in (see Table 1 in~\citep{kahatapitiya2024adaptive}, where one can typically save less than 30\%).

An interesting direction for future work involves adapting the inference scheduler, i.e. what patch size we are using for each denoising step, based on the requirements of each sample. It is natural to assume that when generating more static videos, increasing the temporal patch size, and thus decreasing the amount of compute along the temporal dimension, will result in smaller drops in performance. The same holds for the spatial patch sizes when generating images or videos that require less high-frequency details. 




\section{Additional Experiments and Details}
\label{sec:additional-experiments}

We provide additional experiments, complementary to the main text.

\subsection{Exposure Bias and Accumulation of Error}
\label{sec:exposure_bias}

\begin{figure}[!h]
    \centering
    \qquad
    {{\includegraphics[width=5.5cm]{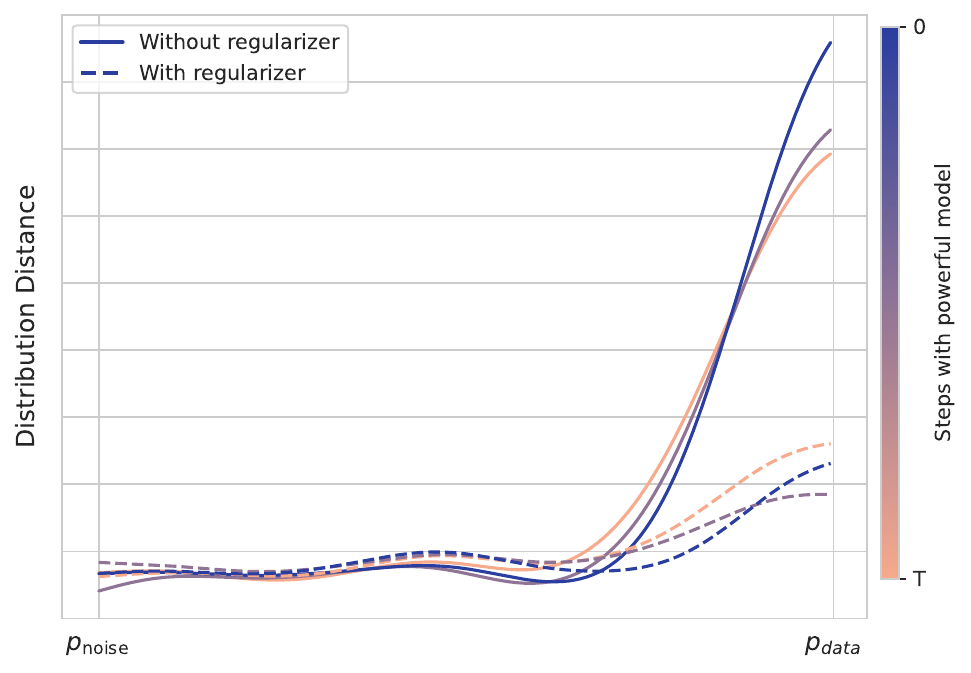} }}%
    \qquad \qquad
    {{\includegraphics[width=8.0cm]{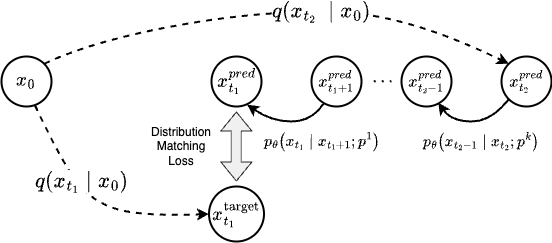} }}%
    \qquad
    \caption{\textbf{Left:} We use maximum mean discrepancy to estimate the distribution mismatch between $\ptheta(\x_{t} | \x_{T:t + 1})$ and $\q (\x_{t} | \x_0)$. \textbf{Right:} The proposed bootstrapped loss. During training, we perform a few denoising steps with a weak model followed by a few denoising steps with a powerful model (reminiscent of the scheduler during inference) and apply a distribution matching loss on the resulting samples.}%
    \label{fig:mmd}
\end{figure}

Inference with diffusion suffers from \emph{exposure bias}, due to the discrepancy of the input distribution during training and inference~\citep{li2023alleviating, li2023error, ning2023elucidating, daras2024consistent}. Briefly, models are trained to denoise images sampled from the $\q (\x_{t} | \x_0)$ distribution~\citep{ho2020denoising}. Inference on the other hand is characterized by repeated model evaluations $\ptheta(\x_{t} | \x_{t + 1})$ and any distribution mismatch between $\ptheta(\x_{t} | \x_{T:t + 1})$ and $\q (\x_{t} | \x_0)$ accumulates, as also shown in Fig.~\ref{fig:mmd} (left). The error at each iteration depends on the model, with a perfect model resulting in $0$ error and thus no error accumulated. In our case, the accumulation of error is exacerbated by the characteristics of our model, where weak models could lead to higher, but also specific in nature, kinds of errors. Training with the standard denoising objective, where real samples are randomly noised for some $t$, does not make the more powerful model aware of the nature of the mistakes made by the weak model, rendering it unable to potentially correct them. We propose to mitigate this issue by introducing a bootstrapped distribution matching loss~\citep{tesauro1995temporal}, as illustrated in Fig.~\ref{fig:mmd} (right). The loss is applied in a patch size-dependent manner, according to the desired inference protocol (from weak to powerful model calls during inference). 

Given natural images $\x_0, \tilde{\x}_0 \sim \q(\x_0)$, we sample two time points $t_1 > t_2$ and corrupt the images with noise $\x_{t_1}^\text{target} \sim \q(\x_{t_1} | \x_0), \tilde{\x}_{t_2}^\text{pred} \sim \q(\tilde{\x}_{t_2} | \tilde{\x_0})$. We then apply a chain of denoising steps $\etheta(\tilde{\x}_{t - 1}^\text{pred} | \tilde{\x}_{t}^\text{pred}; \patchsize)$ for $t \in (t_1, t_2]$ and a patch size $\patchsize$. Ultimately, we wish for the distributions of $\x_{t_1}^\text{target}$ and $\tilde{\x}_{t_1}^\text{pred}$ to match, for which we employ the maximum mean discrepancy (MMD)~\citep{gretton2012kernel}. To let the powerful model learn and potentially correct mistakes of the weak model and to simulate how our inference patch size scheduler works, we perform the first of these denoising steps with the weak model, followed by denoising steps with the powerful model. Given a set of patch sizes $\{ \patchsizeexp[i]\}_{i=1}^{k}$ where $\patchsizeexp[1] < \patchsizeexp[2] < \dots < \patchsizeexp[k]$ and a number of denoising steps to perform with each $\denoisestepexp[1], \denoisestepexp[2], \dots, \denoisestepexp[k]$, where $\sum_{j=1}^k \denoisestepexp[j] = t_2 - t_1$, we denoise with a given patch size $\patchsizeexp[i]$ for all $t$'s in $( t_1 + \sum_{j = i + 1}^k \denoisestepexp[j], t_1 + \sum_{j = i}^k \denoisestepexp[j]]$. An illustration of this process can also be seen in Fig.~\ref{fig:mmd} (right). When sampling a time step $t_1$, we bias our sampling similar to~\citep{sauer2024fast}. The proposed distribution matching loss, inspired by the notion of consistency~\citep{song2023consistency, song2023improved}, provides a principled way to correct the errors accumulated during inference. We note that we are optimizing a simple distribution matching loss, instead of over-optimizing according to desired downstream metrics (namely FID), thus not violating Goodhart's law\footnote{Goodhart's law states that: `When a measure becomes a target, it ceases to be a good measure'.}. Different distribution matching losses (including discriminator-based losses) can also be used. Note that we only correct this exposure bias for the class-conditioned image generation experiments, as this is the only case where we fine-tune the powerful pre-trained model.

\subsection{Inference with Packing}

\begin{figure}[!ht]
    \centering
    \includegraphics[width=\textwidth]{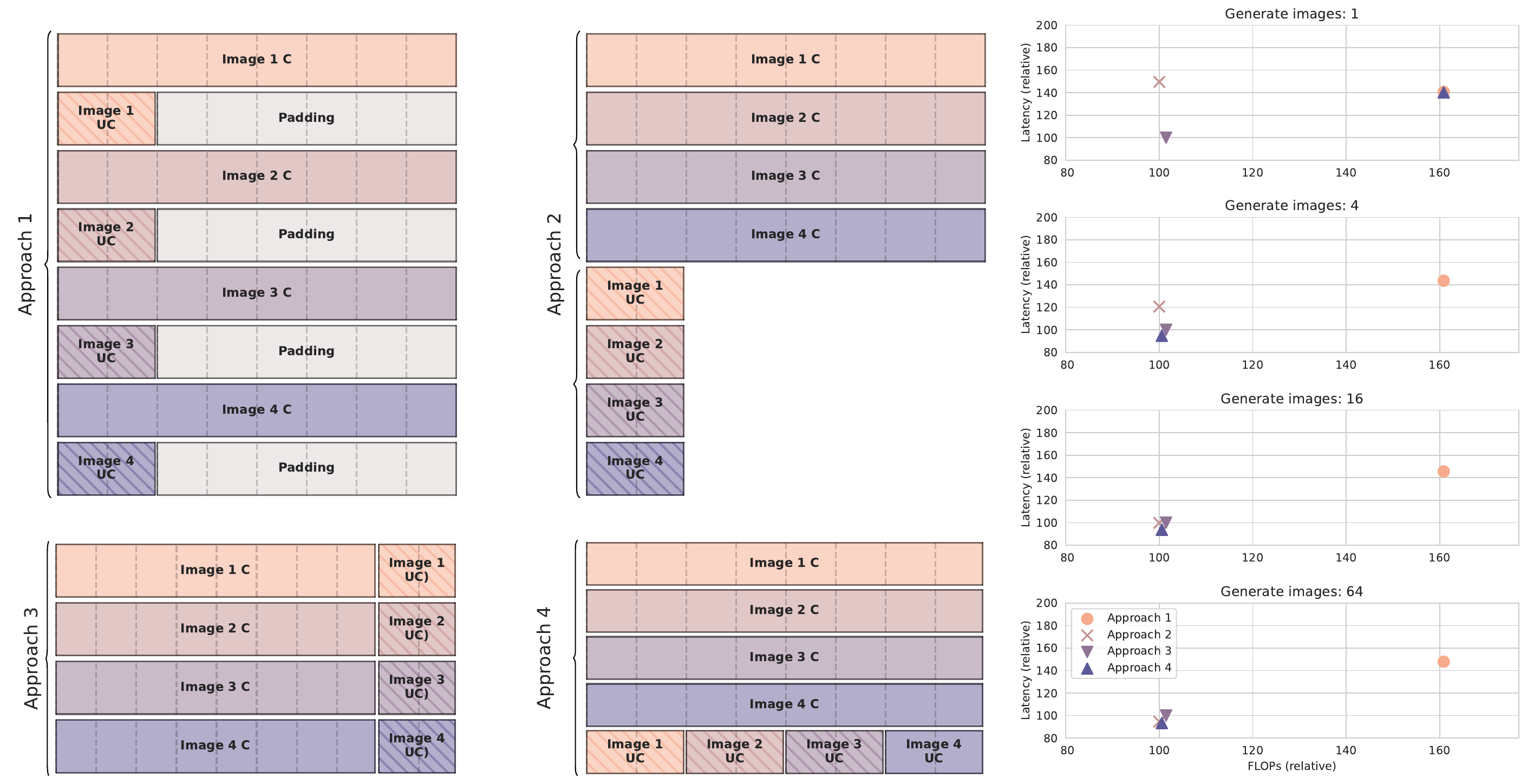}
    \caption{Different approaches can be employed to perform forward passes with CFG when the conditional (C) and unconditional (UC) predictions use different patch sizes. Here, each row corresponds to a sequence of tokens propagated through the DiT, and each bracket corresponds to a batch of sequences for a single NFE. Generally `Approach 2' leads to the smallest amount of FLOPs, but for batch size $1$, inference can be memory bound for low-resolution image generation. `Approach 4' mostly leads to the smallest latency, as long as the number of generated images is larger than $4$, i.e. the ratio of the sequence lengths between the powerful and the weak model. On the right, we plot FLOPs and Latency from the four different approaches of performing inference, for a different number of generated images. Batch size plays a role here (class-conditioned image generation experiments) as generated images are of lower resolution, namely $256 \times 256$, and thus sequence lengths through the Transformer are smaller. Normalized FLOPs are determined based on `Approach 2' and normalized latency based on `Approach 3'. We use \textit{torch.compile} with \textit{fullgraph=True} and \textit{mode = 'reduce-overhead'}.}%
    \label{fig:packing}
\end{figure}

We provide more details in Fig.~\ref{fig:packing}, on how to perform inference with CFG when the conditional and unconditional predictions employ a different patch size. We show this for our class-conditioned model, but results easily generalize for all our~\flexidit~models. Performing CFG entails NFEs with double the batch size (or 2 distinct NFEs), for the conditional and unconditional input, respectively. Performing the conditional and unconditional calls with different patch sizes leads to propagating sequences of different lengths through the DiT. Depending on how these sequences are `packed' together, and for lack of a hardware-specific implementation of masked attention, more FLOPs can be traded for better latency. Our weak model additionally leads to memory benefits, which can be traded for a bigger batch size when serving the model. Notice that current state-of-the-art image generation models in practice require much longer sequences compared to the $256 \times 256$ images generated here (see also Section~\ref{sec:latency}) and so generation is compute-bound even when generating with batch size equal to 1.

\subsection{What does the Model Learn?}
\label{sec:intuition}

\begin{figure}[!h]
  \begin{center}
    \includegraphics[width=0.8\linewidth]{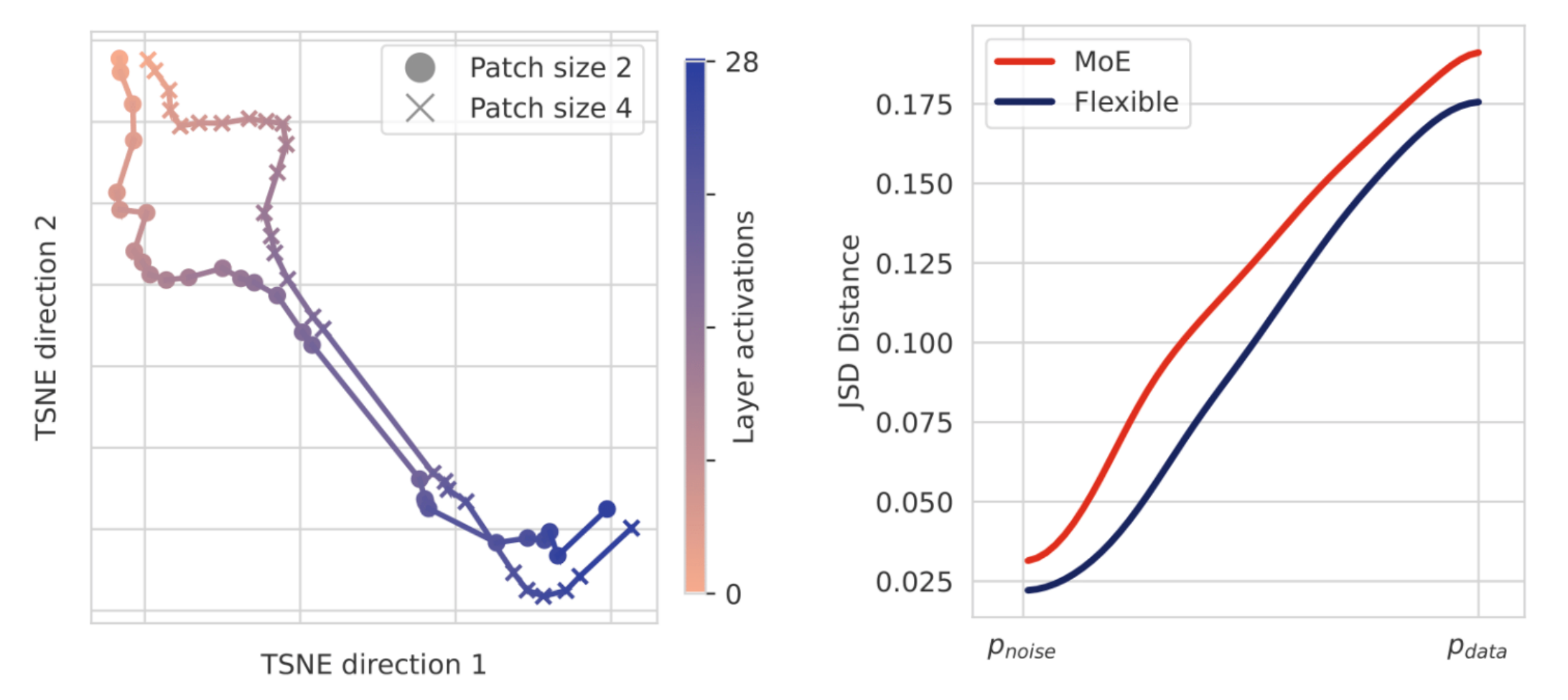}
    \caption{Interpretability of the model activations and attention scores, when propagating samples tokenized with different patch sizes.}
    \label{fig:intuition}
  \end{center}
\end{figure}

Transformers are composed of a series of channel mixing components --- feed-forward layers with shared weight applied to all tokens in the sequence --- and token mixing components --- attention applied to tokens in the sequence. By coercing the model to learn the denoising objective when applied to images processed with different patch sizes, we are enforcing inductive bias in its weights and helping it better understand global structures in the image~\citep{d2021convit, raghu2021vision, xiao2021early}. We test this hypothesis and evaluate what the model is learning in the following ways in Fig.~\ref{fig:intuition}. (left) We visualize using t-SNE, centered kernel alignment (CKA) between feature maps across layers when performing NFEs with different patch sizes. Activations across layers exhibit similar transformations~\citep{von2024language}, except the early layers, where features are lower level, i.e. more patch specific. (right) We visualize the Jensen–Shannon divergence (JSD) between attention maps (interpolated to the same image space) when performing NFEs with different patch sizes. We compare using our~\flexidit~model with different patch sizes (Flexible) versus using two static models trained with different patch sizes (namely \textit{DiT-XL/2} and a trained from scratch \textit{DiT-XL/4}). Our flexible model showcases lower JSD, demonstrating better knowledge transfer between the different patch sizes. We believe that this ``transfer'' of knowledge is crucial to (1) confirm that parameter sharing across patch sizes is valid and (2) ensure that fine-tuning can be fast and sample efficient. 

\subsection{Class-Conditioned Image Generation}

\begin{figure}[!h]
    \centering
    {{\includegraphics[width=0.33\textwidth]{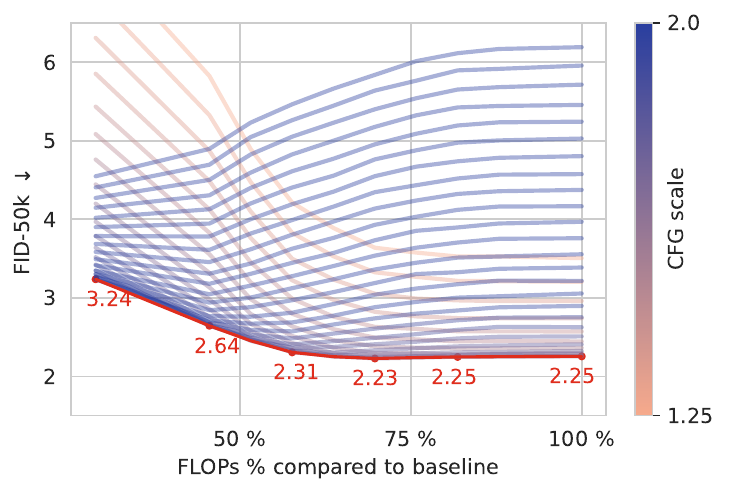} }}%
    \hfill
    {{\includegraphics[width=0.33\textwidth]{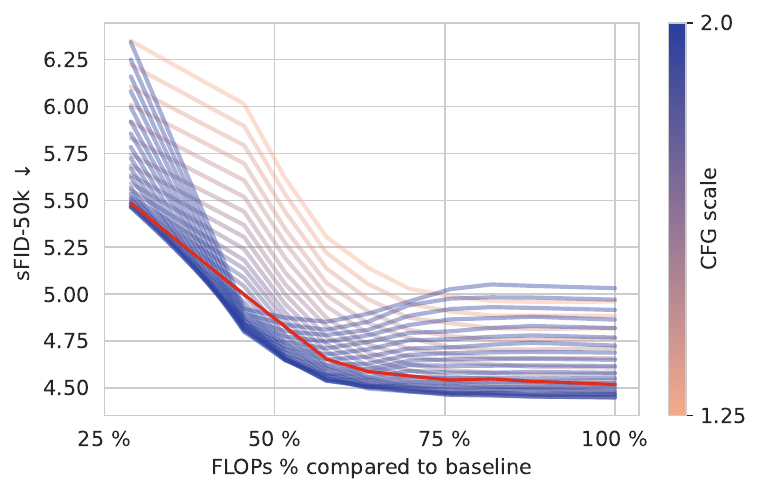} }}%
    \hfill
    {{\includegraphics[width=0.33\textwidth]{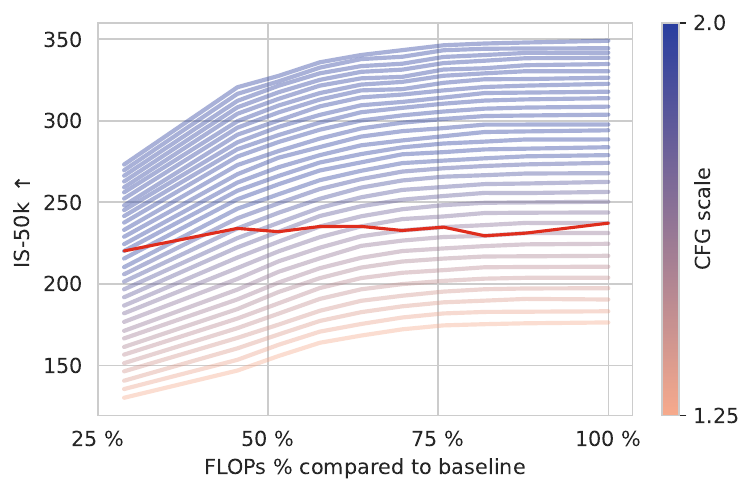} }}%
    \hfill
    {{\includegraphics[width=0.33\textwidth]{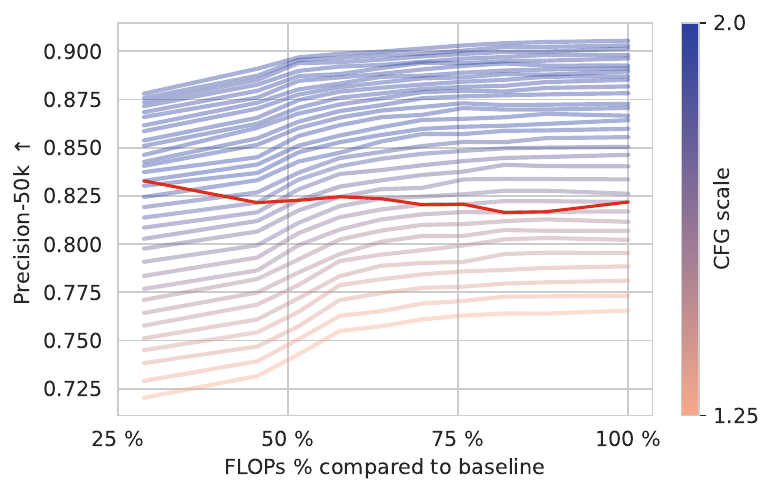} }}%
    {{\includegraphics[width=0.33\textwidth]{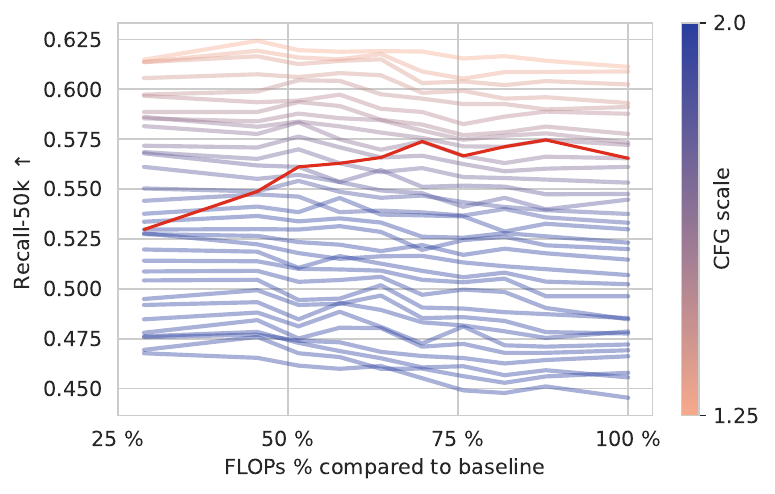} }}%
    \hfill
    \caption{More metrics for our~\flexidit~based on the \textit{DiT-XL/2} model for class-conditioned generation on \ImageNet. We plot (a) FID (b) sFID, (c) inception score, (d) precision, and (e) recall when generating $50,000$ samples with $250$ steps of the DDPM schedule for various values of the CFG scales. Red lines correspond to the values that lead to the optimum FID scores for each compute level.}%
    \label{fig:xl-256-more-results}
\end{figure}

\begin{figure}[!h]
    \centering
    \includegraphics[width=0.6\textwidth]{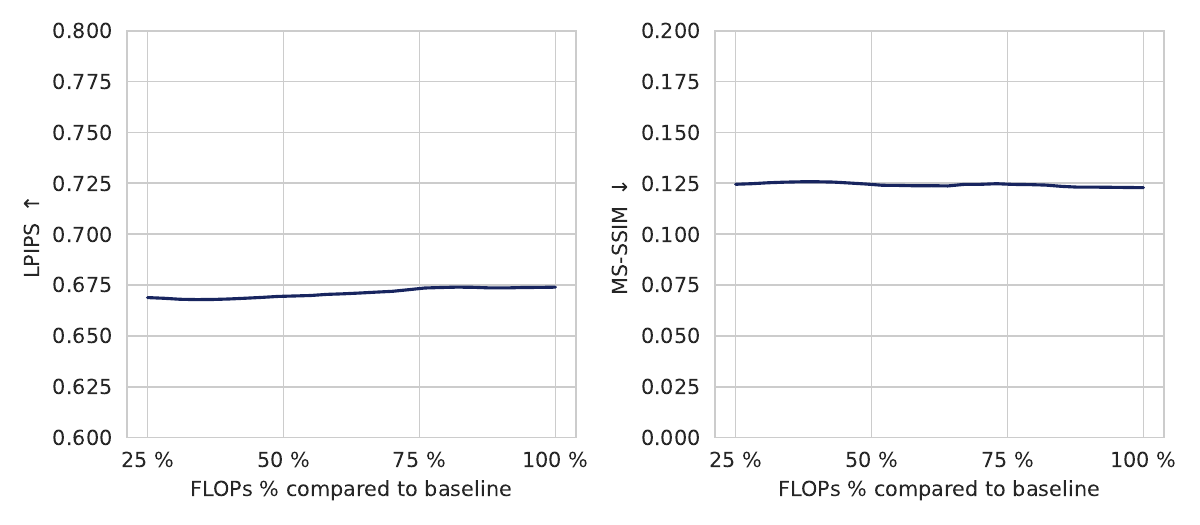}
    \caption{Average distance/similarity of images generated from the same label map. Both metrics take values between $0$ and $1$.}
    \label{fig:diversity}
\end{figure}

\begin{figure}[!h]
    \centering
    \includegraphics[width=\textwidth]{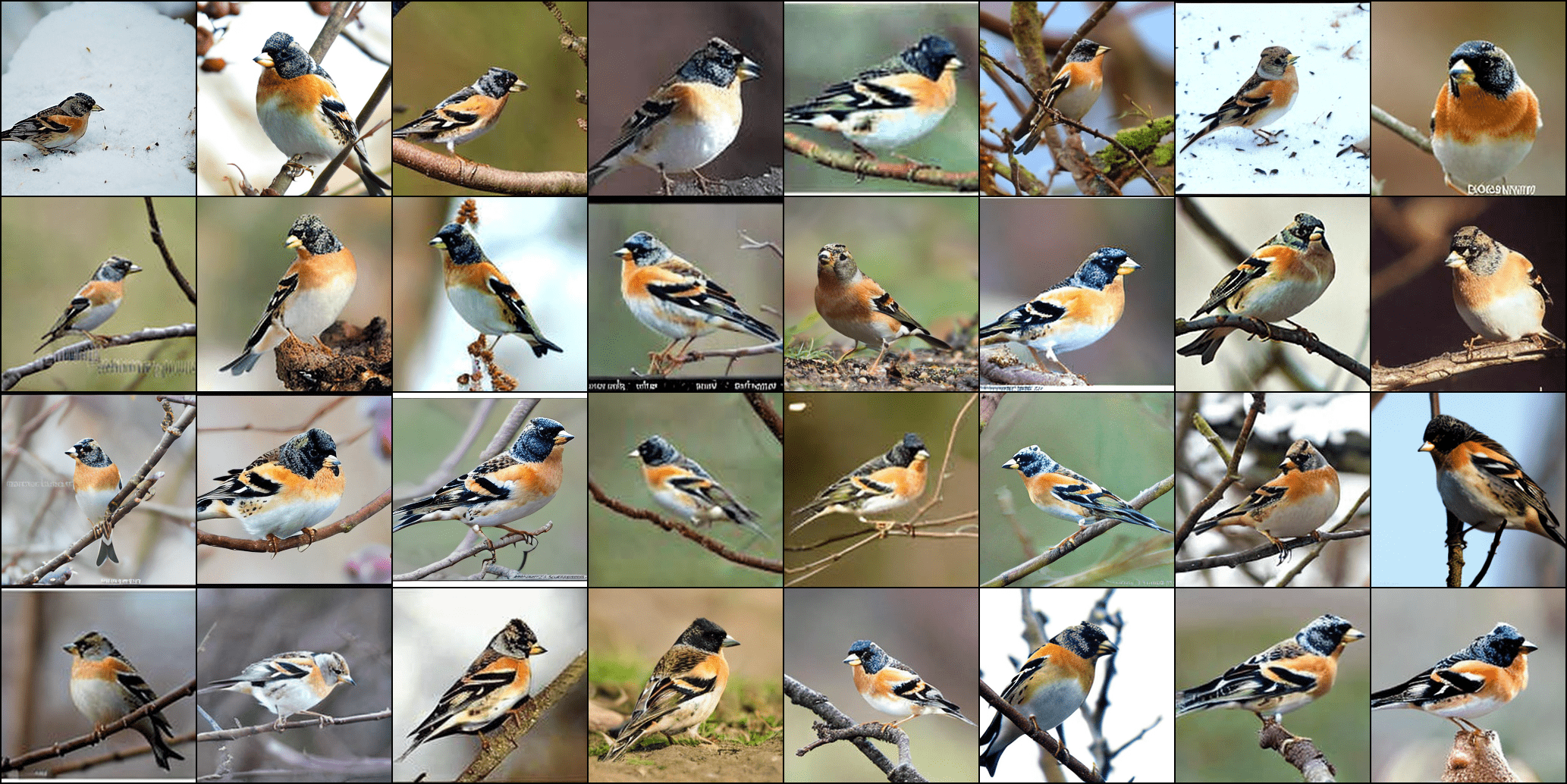}
    \caption{Sample images generated with the baseline \textit{DiT-XL/2} for the \ImageNet~category `Brambling'.}
    \label{fig:baseline_10}
\end{figure}

\begin{figure}[!h]
    \centering
    \includegraphics[width=\textwidth]{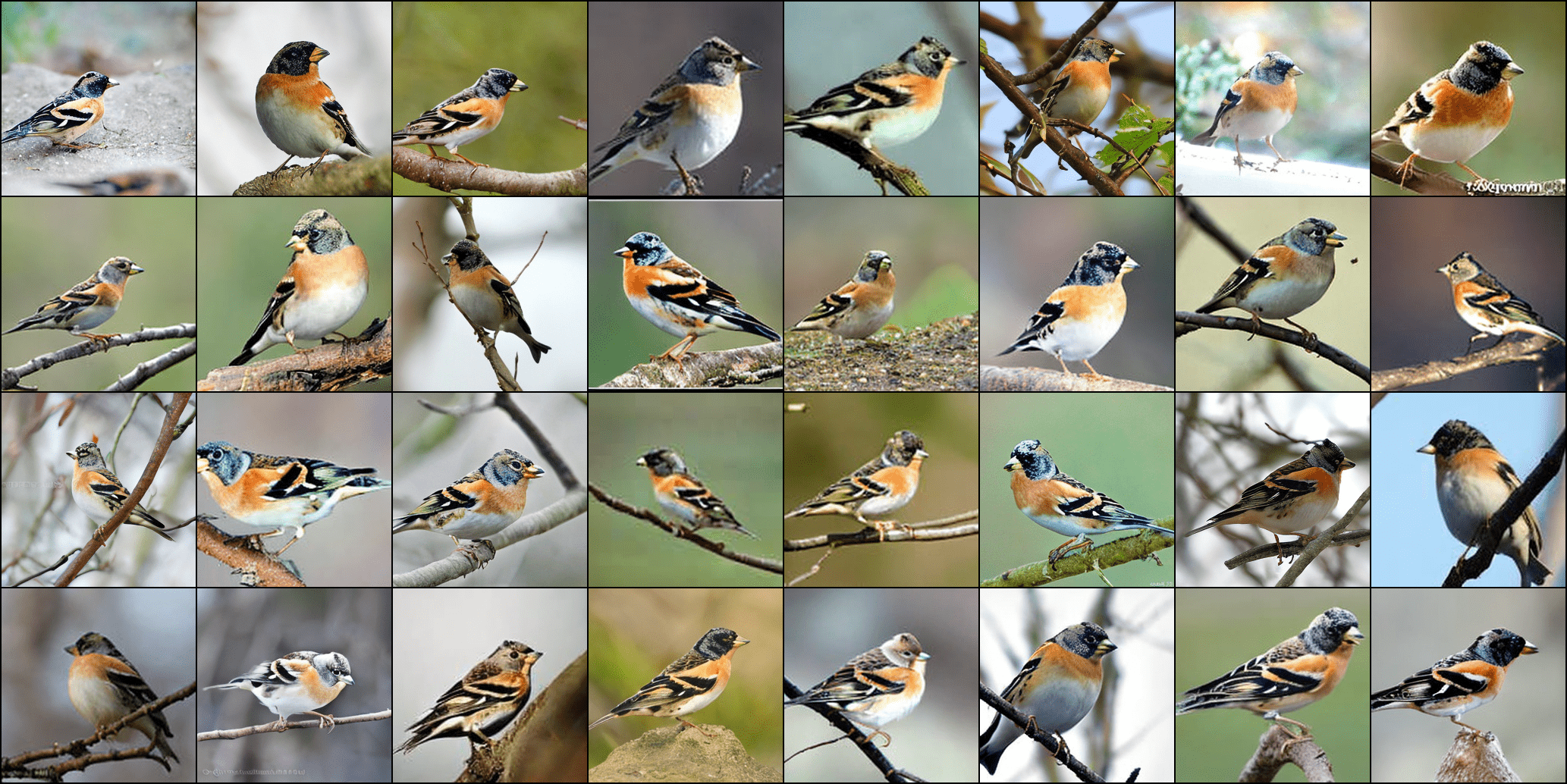}
    \caption{Sample images generated with our flexible \textit{DiT-XL} model when performing inference using only the powerful model, for the \ImageNet~category `Brambling'.}%
    \label{fig:flex_10}
\end{figure}

\paragraph{Additional metrics.}~For class-conditioned experiments on the main text we focused on the \textit{DiT-XL/2}~\citep{peebles2023scalablediffusionmodelstransformers} and FID as a metric. Here, we report more metrics apart from FID, namely Inception Score (IS), sFID, and Precision/Recall. Results are presented in Fig.~\ref{fig:xl-256-more-results} for our flexible \textit{DiT-XL/2} model. 
We remind that for class-conditioned models, we fine-tune models using our distribution matching loss. As a result, the powerful model that we get after fine-tuning is different to the pre-trained checkpoints we start from. 
To verify that our weak model does not lead to less diverse samples, we embark on a small experimental study to guarantee the diversity of generated images. We follow~\citep{sushko2020you} and generate images from the same label map. We then calculate pairwise similarity/distance between these images and average across all similarities/distances and all label maps. We use MS-SSIM~\citep{wang2003multiscale}, LPIPS~\citep{zhang2018unreasonable} and plot results in Fig.~\ref{fig:diversity}. Results indicate very similar values in terms of the diversity of the generated images. We also provide some sample images demonstrating diversity from the baseline model in Fig.~\ref{fig:baseline_10} and our tuned model in Fig.~\ref{fig:flex_10}. Note that the diversity of the generated images is in general high and not affected much by using the weak model for more of the initial denoising steps.

\begin{figure}[!ht]
    \centering
    \includegraphics[width=\textwidth]{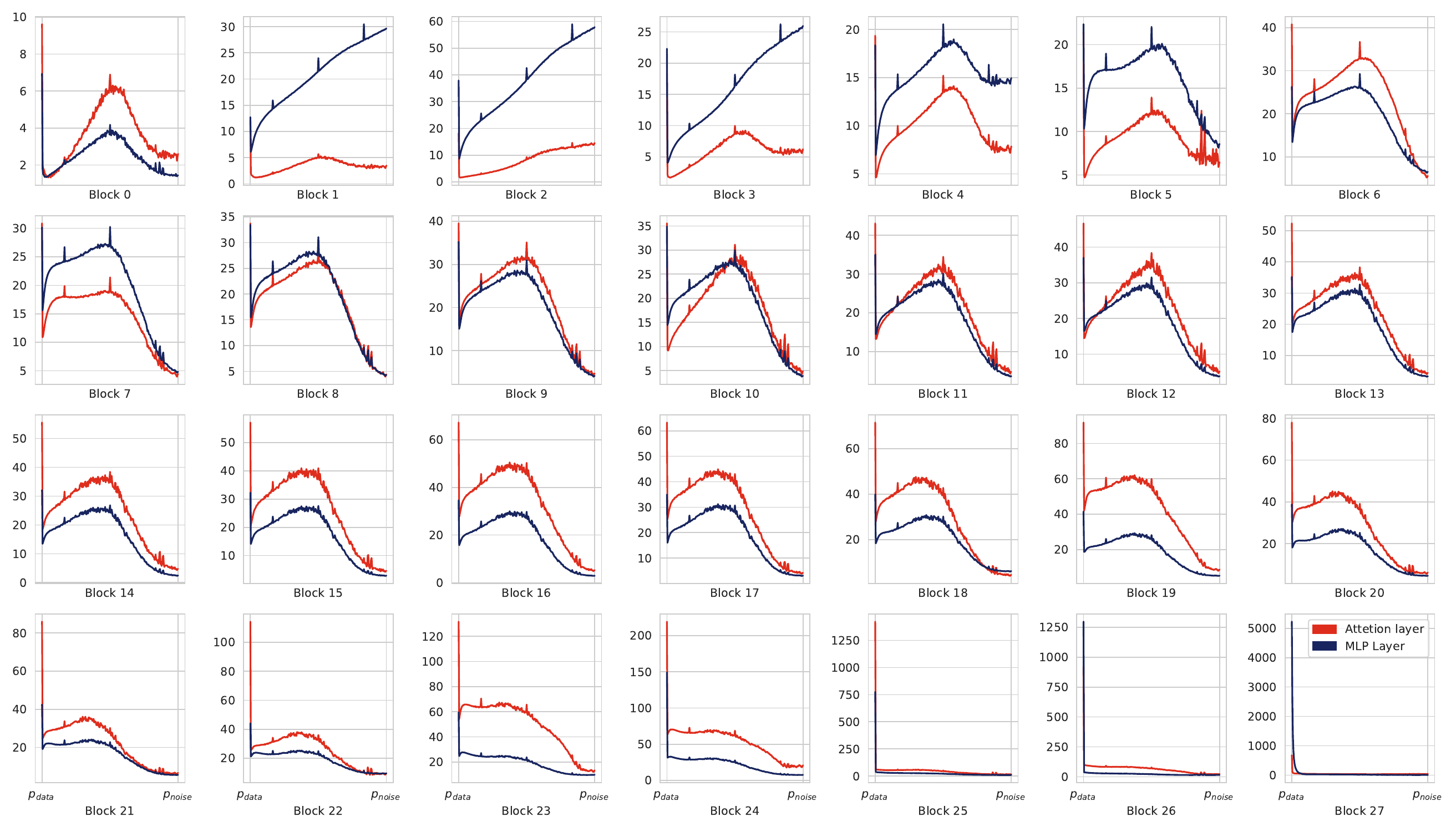}
    \caption{We plot average distance ($L_2$-norm) between activation of different layers during successive steps of the denoising process of the \textit{DiT-XL/2} model. Different layers exhibit different characteristics. Similar observations have been made in~\citep{ma2024learning}.}%
    \label{fig:caching_layers}
\end{figure}

\paragraph{Caching distance.}
In the main text, we have shown how weak and powerful models generate more similar predictions during the early steps of the denoising process. Previous papers to accelerate diffusion have largely relied on caching~\citep{wimbauer2024cache, yuan2024ditfastattn, chen2024asyncdiff, ma2024deepcache, zhao2024real} previous activations, by taking advantage of the similarity in activations between successive steps. For completeness, we also plot the caching distance between activations of the same layer between successive generation steps in Fig.~\ref{fig:caching_layers}. In this paper, we do not employ caching but focus on an orthogonal approach. We advocate that all steps are important for high-quality image generation, as demonstrated by our experiments on reducing the overall number of generation steps. Instead of completely skipping steps, we simply invest less compute for them, and let the model decide how to allocate this compute.

\paragraph{Additional schedulers.}~Based on the results on activation distance between successive denoising steps (Fig.~\ref{fig:caching_layers}), one could argue that first denoising steps are also important and thus a better inference scheduler would deploy the powerful model for these as well. In practice, we found no benefit from deploying a scheduler that works like that. Notice though how activation distance is high for the first denoising steps only for some of the layers. We additionally experimented with dynamic schedulers that choose the patch size of each denoising step based on the activation distance of different layers between successive denoising steps. We did not find additional potential benefits. 

\begin{figure}[!h]
    \centering
    \includegraphics[width=0.5\linewidth]{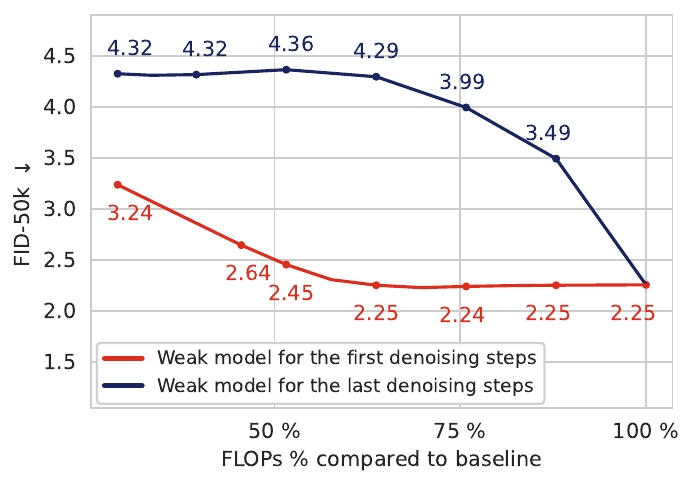}
    \caption{We compare our scheduler versus a different scheduler that uses the weak model for the last denoising steps when generating class-conditioned images. Points correspond to the minimum --- concerning CFG scale --- FID values.}
    \label{fig:opposite-scheduler}
\end{figure}

\begin{figure}[!h]
    \centering
    \includegraphics[width=\linewidth]{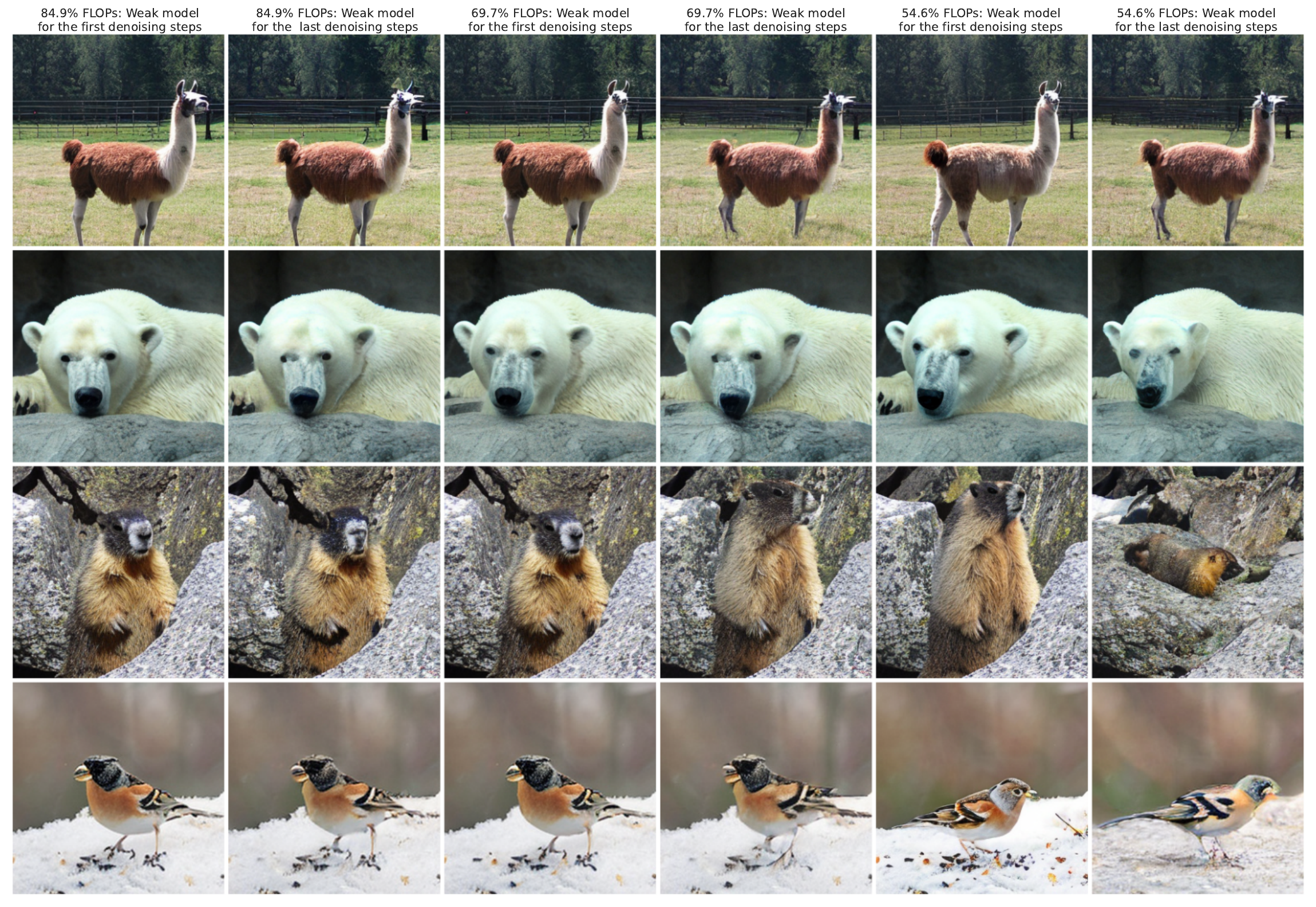}
    \caption{We compare our scheduler versus a different scheduler that uses the weak model for the last denoising steps when generating class-conditioned images. Using the weak model for the last denoising steps leads to images with lower image fidelity.}
    \label{fig:opposite-examples}
\end{figure}

In this paper, we are training a single model that can denoise images with any patch size for any denoising step. Given a fixed desired inference scheduler --- i.e. if we know exactly which $t$'s to run with the powerful and the weak model ---, one can train a model specifically based on that, leading to undoubtedly better quality images for the same compute. Similar techniques are regularly applied in consistency models~\citep{song2023consistency}. Finally, we compare our scheduler --- performing the first $\diffusionstepsweak$ denoising steps with a weak model --- versus the opposite scheduler, i.e. performing the last $\diffusionstepsweak$ denoising steps with a weak model. Results in Fig.~\ref{fig:opposite-scheduler} indicate that, as expected, using the weak model in the last diffusion steps is suboptimal, leading to a loss in fine-grained details. We also provide qualitative examples of how these different schedulers affect image quality in Fig.~\ref{fig:opposite-examples}.


\paragraph{More results on CFG.}~In the main text, we presented results on performing inference with different CFG scales and different invocations to our weak model for the unconditional and conditional part. The $4$ generated curves in Fig.~\ref{fig:xl-experiments} (middle) correspond to performing our scheduler as $250/250$, $130/130$, $70/70$, and $30/0$ where $x/y$ means using the powerful model for the last $x$ denoising steps for the conditional and $y$ denoising steps for the unconditional part. When performing CFG, we use the update rule as presented in the main text \begin{align*}
    \begin{cases}
      \etheta(\x_{t - 1} | \x_{t}, \emptyset; \patchsizeuncondition) + \guidancescaleone (\etheta(\x_{t - 1} | \x_{t}, \condition; \patchsizecondition) - \etheta(\x_{t - 1} | \x_{t}, \emptyset; \patchsizeuncondition)), & \text{if}\ \patchsizecondition = \patchsizeuncondition \\
      \etheta(\x_{t - 1} | \x_{t}, \condition; \patchsizeuncondition) + \guidancescaletwo (\etheta(\x_{t - 1} | \x_{t}, \condition; \patchsizecondition) - \etheta(\x_{t - 1} | \x_{t}, \condition; \patchsizeuncondition)), & \text{if}\ \patchsizecondition < \patchsizeuncondition
    \end{cases}.
\end{align*}
\begin{figure}[!h]
  \begin{center}
    \includegraphics[width=0.55\textwidth]{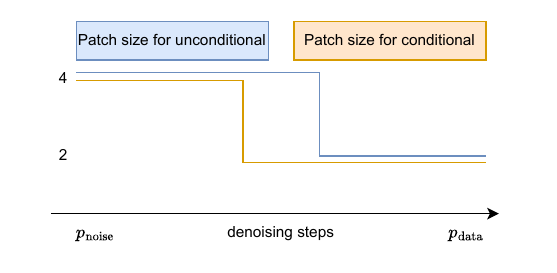}
  \end{center}
\end{figure}
This guidance scheme seeks to reduce errors made by the powerful model, enhancing potential differences in predictions of the corresponding weak model, when the two models disagree, indicating the general direction towards higher-quality samples. In practice, different values of $\guidancescaleone$ and $\guidancescaletwo$ lead to the best results. We find that the rule $(1 - \guidancescaleone) / (1 - \guidancescaletwo) = 2.5$, works consistently across experiments. Although we fix the value of the CFG scale during inference, different combinations are likely to lead to higher quality images as demonstrated by previous work~\citep{castillo2023adaptive}, which we leave for future exploration.

\begin{figure}[!h]
    \centering
    {{\includegraphics[width=0.33\textwidth]{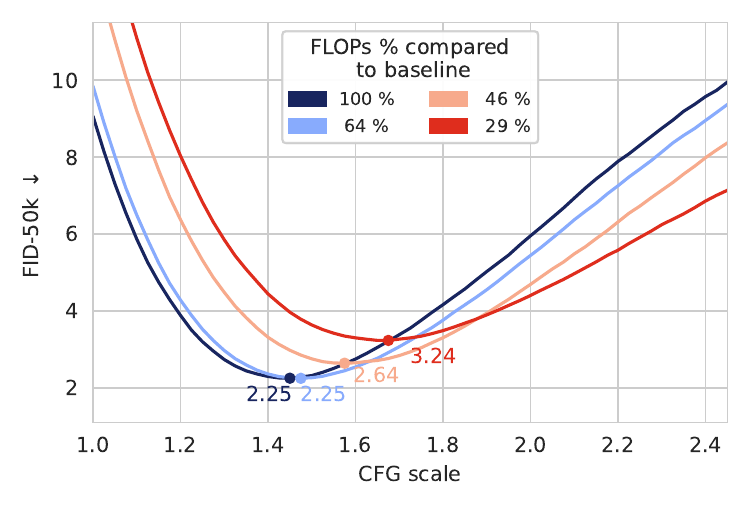} }}%
    \hfill
    {{\includegraphics[width=0.33\textwidth]{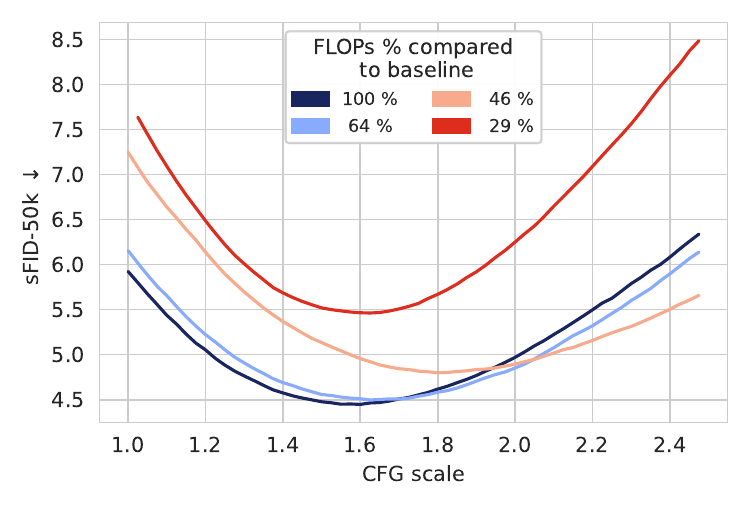} }}%
    \hfill
    {{\includegraphics[width=0.33\textwidth]{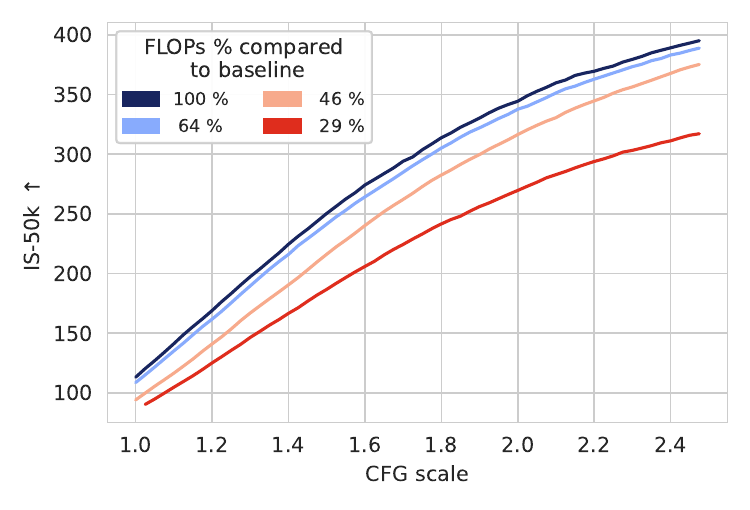} }}%
    \hfill
    {{\includegraphics[width=0.33\textwidth]{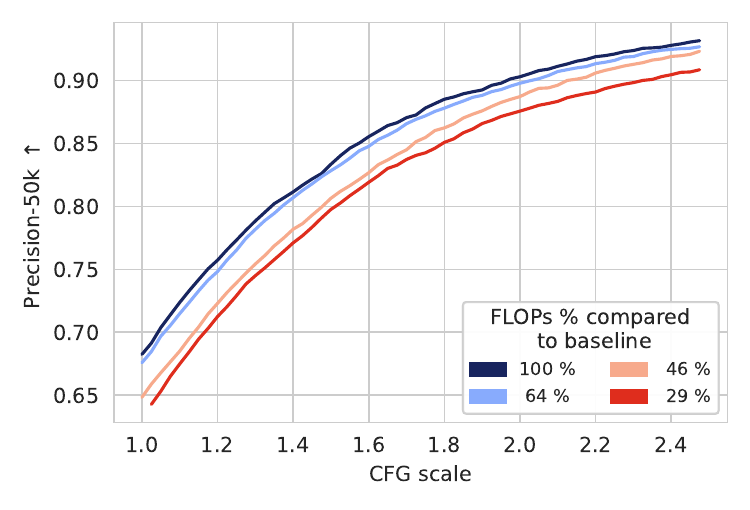} }}%
    {{\includegraphics[width=0.33\textwidth]{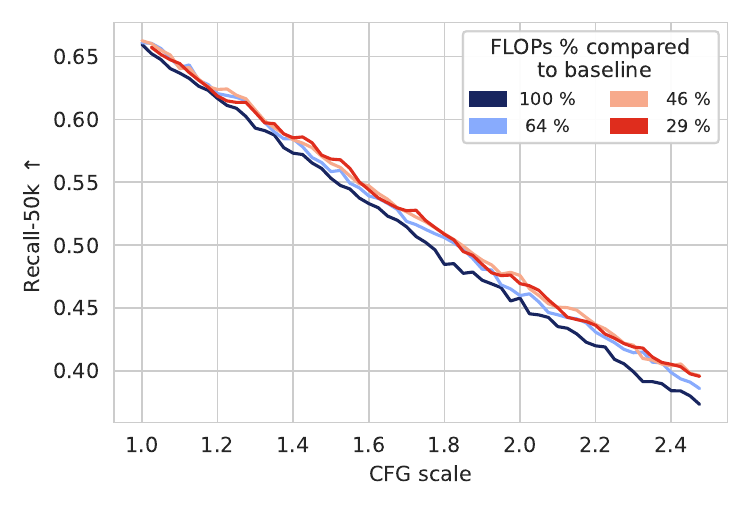} }}%
    \hfill
    \caption{Effect of CFG scale on the generated images from our class-conditioned~\flexidit~model. We plot (a) FID, (b) sFID, (c) inception score, (d) precision, and (e) recall when generation $50,000$ samples with $250$ steps of the DDPM scheduler.}%
    \label{fig:detailed-cfg-results}
\end{figure}

We point out that our scheduler is very stable in terms of performance attained for similar compute. For instance, performing inference with a $70/70$ scheduler or a $90/50$ scheduler, which both require the same overall compute, produces FID results of $2.64$ and $2.65$ respectively. Finally, we present detailed experiments on the effect of the CFG scale for different levels of compute and more metrics in Fig.~\ref{fig:detailed-cfg-results}.

\subsection{Text-to-Image Experiments}
\label{sec:details_text_to_image}

Generally, T2I generation is performed for a fixed target CFG scale. For our experiments we choose $\guidancescale = 4.5$ for the~\pixart~model, as this is the value used in~\citep{chen2023pixart} and $\guidancescale = 6.0$ for the~\emu~model, as for these values we observed the best quality images. In general, we can match with our dynamic inference other target values of the CFG scale. One simply needs to adjust the used CFG scale for the dynamic inference accordingly.

\begin{figure}[!htb]
    \begin{minipage}{.31\textwidth}
        \centering
        \includegraphics[width=\textwidth]{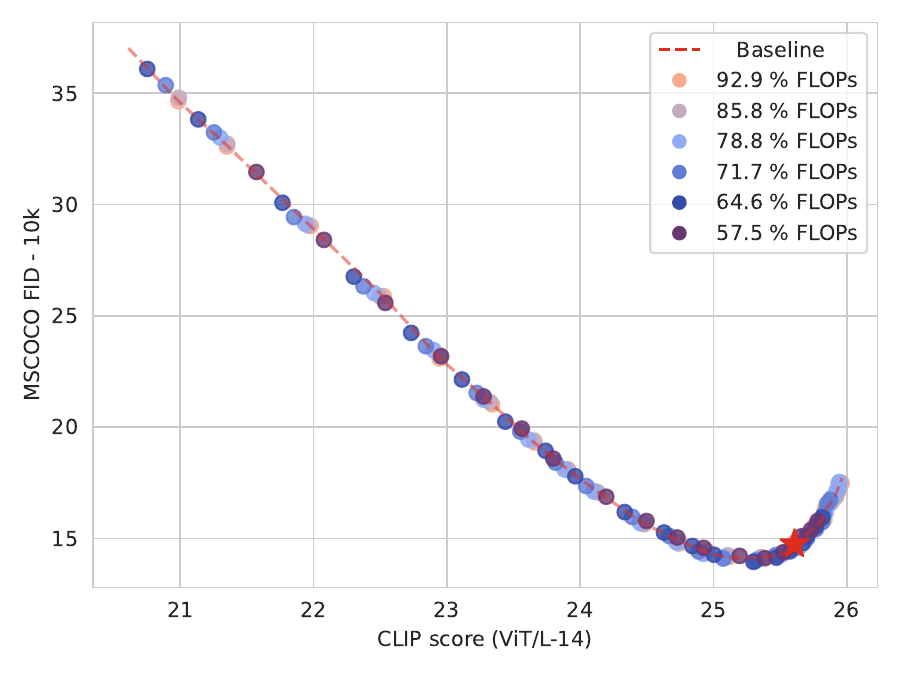}
        \caption{FID vs CLIP score using iDDPM for $100$ steps for the~\pixart~model.}%
        \label{fig:pixart_pareto_iddpm}
    \end{minipage}%
    \hfill
    \begin{minipage}{0.31\textwidth}
        \centering
        \includegraphics[width=\textwidth]{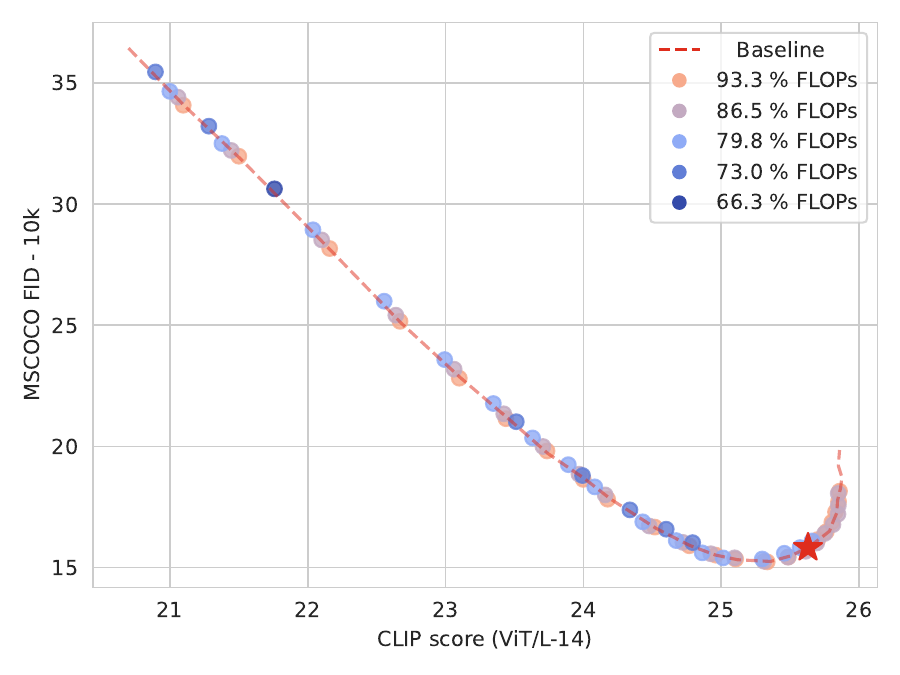}
        \caption{FID vs CLIP score using the DPM-solver for $20$ steps for the~\pixart~model.}%
        \label{fig:pixart_pareto_dpm}
    \end{minipage}
    \hfill
    \begin{minipage}{0.31\textwidth}
        \centering
        \includegraphics[width=\textwidth]{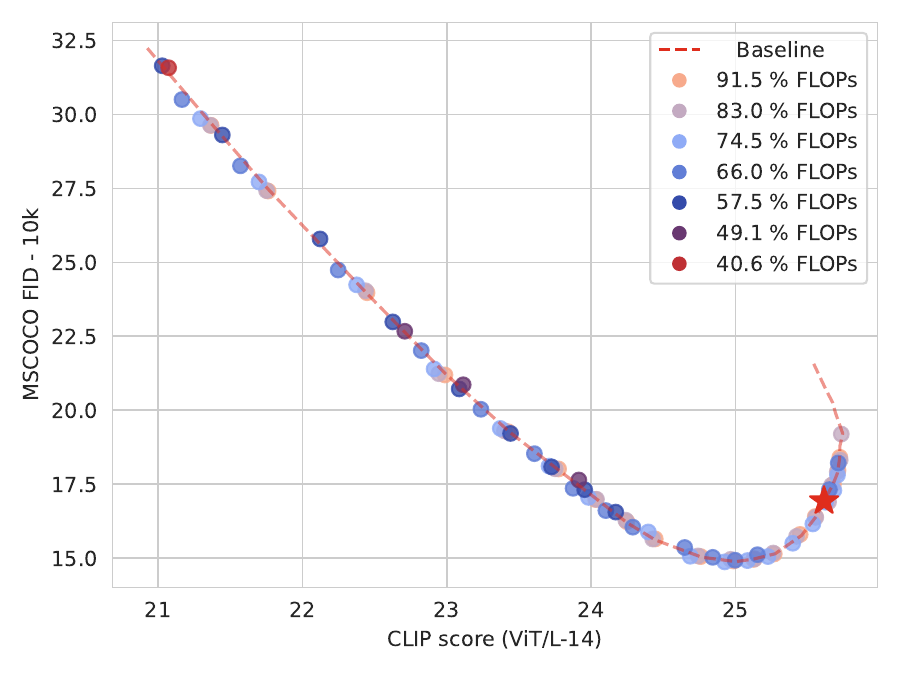}
        \caption{FID vs CLIP score using the SA-solver for $25$ steps for the~\pixart~model.}%
        \label{fig:pixart_pareto_sa}
    \end{minipage}
\end{figure}

We follow the evaluation protocol of PIXART-$\alpha$~\citep{chen2023pixart} and perform inference using the same solvers as they do, namely iDDPM~\citep{dhariwal2021diffusion} for $100$ steps, DPM solver~\citep{lu2022dpm} for $20$ steps, and SA solver~\citep{xue2024sa} for $25$ steps. In the main text --- Fig.~\ref{fig:t2i-experiments} (left) --- we presented results for the iDDPM solver. We present results for all the schedulers with the settings used in PIXART-$\alpha$ in Fig.~\ref{fig:pixart_pareto_iddpm},~\ref{fig:pixart_pareto_dpm} and~\ref{fig:pixart_pareto_sa}. For all the schedulers, there are settings where we reach the Pareto front of FID vs CLIP score of the baseline model with a lot less required compute.

\begin{figure}[!h]
    \centering
    \includegraphics[width=\textwidth]{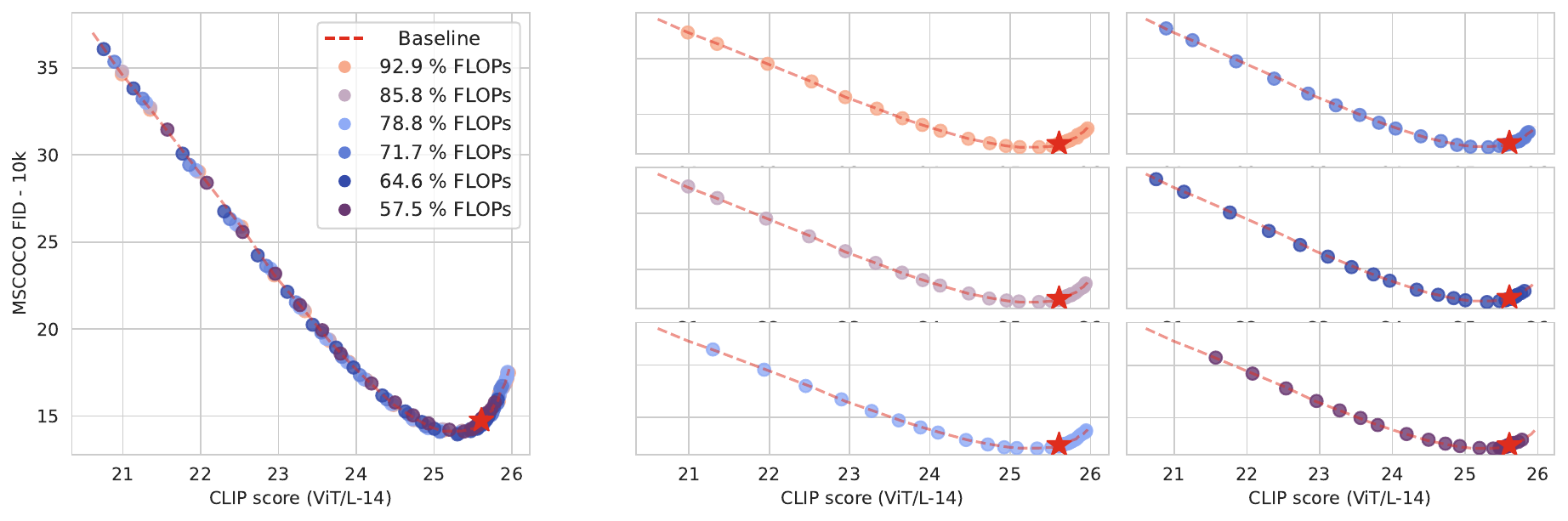}
    \caption{FID vs CLIP score using DDPM for $100$ steps for the~\pixart~model, for different levels of compute. On the right, we present results separately for each compute level.}%
    \label{fig:pareto-full-100-steps}
\end{figure}

To better characterize the effect of reducing compute, i.e. heavier use of the weak model, we also present more detailed results for the DDPM scheduler in Fig.~\ref{fig:pareto-full-100-steps}. Less compute-heavy inference schedulers, often produce images with smaller possible maximum CLIP scores (for large CFG guidance scales $\guidancescale$). In practice, as large CFG scale values lead to larger values of FID, these are less preferred. In every case, our weak models can max the FID vs CLIP score tradeoff of the base model for the default configuration used, i.e. $\guidancescale = 4.5$.

\begin{figure}[!h]
    \centering
    \includegraphics[width=0.5\textwidth]{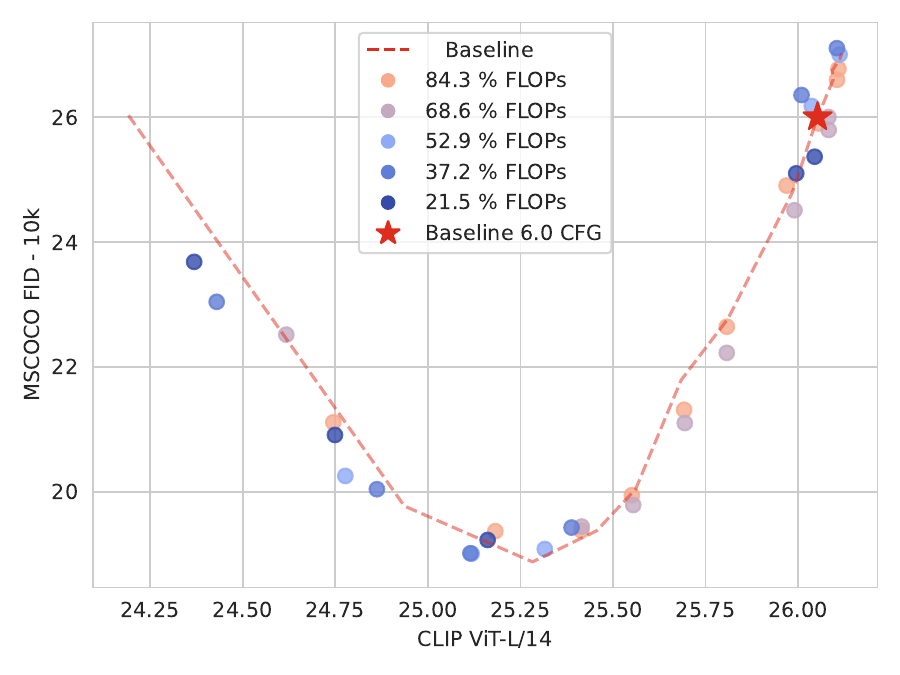}
    \caption{FID vs CLIP score using DDIM for $50$ steps for the~\emu~model.}%
    \label{fig:pareto-emu-50-steps}
\end{figure}

\begin{figure}[!h]
        \centering
        \includegraphics[width=0.7\textwidth]{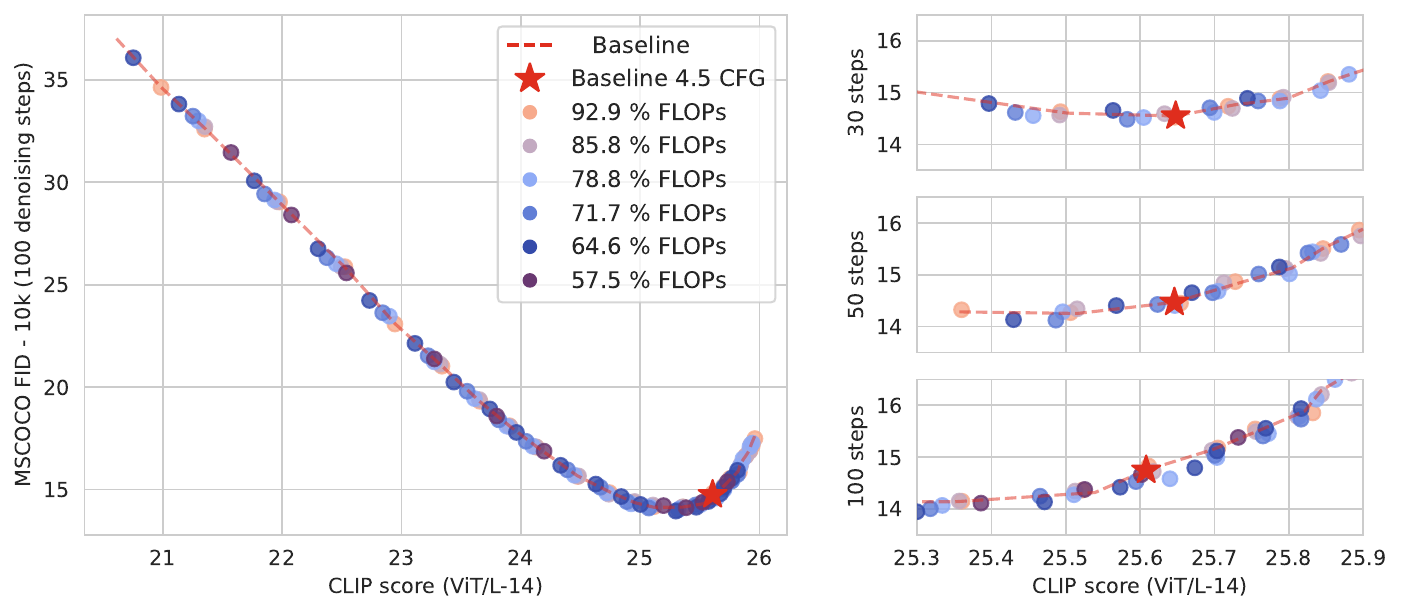}
        \caption{We plot FID vs CLIP score when generating images with different CFG scales. (left) Overall Pareto front when generating images with $100$ denoising steps. (right) Pareto front generating images with a different number of steps zoomed in the typical tradeoff generation values.}
        \label{fig:pixart_pareto_steps}
\end{figure}

We also provide results on using a smaller overall number of steps with the DDPM solver in Fig.~\ref{fig:pixart_pareto_steps}. To generate the baseline curves, we sample $10,000$ samples using a CFG scale $\guidancescale$ from the set $\{1.0, 1.125, 1.25, 1.375, 1.5, 1.625, 1.75, 2.0, 2.25, 2.5, 2.75, 3.0, 3.5, 4.0, 4.5, 5.0, 5.5, 6.0, 6.5, 7.0, 7.5, 8.0, 8.5\}$. The same values are used when sampling with our flexible models. Finally, we also provide FID vs CLIP for the~\emu~model in Fig.~\ref{fig:pareto-emu-50-steps}. In this case, we take CFG scales $\guidancescale$ from the set $\{1.0, 1.5, 2.0, 2.5, 3.0, 4.5, 6.0, 7.5, 8.0, 9.0\}$. We use captions from the training set of \text{MS COCO} to generate images. Neither of the models was trained on images from this dataset.

\paragraph{VQA results.}~VQA scores are calculated by querying a visual-question-answering model to produce an alignment
score by computing the probability of a "Yes"  answer to a simple "Does this
figure show \{text\}?". question. To calculate this score, we use the \textit{clip-flant5-xxl} model from huggingface\footnote{\url{https://huggingface.co/zhiqiulin/clip-flant5-xxl}} as suggested in ~\citep{lin2025evaluating}. We provide more detailed results on the VQA benchmark in Tables~\ref{tab:pixart_vqa_detailed} and~\ref{tab:emu_vqa_detailed}. More specifically, we provide per dataset VQA scores, along with the CFG scale $\guidancescale$ used to generate the images for each case  . As we can see, using the weak model requires a bigger CFG scale to reach the same level of optimality (calculated from the FID vs CLIP score tradeoff). We also note that using the weak model often leads to images with better text alignment. We hypothesize that fewer tokens (as a result of larger patch sizes) help with spatial consistency at the beginning of the denoising process. To calculate VQA scores, we take the first $200$ prompts from each dataset and use the \textit{train} split from the \textit{DrawBench}~\citep{saharia2022photorealistic}, \textit{train} split from the \textit{Pick-a-Pic}~\citep{kirstain2023pick}, \textit{test} split from the \textit{Winoground}~\citep{thrush2022winoground} and \textit{tifa\_v1.0\_text\_inputs}\footnote{\url{https://github.com/Yushi-Hu/tifa/blob/main/tifa_v1.0/tifa_v1.0_text_inputs.json}} from the \textit{TIFA160}~\citep{hu2023tifa} dataset.

\begin{table}[!h]
    \centering
    {\normalsize
    \begin{tabular}{ |c|c|cccc|c| } 
    \toprule
     & \makecell{CFG scale \\ $\guidancescale$} & \textit{DrawBench} & \textit{Pick-a-Pic} & \textit{Winoground} & \textit{TIFA160} & Average \\
    \midrule
    \pixart~($100$ \%) & $4.5$ & $58.93$ & $50.56$ & $62.01$ & $81.65$ & $63.29$ \\
    \pixart~($92.9$ \%) & $4.5$ & $58.37$ & $51.53$ & $62.09$ & \underline{$82.16$} & $63.54$ \\
    \pixart~($85.8$ \%) & $4.5$ & $57.58$ & $51.67$ & $62.41$ & $81.53$ & $63.30$ \\
    \pixart~($78.8$ \%) & $4.7$ & $58.62$ & $51.97$ & $62.04$ & $80.60$ & $63.31$ \\
    \pixart~($71.7$ \%) & $4.7$ & $58.44$ & $51.56$ & $63.03$ & $80.57$ & $63.40$ \\
    \pixart~($64.6$ \%) & $4.9$ & \underline{$60.16$} & \underline{$52.34$} & $62.98$ & $80.06$ & \underline{$63.89$} \\
    \pixart~($57.7$ \%) & $5.0$ & $59.04$ & $50.80$ & \underline{$63.27$} & $79.90$ & $63.26$ \\
    \pixart~($50.5$ \%) & $5.0$ & $56.77$ & $51.72$ & $61.87$ & $79.38$ & $62.44$ \\
    \pixart~($43.4$ \%) & $5.0$ & $56.87$ & $51.92$ & $61.30$ & $78.33$ & $62.11$ \\
    \bottomrule
    \end{tabular}
    }
    \caption{Detailed VQA evaluations for the benchmarks tested with the~\pixart~model. As in the class-conditioned experiments, using more of the weak model during denoising requires a higher CFG scale $\guidancescale$ to reach optimum performance.}
    \label{tab:pixart_vqa_detailed}
\end{table}

\begin{table}[!h]
    \centering
    {\normalsize
    \begin{tabular}{ |c|c|cccc|c| } 
    \toprule
     & CFG-scale $\guidancescale$ & \textit{DrawBench} & \textit{Pick-a-Pic} & \textit{Winoground} & \textit{TIFA160} & Average \\
    \midrule
    \emu~($100$ \%) & $6.0$ & $69.44$ & $58.70$ & $65.75$ & \underline{$86.77$} & $70.17$ \\
    \emu~($84.3$ \%) & $6.0$ & $68.00$ & $58.93$ & \underline{$67.33$} & $86.51$ & $70.19$ \\
    \emu~($68.6$ \%) & $6.25$ & $69.53$ & \underline{$60.62$} & $66.00$ & $85.33$ & \underline{$70.37$} \\
    \emu~($52.9$ \%) & $6.5$ & \underline{$69.79$} & $58.14$ & $66.23$ & $86.20$ & $70.09$ \\
    \bottomrule
    \end{tabular}
    }
    \caption{Detailed VQA evaluations for the benchmarks tested with the~\emu~model. As in the class-conditioned experiments, using more of the weak model during denoising requires a higher CFG scale $\guidancescale$ to reach optimum performance.}
    \label{tab:emu_vqa_detailed}
\end{table}

\paragraph{Alignment between powerful and weak model.}~It is common practice nowadays to train images (and especially videos) in different stages, where a large (potentially lower quality) dataset is used for the first stage, followed by a shorter fine-tuning stage, characterized by higher quality and aesthetically more pleasing images. Although we are directly distilling the weak model from the predictions of the powerful model, the data used throughout training are still important. In practice, our fine-tuning is sample efficient, and we find that even a few thousand images ($<5000$) are enough to succeed. We thus suggest fine-tuning on the last (potentially smaller) but higher-quality dataset. When generating images based on shorter prompts with~\emu, we use a prompt re-writer, prompting a small LLM to expand on the information provided. We consider this prompt re-writer as part of the model. 






\section{Implementation Details}
\label{app:implementation-details}

We provide additional details on the experiments in the main text.

\subsection{Figure Details}

\paragraph{Prompts used for Fig.~\ref{fig:emu-examples-0}.}~We provide in Table~\ref{tab:emu-prompts} the exact prompts used to generate the images.

\begin{table}[!h]
    \centering
    \begin{tabular}{|p{16cm}|}
        \toprule
        \multicolumn{1}{|c|}{Prompts for Fig.~\ref{fig:emu-examples-0}} \\
        \midrule
        The image shows a frog wearing a golden crown with intricate designs, sitting on a wooden log in a serene environment reminiscent of a Japanese anime setting. The frog's crown is adorned with small gems and its eyes are large and expressive. The log is covered in moss and surrounded by lush greenery, with a few cherry blossoms visible in the background. The frog's skin is a vibrant shade of green with blue stripes, and it has a regal demeanor, as if it is a monarch of the forest. The overall atmosphere is peaceful and whimsical. \\
        \midrule
        The image shows a serene waterfall cascading down a rocky slope in a lush tropical forest, reminiscent of Claude Monet's impressionist style. Sunlight filters through the dense foliage above, casting dappled shadows on the misty veil surrounding the falls. The water plunges into a crystal-clear pool, surrounded by large rocks and vibrant greenery. The atmosphere is tranquil, with a warm color palette and soft brushstrokes evoking a sense of serenity. The forest floor is covered in a thick layer of leaves, and the sound of the waterfall echoes through the air.\\
        \bottomrule
    \end{tabular}
    \caption{Details on the prompts used to generate the images in the paper.}
    \label{tab:emu-prompts}
\end{table}

\paragraph{Details on Fig.~\ref{fig:motivation}.}~In Fig.~\ref{fig:motivation} (b), we fix randomness of the denoising process in terms of the initial sampled image $\p(\x_\diffusionsteps)$, and from the denoising process in~\cref{eq:denoise_images}. Then we generate images with $250$ steps using the \textit{DiT-XL/2} official public checkpoint. During denoising, we modify only $1$ of the $250$ denoising steps, bypassing the model predictions from that step through a high/low pass filter. We then compare the resulting generated images in terms of LPIPS, $L_2$ distance, SSIM, and DreamSim. In general, modifying one of the first denoising steps, leads to larger final image differences, due to the accumulation of differences. We still note a distinctive pattern: a high-pass filter, i.e. removing low-pass components, leads to larger image differences during the first denoising steps. The opposite holds for the last denoising steps. We can thus argue, that low-pass components, i.e. `coarser' image details are more important compared to high-frequency details, for the first denoising steps. To calculate spatial frequencies, we keep the corresponding values of the FFT of an image.

\paragraph{Details on Fig.~\ref{fig:latency}.}~In Fig.~\ref{fig:latency}, we plot GPU utilization when propagating different sequence lengths. All experiments are conducted in \textit{bfloat16} using \textit{PyTorch 2.5} and \textit{CUDA 12.4} and with our~\emu~DiT that has $24$ layers and a hidden dimension of $2048$. We also plot peak FLOPs and memory bandwidth for the GPU tested, an \textit{NVIDIA H100 SXM5} in this case. When we report FLOPs, we count additions and multiplications separately, as it is commonly done~\citep{hoffmann2022training}. We count FLOPs as the theoretical required operations for a forward pass, and not the actual GPU operations, which might be a higher number due to the potential recalculation of partial results. As bytes for the x-axis, we only consider the model parameters ($2$ bytes per model parameter). Note that for our choice of weak models, the GPU is fully utilized (it reaches the maximum compute intensity that can be achieved for this application). In reality, compute intensity drops when larger sequence lengths are used, mainly due to the larger memory footprint of intermediate activations. Thus, latency benefits are indeed even larger than FLOPs benefits reported in the paper. Compiling and fusing operations is crucial to ensure that the GPU is not bottlenecked by waiting instructions from the CPU. When we are performing inference with the weak model without first merging the LoRAs, inference time is proportional to the additional FLOPs required. For the attention operation, we use the \textit{memory\_efficient\_attention} operation from the \textit{xformers} library. Other efficient implementations of attention do not lead to significant differences. Our T2I model operates in a $128 \times 128$ latent space, which means that using a patch size of $(1, 2, 2)$ results in $4096$ tokens, compared to $1024$ tokens for the $(1, 4, 4)$ patch size. Our T2V model operates in a $32 \times 88 \times 48$ latent space, which means that using a patch size of $(1, 2, 2)$, $(2, 2, 2)$, $(1, 4, 4)$ leads to $33792$, $16896$ and $8448$ tokens respectively. We obtain similar results using different-sized models, like our~\moviegen~model.

When we report FLOPs in the paper, we report numbers for the denoising process, as it is commonly done. We thus ignore latency induced by decoders that map samples from a latent space, or potential prompt-rewrite modules. The added latency by the decoder, is in our settings negligible.

\subsection{Flexifying Diffusion Transformers}

\begin{wrapfigure}{r}{0.14\textwidth}
  \begin{center}
    \includegraphics[width=0.14\textwidth]{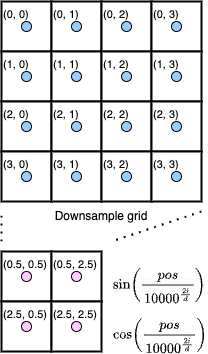}
  \end{center}
\end{wrapfigure}

Although for the class-conditioned experiments, we use a single embedding and de-embedding layer that we always project to the required patch size, we note that this projection can be done once to pre-calculate embedding and de-embeddding parameters during inference. These projected embeddings can then be used out of the box, for the tradeoff of some minuscule additional memory. The choice of $\patchsizeexp[\prime]$ as the underlying patch size is not too important for our experiments. In practice, we use a value of $\patchsizeexp[\prime] = 4$. As mentioned in the main text, we add positional embeddings according to the coordinates of each patch in the original image. A schematic of this can be found on the right.

Apart from the architecture modifications listed in the main text, we experimented with adding patch size specific temperatures in the self-attention computation:

\begin{equation*}
    \text{softmax}\left( \frac{\mathbf{Q}\mathbf{K}^\top}{\tau_{\patchsize}\sqrt{\hiddensize}} \right) \mathbf{V},
\end{equation*}
where $\mathbf{Q}, \mathbf{K}, \mathbf{V}$ are the queries, keys and values respectively and $\tau_{\patchsize}$ is a patch size specific temperature initialized as $1$. We do not include this in the end, as it occasionally leads to instabilities during fine-tuning, even under different parametrizations.

\paragraph{Class-conditioned implementation details.}~We largely use the same hyperparameters as~\citep{peebles2023scalablediffusionmodelstransformers} to fine-tune. When fine-tuning to match distributions, we train to minimize the MMD loss, as introduced in Section~\ref{sec:exposure_bias}. During bootstrapping, we denoise images with a DDPM scheduler, operating on $\diffusionsteps = 250$ steps, the same as the target inference scheduler. As we found that MMD distance is higher for diffusion steps closer to $x_0 \sim \q(\x_0)$ --- see also Fig.~\ref{fig:mmd} ---, we bias the sampling to reflect that during training as well.

\begin{table}[!h]
    \centering
    \begin{tabular}{ |c|c| } 
    \toprule
    Parameter & Value \\
    \midrule
    training data & \ImageNet \\
    learning rate & $10^{-4}$ \\
    weight decay & $0.0$ \\
    EMA update frequency & every step \\
    EMA update rate & $0.9999$ \\
    \bottomrule
    \end{tabular}
    \caption{Class-conditioned implementation details.}
    \label{tab:dit-implementation-details}
\end{table}

In our experiments, we focused on fine-tuning pre-trained models, as we were interested in efficiency. We note that training with different patch sizes has been used in the past to also accelerate pre-training~\citep{beyer2023flexivit, anagnostidisnavigating}. We believe that flexible patch sizes can also be used in this application to accelerate pre-training.

\paragraph{\pixart~implementation details.}
For the~\pixart~model, we follow exactly the recipe of~\citep{chen2023pixart}, and fine-tune a $256 \times 256$ pre-trained variant\footnote{Our starting pre-trained model exactly matches the public checkpoint~\url{https://huggingface.co/PixArt-alpha/PixArt-XL-2-SAM-256x256}.}. For fine-tuning, we use the same image dataset, namely the SAM dataset\footnote{\url{https://segment-anything.com/}.} with captions generated from a vision-language model. The model has overall the same parameters as the \textit{DiT-XL/2} model, with the addition of cross-attention blocks. When adding new embedding and de-embedding layers, we initialize them as we did for the class-conditioned experiments. Embedding layers are initialized to $\projectembedding^\dagger \weightembedding$ and de-embedding layers are initialized to $\weightdeembedding \projectdeembedding^\dagger$. Here $\weightembedding$, $\weightdeembedding$ are the pre-trained model parameters and $\projectembedding$, $\projectdeembedding$ are the same --- patch size dependent --- fixed projection matrices that better preserve the norm of the output activations at initialization. As aforementioned, we add a patch size embedding that is added to all tokens in the sequences after the tokenization step. This embedding is equal to $0$ for the pre-trained patch size, to ensure functional preservation. We fine-tune the~\pixart~model on a small subset of the SAM dataset used to originally train the target model. We add LoRAs on the self-attention and feed-forward layers, with a LoRA dimension of $32$. We use a higher learning rate, due to the different learning objectives --- distilling a powerful model's predictions into the ones of a weal model.

\begin{table}[!h]
    \centering
    \begin{tabular}{ |c|c| } 
    \toprule
    Parameter & Value \\
    \midrule
    training data & SAM with captions from a VLM model \\
    optimizer & \textit{AdamW} \\
    learning rate & $8 \times 10^{-4}$ \\
    weight decay & $10^{-2}$ \\
    gradient clipping & $0.02$ \\
    batch size & $512$ \\
    EMA update frequency & every step \\
    EMA update rate & $0.9999$ \\
    LoRA rank & $32$ \\
    \bottomrule
    \end{tabular}
    \caption{Image text-conditioned implementation details.}
    \label{tab:pixart-implementation-details}
\end{table}

\paragraph{\emu~implementation details.}~Our~\emu~model is fundamentally identical to the~\pixart~model. Small variations are due to different ways to calculate text embeddings, which lead to a different number and size of the cross-attention tokens, and slight architectural modifications --- primarily the use of QK-normalization~\citep{dehghani2023scaling} and the use of learnable positional embeddings. We train using a high-quality aesthetic dataset. To calculate metrics based on the $1024 \times 1024$ images generated with this model, we follow the evaluation protocol of~\citep{xu2023restart} and resize images to $512 \times 512$. For both our~\emu~and our~\moviegen~model, we use a LoRA rank of $64$.

\paragraph{T2V implementation details.}~As aforementioned, our~\moviegen~model has a pre-trained patch size of $(\patchsizef, \patchsizeh, \patchsizew) = (1, 2, 2)$. Compared to our T2I experiments, we only change the $2$D convolutional layers used for tokenization with a $3$D convolution layer. When increasing the temporal patch size $\patchsizef$, we duplicate parameters along that dimension. When interpolating positional embeddings, we also do that along the temporal dimension. No additional changes are made. For evaluation, we use the prompts from \url{https://github.com/facebookresearch/MovieGenBench/blob/main/benchmark/MovieGenVideoBench.txt} to generate videos of length equal to $256$ frames. We evaluate according to VBench~\citep{huang2024vbench} and report the average over~\textit{Subject Consistency},~\textit{Background Consistency},~\textit{Temporal Flickering},~\textit{Motion Smoothness},~\textit{Dynamic Degree},~\textit{Aesthetic Quality},~\textit{Imaging Quality},~\textit{Temporal Style} and~\textit{Overall Consistency}.

\subsection{Human Evaluation Details}
\label{sec:human_evals}

We prompt humans, asking them to: ``Compare the two side-by-side images. Focus on visual appeal and flawlessness, considering factors like aesthetic appeal, clarity, composition, color harmony, lighting, and overall quality, then select 'left' if the left image is better, 'right' if the right image is better, or 'tie' if both are equally appealing or you have no preference.''. In total, we collected votes for the $4$ different settings presented in the paper and aggregated them across $200$ prompts. For each setting and each prompt, we ask $3$ people for votes. In cases where there are $3$ votes for each of 'left', 'right', and 'tie', we ask a fourth labeler to break the tie.

\section{Generated Samples}
\label{app:generations}

We provide more examples of generated samples.

\subsection{Text-Conditioned Image Generation}

\begin{figure}[!h]
    \centering
    \includegraphics[width=0.92\linewidth]{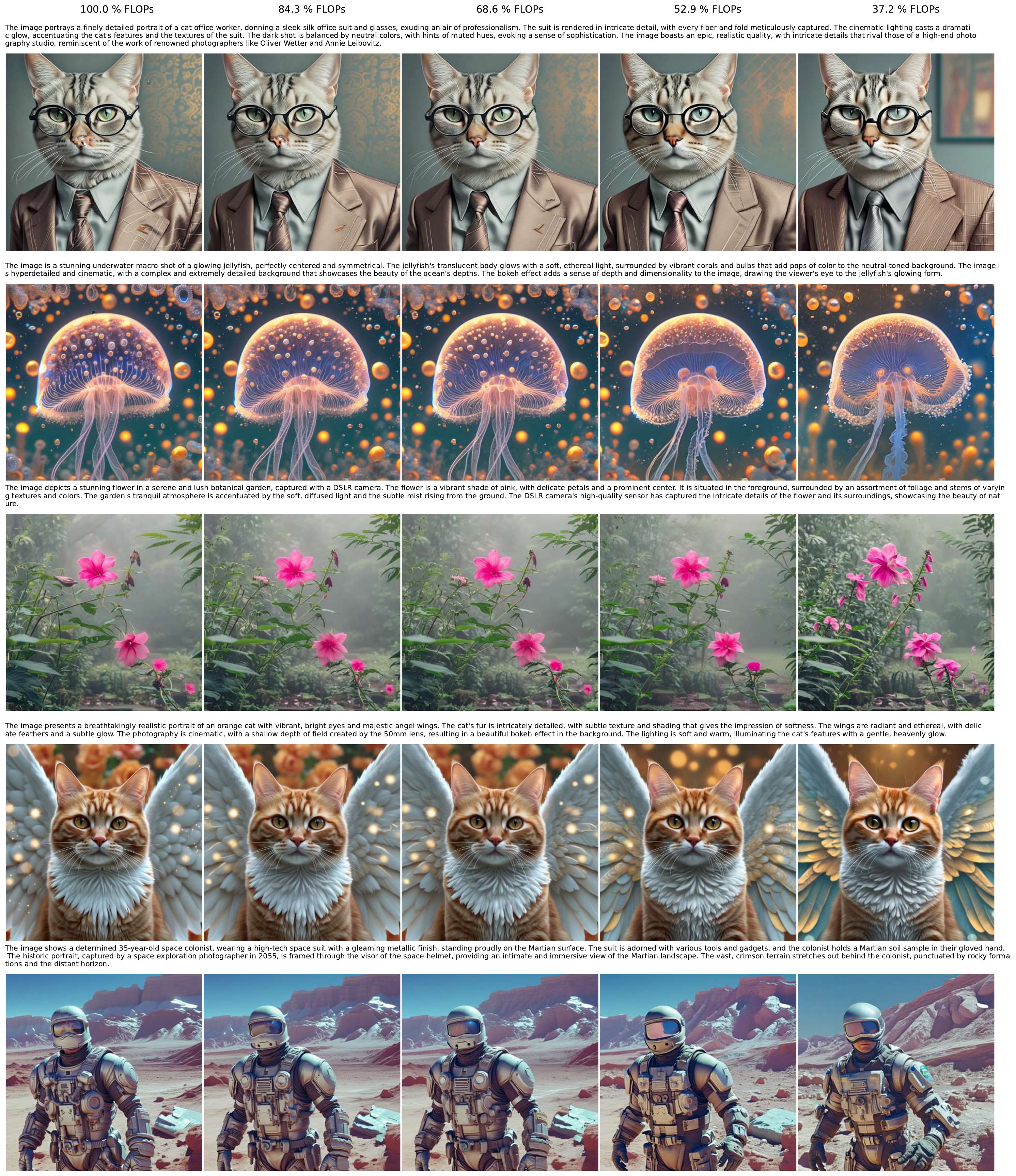}
    \caption{More samples generated by our~\emu~model for varying amounts of compute.}
    \label{fig:emu-examples-1}
\end{figure}

\begin{figure}[!h]
    \centering
    \includegraphics[width=0.9\linewidth]{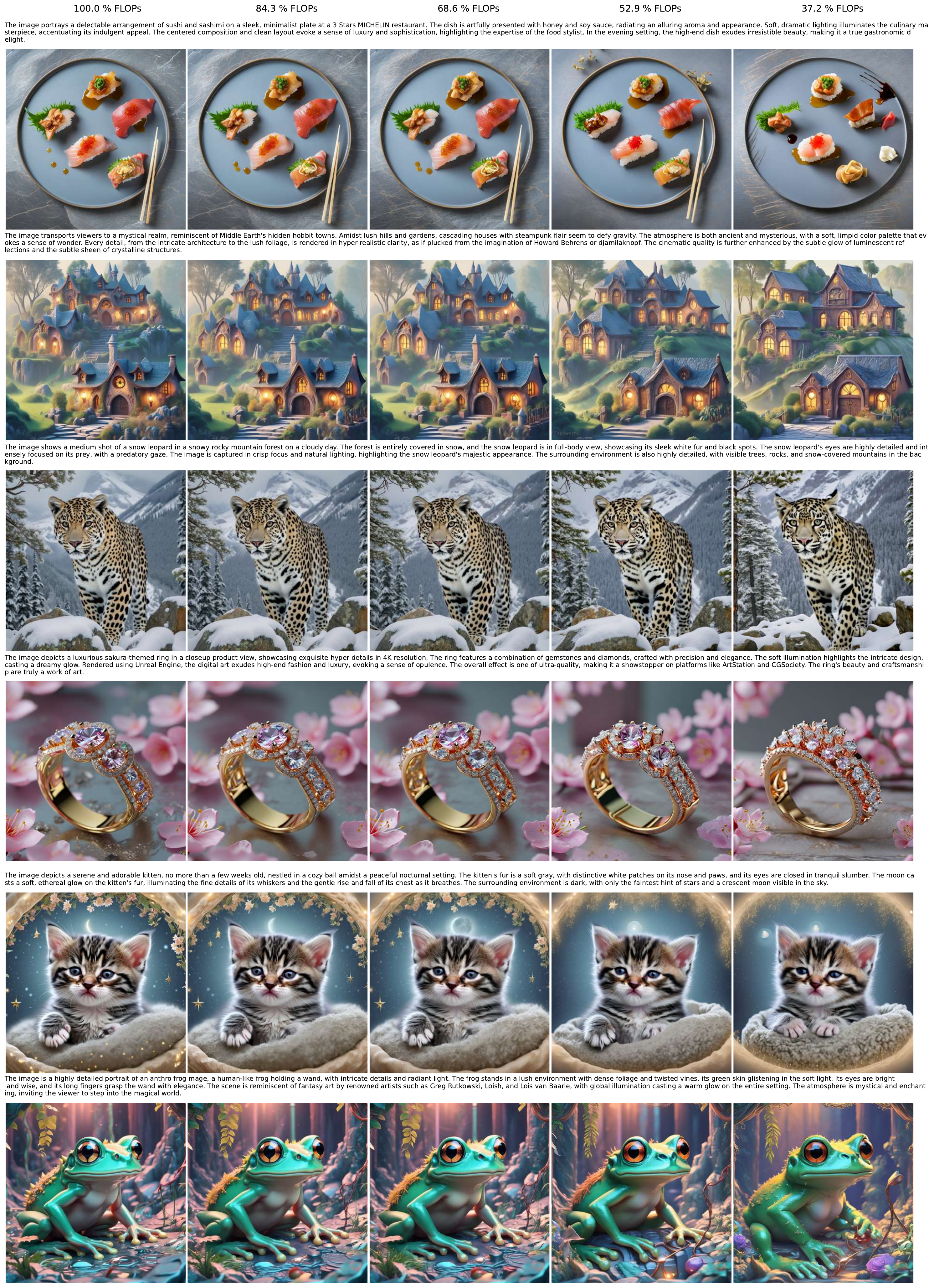}
    \caption{More samples generated by our~\emu~model for varying amounts of compute.}
    \label{fig:emu-examples-2}
\end{figure}

\begin{figure}[!h]
    \centering
    \includegraphics[width=0.9\linewidth]{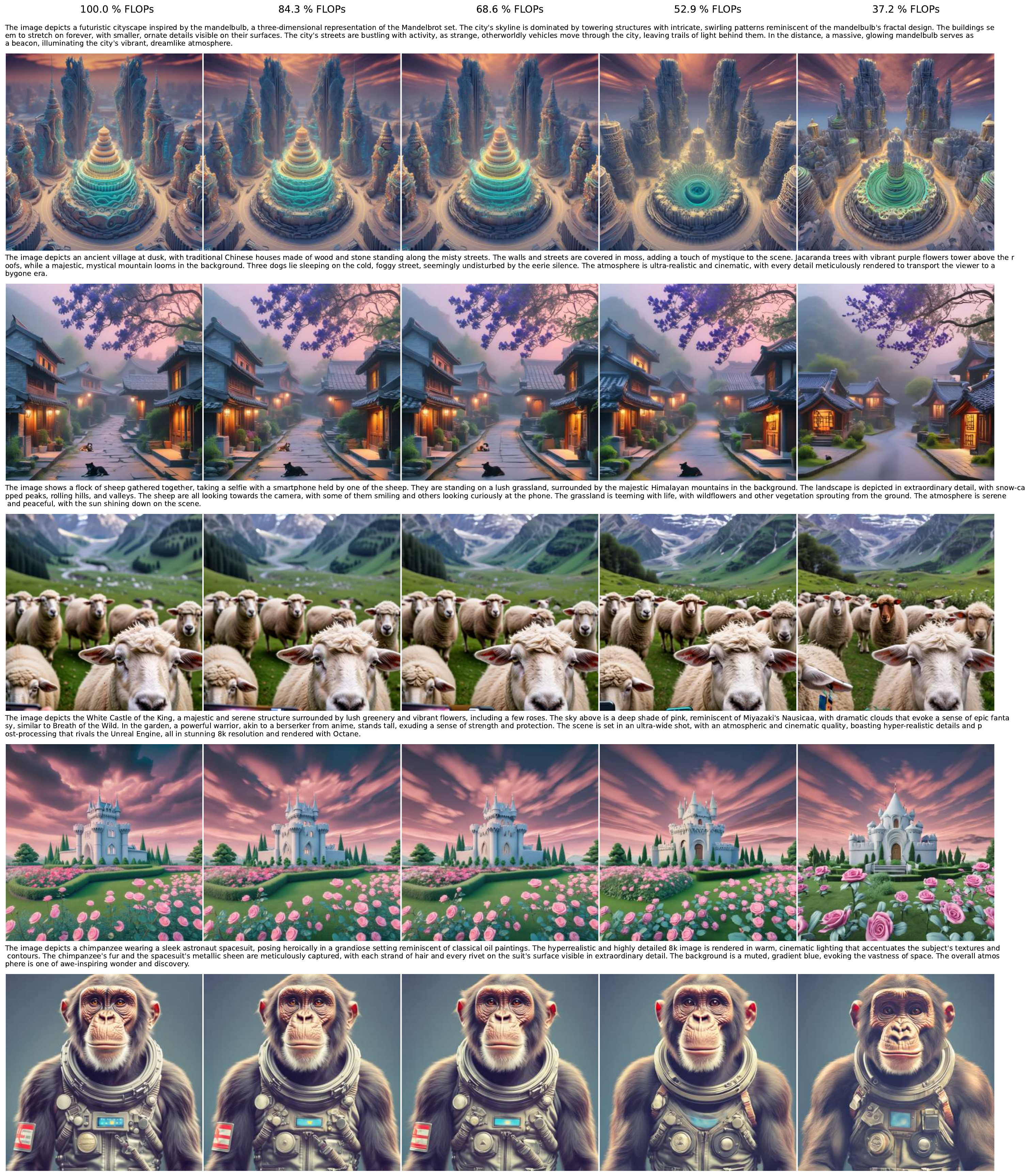}
    \caption{More samples generated by our~\emu~model for varying amounts of compute.}
    \label{fig:emu-examples-3}
\end{figure}

\begin{figure}[!h]
    \centering
    \includegraphics[width=0.95\linewidth]{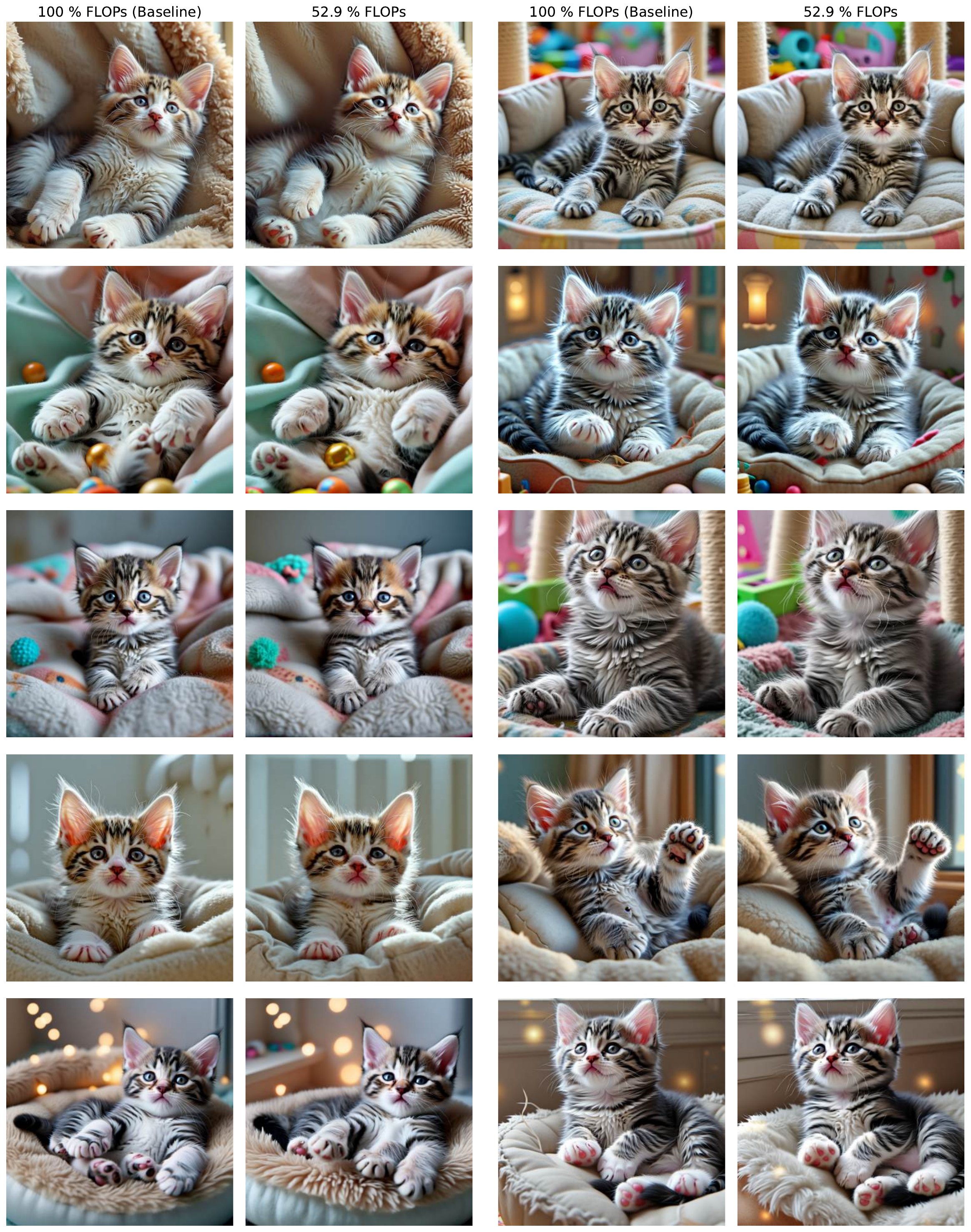}
    \caption{Samples for the prompt: "A playful kitten just waking up.". We showcase the image generated by the baseline and our flexible scheduler using only 53\% of FLOPs.}
    \label{fig:emu-kitten}
\end{figure}

\begin{figure}[!h]
    \centering
    \includegraphics[width=0.95\linewidth]{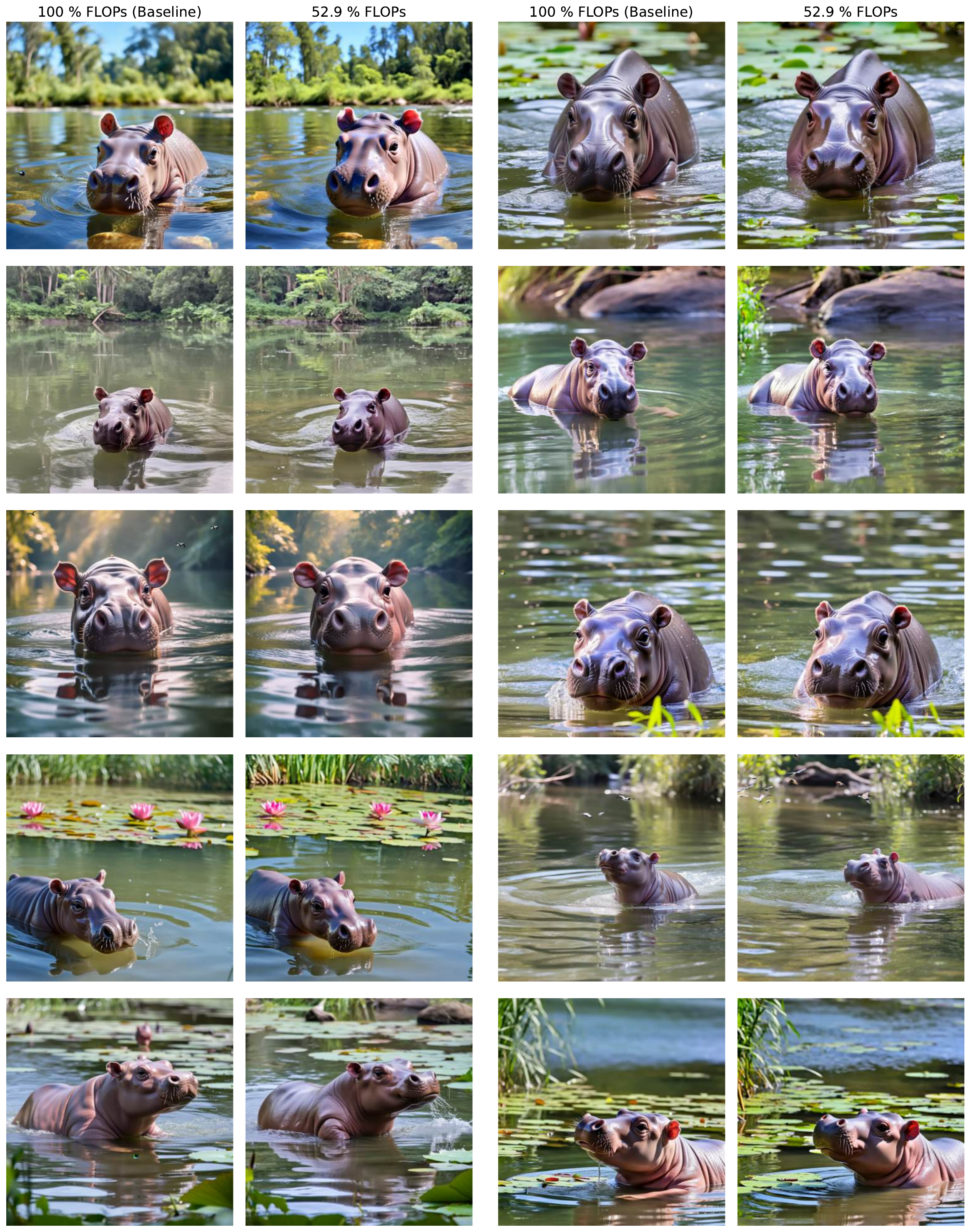}
    \caption{Samples for the prompt: "A baby hippo swimming in the river.". We showcase the image generated by the baseline and our flexible scheduler using only 53\% of FLOPs.}
    \label{fig:emu-hippo}
\end{figure}

\begin{figure}[!h]
    \centering
    \includegraphics[width=0.95\linewidth]{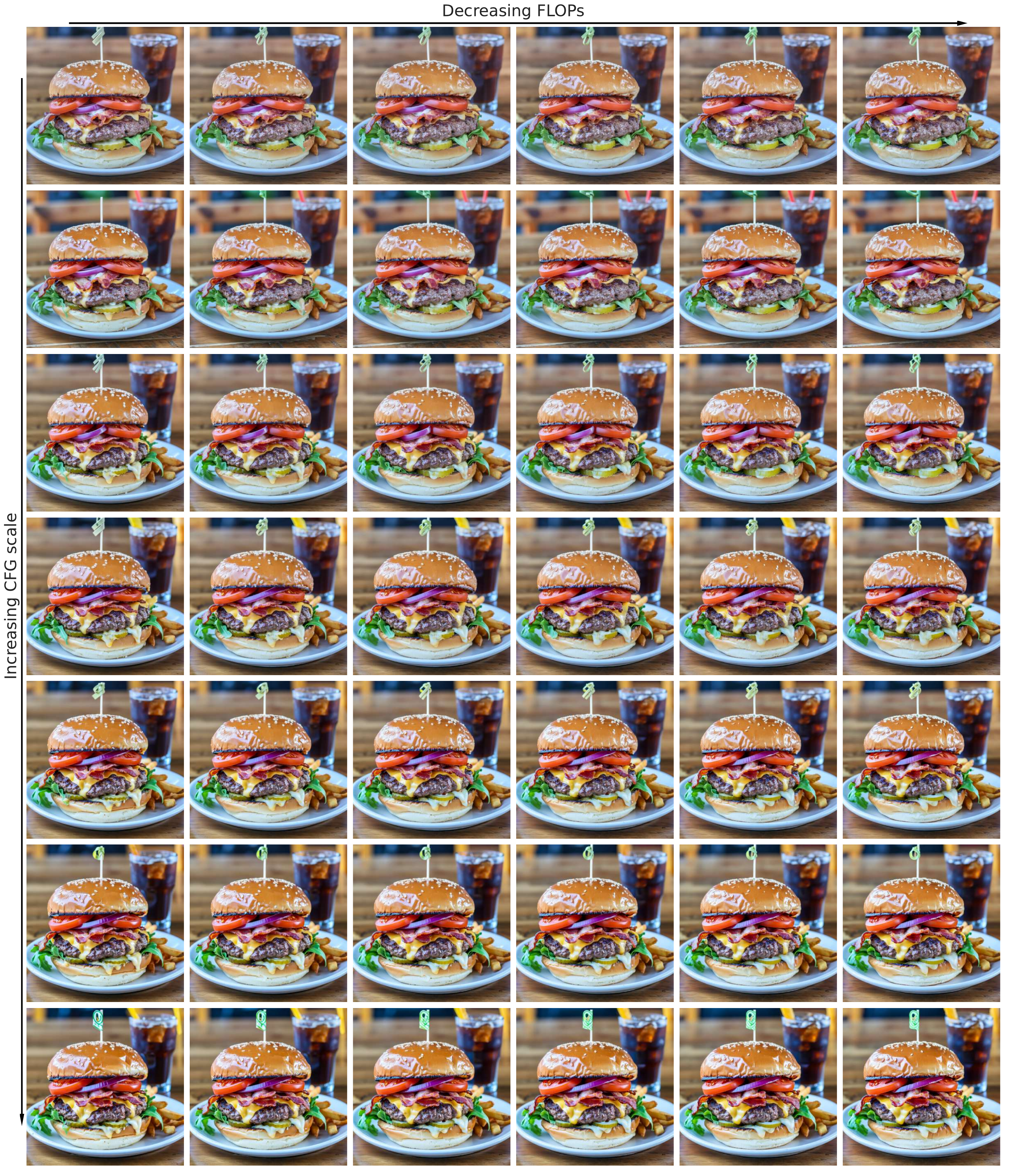}
    \caption{Effect of CFG and total compute for the prompt: `The image shows a gigantic juicy burger placed on a white plate on a wooden table. The burger is composed of a large beef patty, crispy bacon, melted cheese, lettuce, tomato, onion, pickles, and a slice of red tomato, all sandwiched between a soft bun. The burger is so large that it occupies most of the plate, with some toppings falling out of the sides. The bun is slightly toasted, and the cheese is melted to perfection, giving off a savory aroma. The burger is garnished with a side of crispy fries and a refreshing glass of cola.`.}
    \label{fig:grid_image}
\end{figure}

We showcase more examples with varying amounts of compute in Fig.~\ref{fig:emu-examples-1},~\ref{fig:emu-examples-2} and~\ref{fig:emu-examples-3}. We further show more examples of how performance and diversity in image generation are preserved in Fig.~\ref{fig:emu-kitten} and~\ref{fig:emu-hippo}. Finally, we show examples of the effect of CFG scale $\guidancescale$ and reducing the overall number of FLOPs used to generate an image using our method, in Fig.~\ref{fig:grid_image}. For all the images seen in the paper with our~\emu~model, we use $50$ steps of the DDIM scheduler.

\subsection{Text-Conditioned Video Generation}

\begin{figure}[!h]
    \centering
    \includegraphics[width=0.92\linewidth]{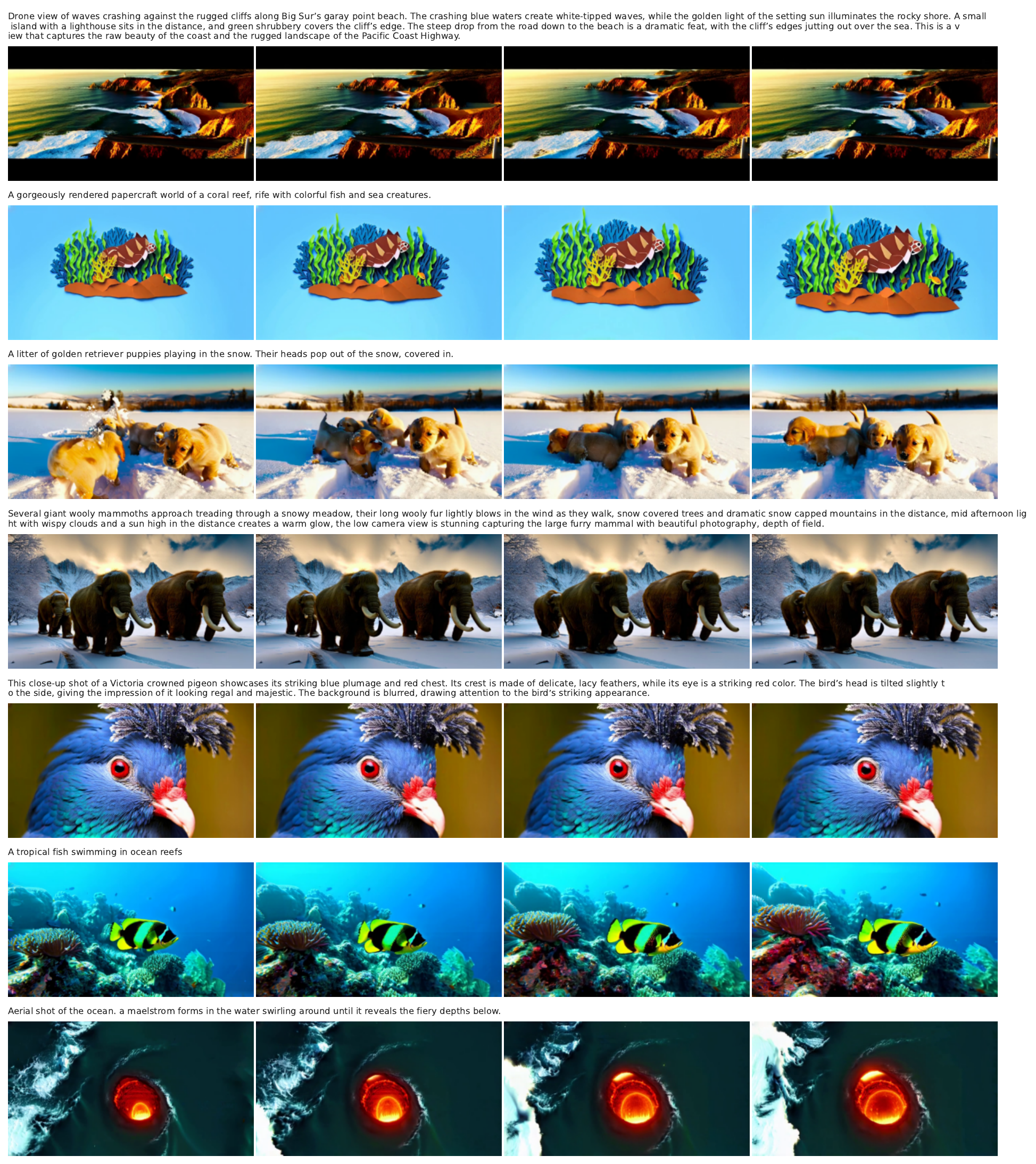}
    \caption{More samples generated by our~\moviegen~model, using $25.2$ \% compute compared to the pre-trained baseline.}
    \label{fig:moviegen-examples-1}
\end{figure}

\begin{figure}[!h]
    \centering
    \includegraphics[width=0.92\linewidth]{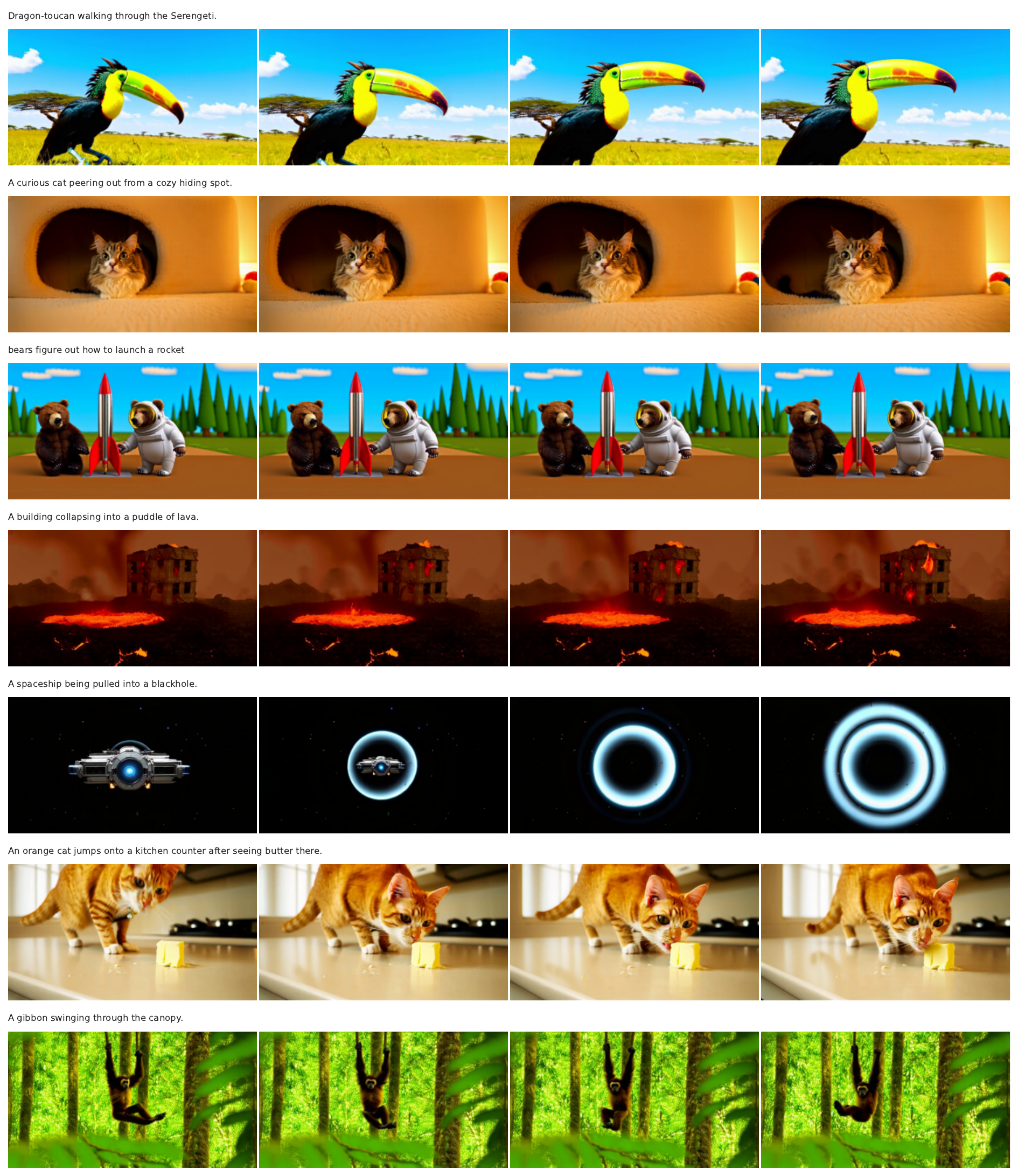}
    \caption{More samples generated by our~\moviegen~model, using $25.2$ \% compute compared to the pre-trained baseline.}
    \label{fig:moviegen-examples-2}
\end{figure}

We showcase more examples of video generation with our flexible model in Fig.~\ref{fig:moviegen-examples-1} and~\ref{fig:moviegen-examples-2}.

\subsection{Class-Conditioned \ImageNet~Generation}

\begin{figure}[!h]
    \centering
    \includegraphics[width=0.88\linewidth]{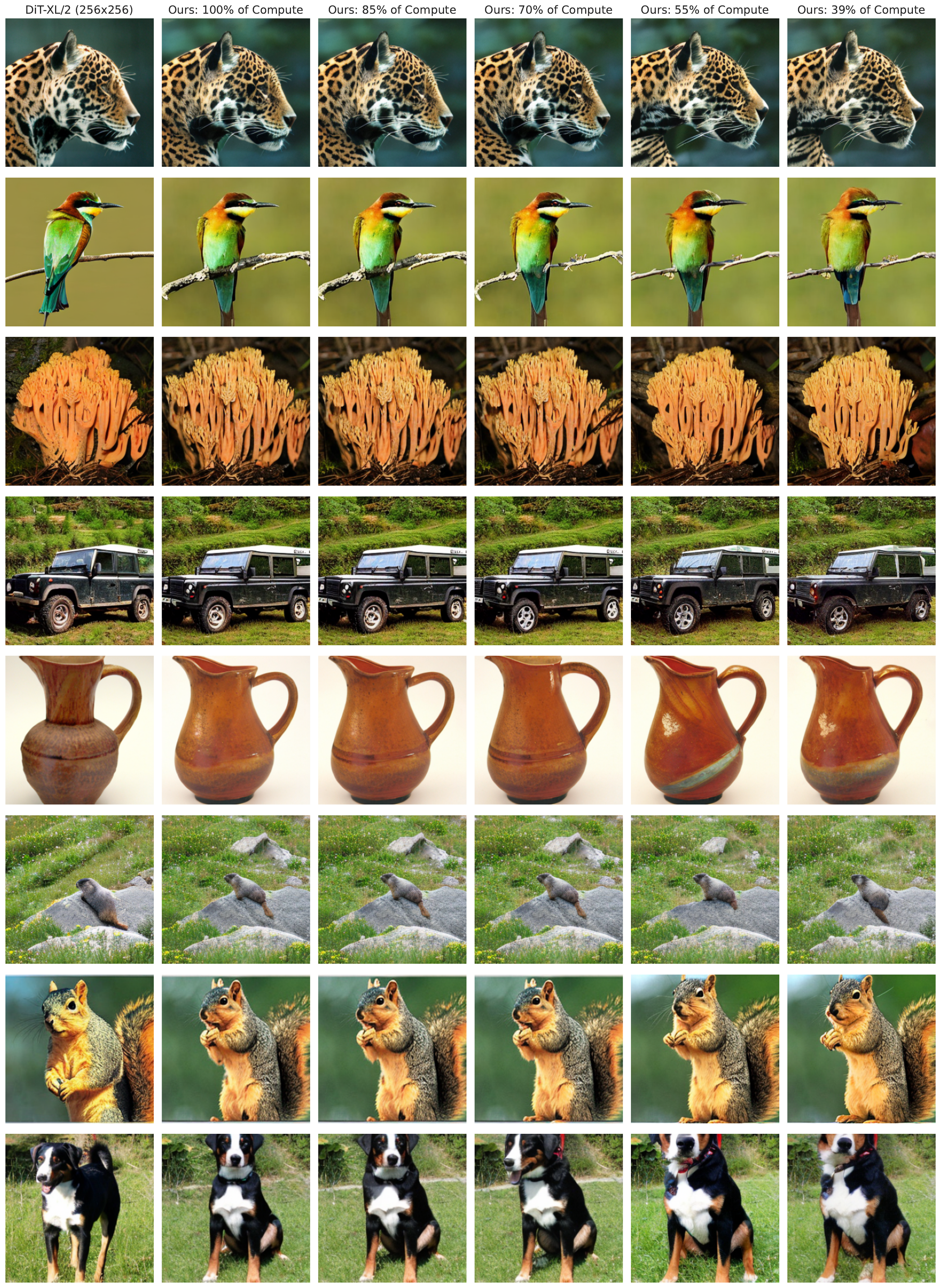}
    \caption{Sample for the \ImageNet~dataset comparing the baseline model and varying inference schedulers of our model, using different levels of compute.}
    \label{fig:sample-xl-256}
\end{figure}

\begin{figure}[!h]
    \centering
    \includegraphics[width=0.95\linewidth]{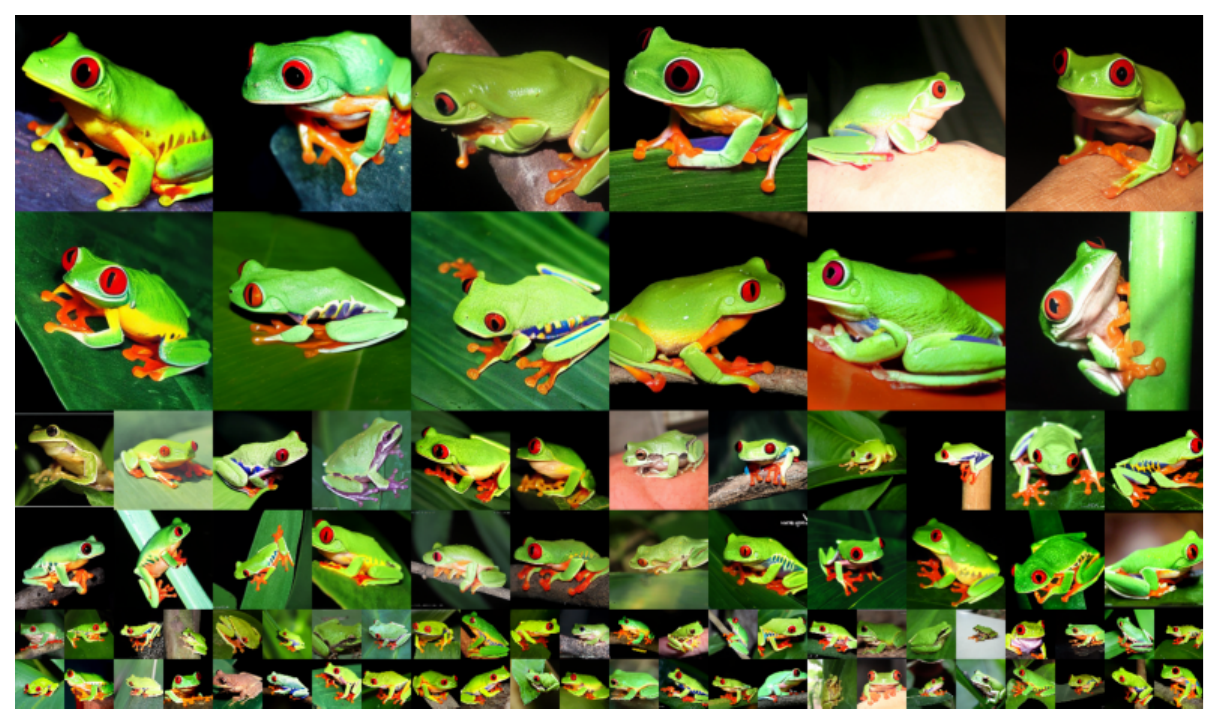}
    \caption{Samples for the \ImageNet~class `tree frog, tree-frog' from our~\flexidit~model that uses only $46$\% of the compute compared to the baseline model.}
    \label{fig:sample-xl-256-31}
\end{figure}

\begin{figure}[!h]
    \centering
    \includegraphics[width=0.95\linewidth]{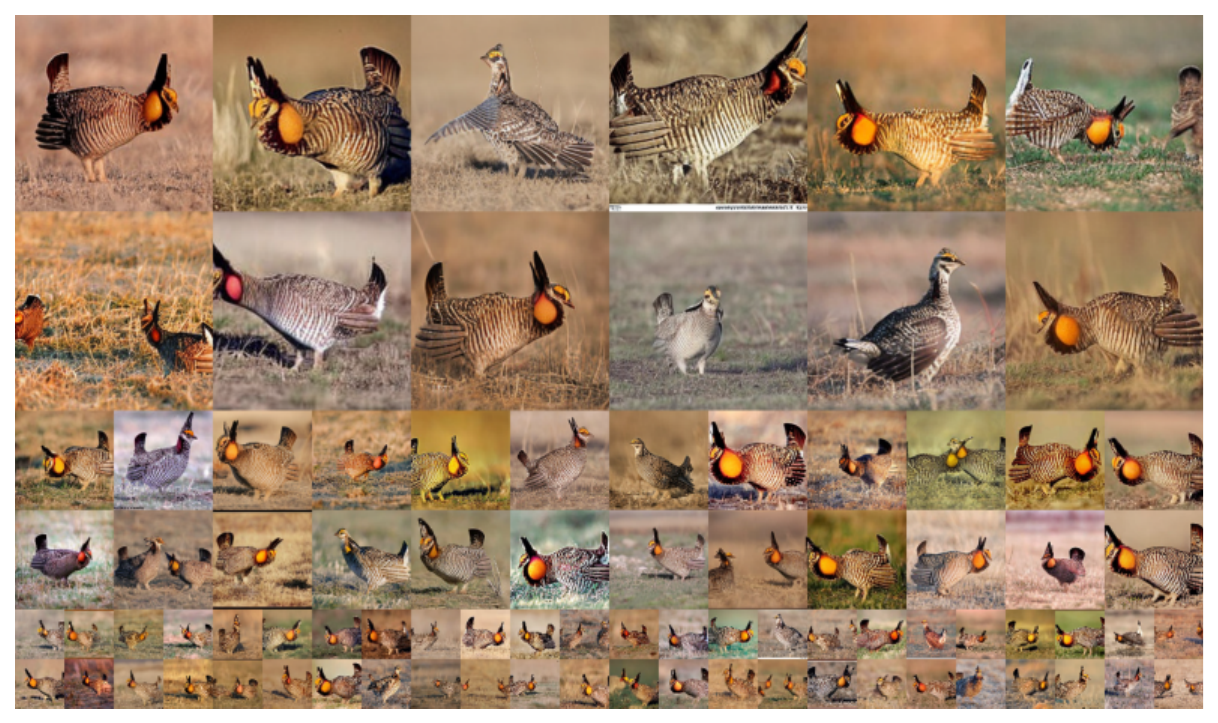}
    \caption{Samples for the \ImageNet~class `prairie chicken, prairie grouse, prairie fowl' from our~\flexidit~model that uses only $46$\% of the compute compared to the baseline model.}
    \label{fig:sample-xl-256-83}
\end{figure}

\begin{figure}[!h]
    \centering
    \includegraphics[width=0.95\linewidth]{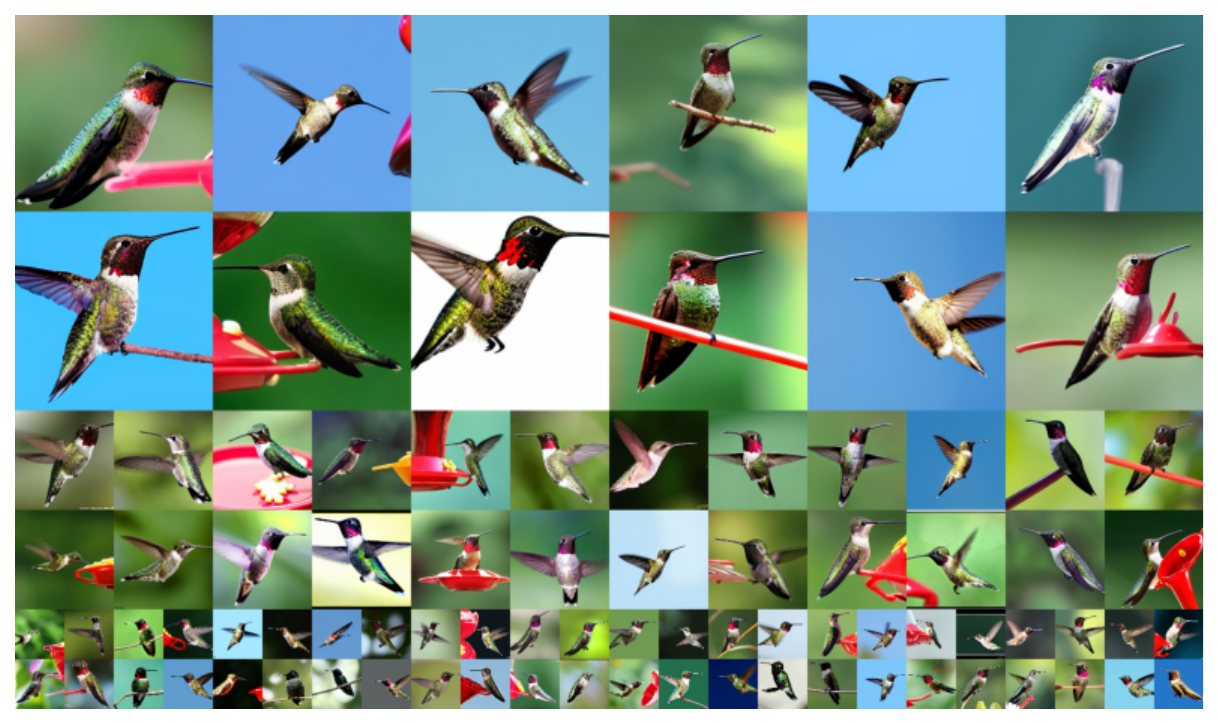}
    \caption{Samples for the \ImageNet~class `hummingbird' from our~\flexidit~model that uses only $46$\% of the compute compared to the baseline model.}
    \label{fig:sample-xl-256-94}
\end{figure}

\begin{figure}[!h]
    \centering
    \includegraphics[width=0.95\linewidth]{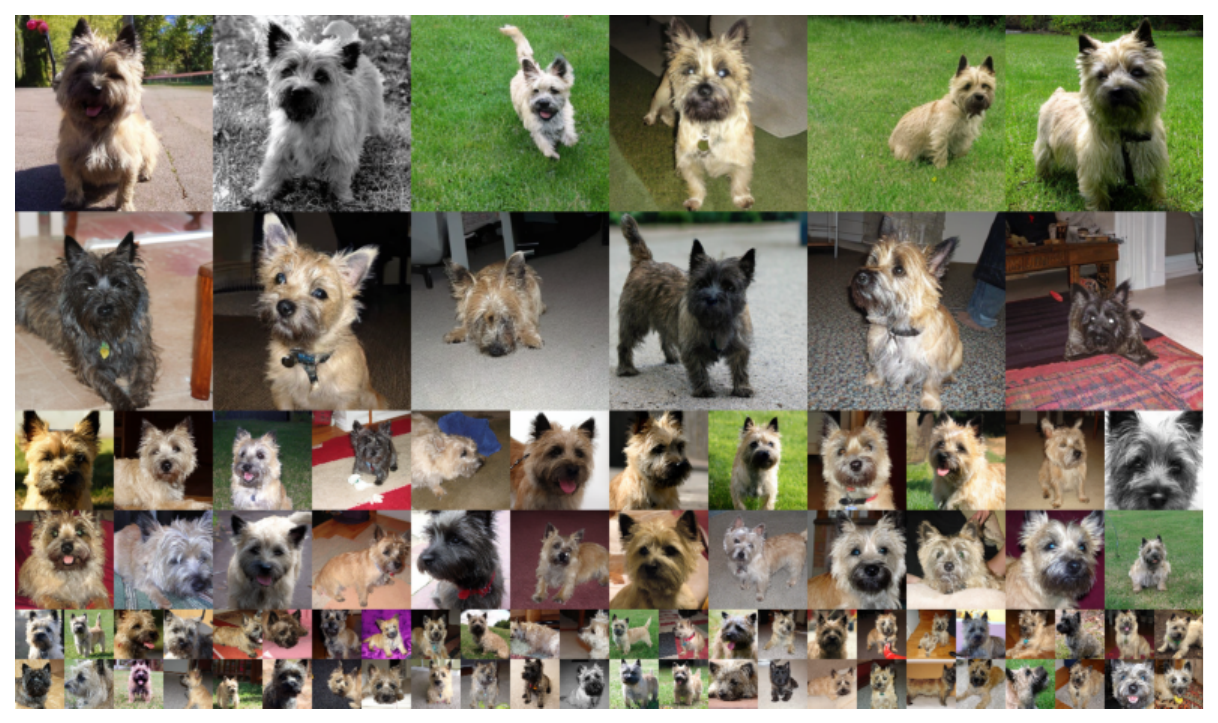}
    \caption{Samples for the \ImageNet~class `cairn, cairn terrier' from our~\flexidit~model that uses only $46$\% of the compute compared to the baseline model.}
    \label{fig:sample-xl-256-192}
\end{figure}

\begin{figure}[!h]
    \centering
    \includegraphics[width=0.95\linewidth]{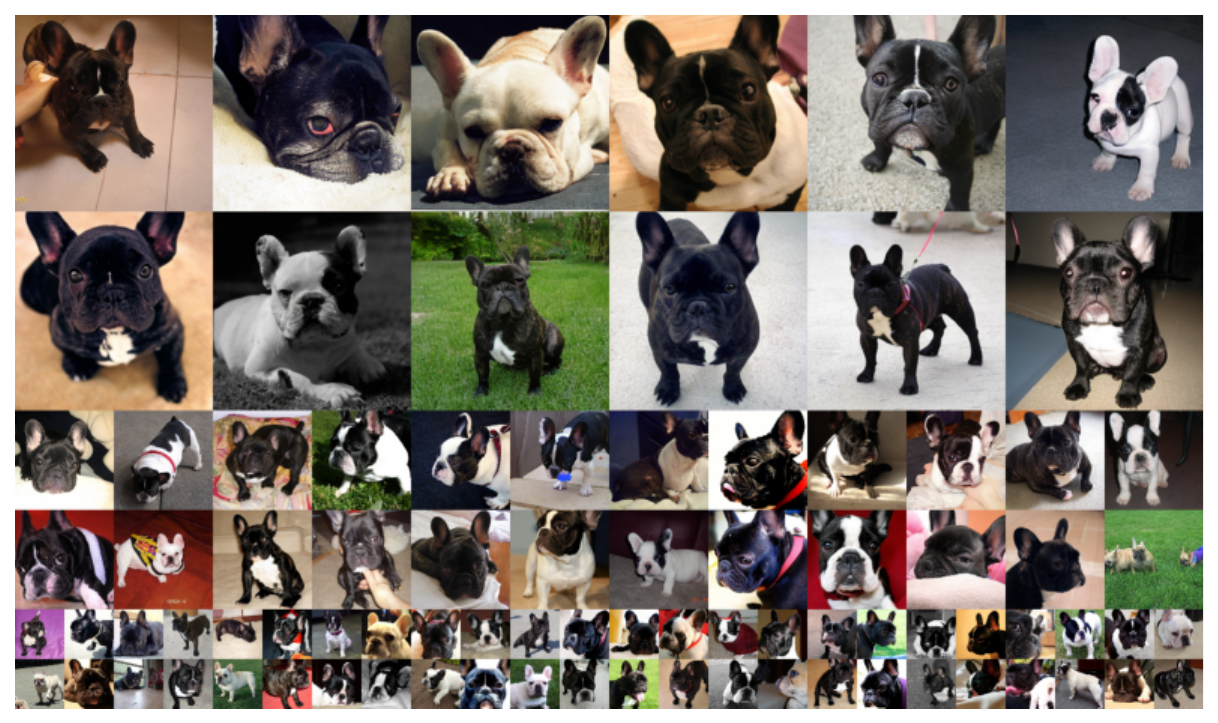}
    \caption{Samples for the \ImageNet~class `French bulldog' from our~\flexidit~model that uses only $46$\% of the compute compared to the baseline model.}
    \label{fig:sample-xl-256-245}
\end{figure}

\begin{figure}[!h]
    \centering
    \includegraphics[width=0.95\linewidth]{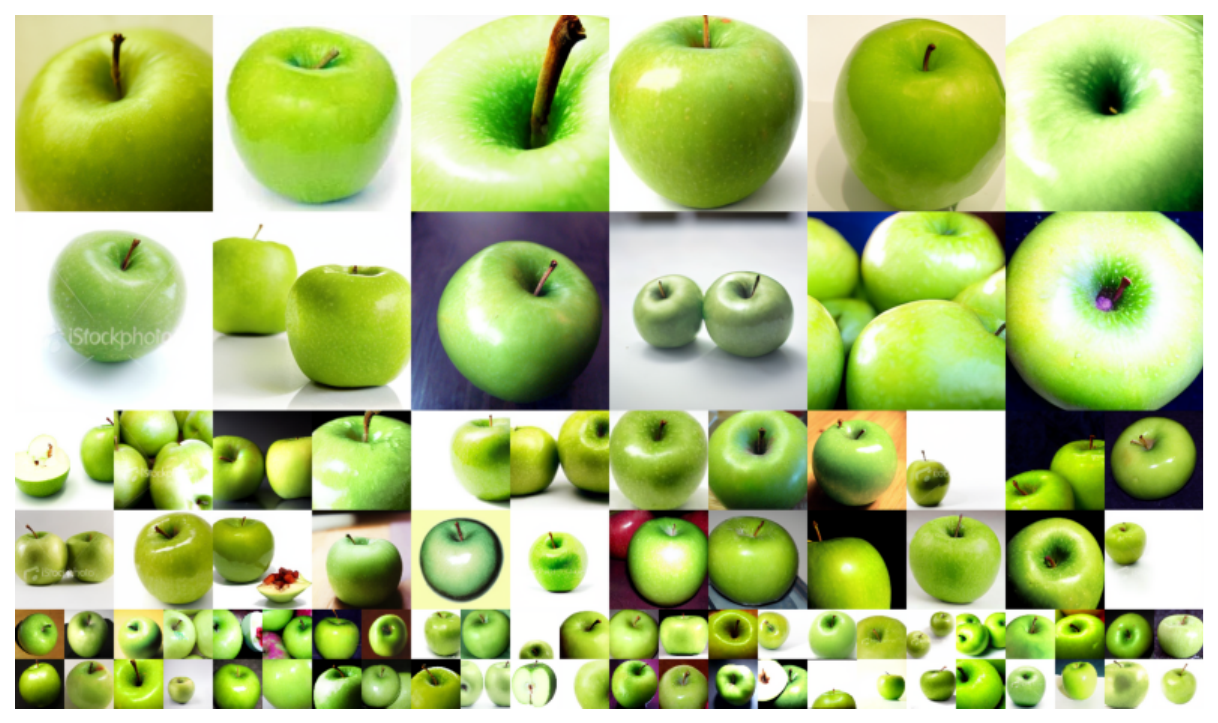}
    \caption{Samples for the \ImageNet~class `Granny Smith' from our~\flexidit~model that uses only $46$\% of the compute compared to the baseline model.}
    \label{fig:sample-xl-256-948}
\end{figure}

\begin{figure}[!h]
    \centering
    \includegraphics[width=0.95\linewidth]{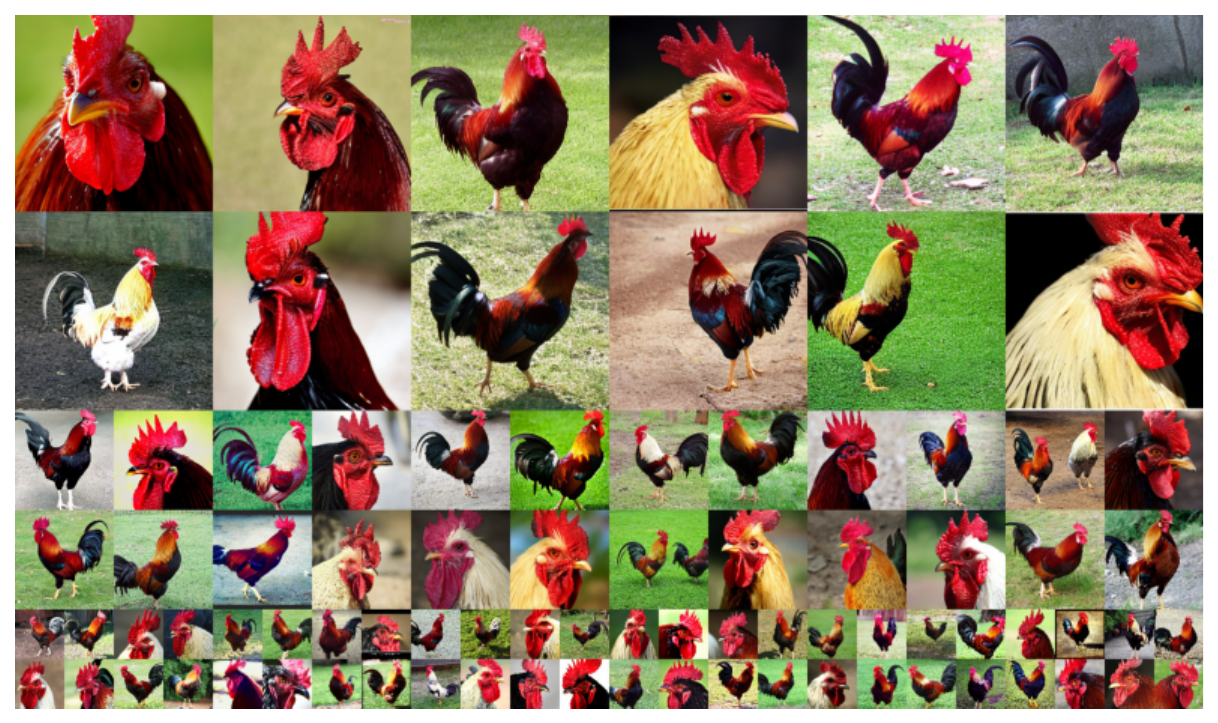}
    \caption{Samples for the \ImageNet~class `cock' from our~\flexidit~model that uses only $46$\% of the compute compared to the baseline model.}
    \label{fig:sample-xl-256-7}
\end{figure}

We provide a more comprehensive comparison for generated images from the same original sample from $\pnoise$, using varying amounts of compute from our flexible model in Fig.~\ref{fig:sample-xl-256}. Note that for class-conditioned generation we do not use LoRAs, and images generated from the original baseline model may not be exactly the same. They do however have the same characteristics (FID score) as seen in Sec~\ref{sec:class_conditioned_experiments}. We also show samples of our flexible \textit{DiT-XL/2} model when using only $64$\% of the compute of the original model in Fig.~\ref{fig:sample-xl-256-31},~\ref{fig:sample-xl-256-83},~\ref{fig:sample-xl-256-94},~\ref{fig:sample-xl-256-192},~\ref{fig:sample-xl-256-245},~\ref{fig:sample-xl-256-948} and~\ref{fig:sample-xl-256-7}. All images are generated using $250$ DDPM steps and a CFG-scale equal to $4.0$. 

